\documentclass[twocolumn, switch]{article} 

\usepackage{preprint}

\usepackage[utf8]{inputenc}	
\usepackage[T1]{fontenc}	
\usepackage{amsmath, amsthm, amssymb, amsfonts, mathrsfs}
\usepackage{algorithm}
\usepackage{algorithmicx}
\usepackage{algpseudocode}
\usepackage{xcolor}		
\usepackage{booktabs} 		
\usepackage{nicefrac}		
\usepackage{microtype}		
\usepackage[switch]{lineno}	
\usepackage[colorlinks = true,
linkcolor = cyan,
urlcolor  = blue,
citecolor = cyan,
anchorcolor = black]{hyperref}	
\usepackage{array}
\usepackage{multirow}
\usepackage{textcomp}
\usepackage{stfloats}
\usepackage{url}
\usepackage{verbatim}
\usepackage{graphicx}
\usepackage{graphics}
\usepackage{lipsum}
\usepackage{subcaption}
\usepackage{xcolor}
\usepackage{soul}
\usepackage{tikz,hyperref}

\usepackage[numbers,square]{natbib}
\usepackage{titlesec}
\titlespacing\section{0pt}{12pt plus 3pt minus 3pt}{1pt plus 1pt minus 1pt}
\titlespacing\subsection{0pt}{10pt plus 3pt minus 3pt}{1pt plus 1pt minus 1pt}
\titlespacing\subsubsection{0pt}{8pt plus 3pt minus 3pt}{1pt plus 1pt minus 1pt}

\definecolor{lime}{HTML}{A6CE39}
\DeclareRobustCommand{\orcidicon}{
	\begin{tikzpicture}
		\draw[lime, fill=lime] (0,0) 
		circle [radius=0.16] 
		node[white] {{\fontfamily{qag}\selectfont \tiny ID}};
		\draw[white, fill=white] (-0.0625,0.095) 
		circle [radius=0.007];
	\end{tikzpicture}
	\hspace{-2mm}
}
\foreach \x in {A, ..., Z}{\expandafter\xdef\csname orcid\x\endcsname{\noexpand\href{https://orcid.org/\csname orcidauthor\x\endcsname}
		{\noexpand\orcidicon}}
}

\title{Kinodynamic Motion Planning for Collaborative Object Transportation by Multiple Mobile Manipulators}

\usepackage{xwatermark}
\newwatermark[firstpage,color=gray!90,angle=0,scale=0.28, xpos=0in,ypos=-5in]{*correspondence: \texttt{keshabpatra19@gmail.com}}

\usepackage{authblk}

\author[1\thanks{\tt{asmith@college.edu}}]{Keshab Patra\orcidA{}}
\author[2]{Arpita Sinha\orcidB{}}
\author[1]{Anirban Guha}

\affil[1]{Department of Mechanical Engineering,
	Indian Institute of Technology Bombay,
	Mumbai, Maharashtra, India}
\affil[2]{Center for Systems and Control,
	Indian Institute of Technology Bombay,
	Mumbai, Maharashtra, India}

\begin{document}

\twocolumn[ 
  \begin{@twocolumnfalse} 
  
\maketitle

\begin{abstract}
This work proposes a kinodynamic motion planning technique for collaborative object transportation by multiple mobile manipulator robots (MMRs) in dynamic environments. A global path planner computes a linear piece-wise path connecting the start to the goal. We proposed an algorithm that aids the path planner in defining the narrow regions between the static obstacles and enhances the feasibility of the global path. We formulate a novel online motion planning technique for the trajectory generations, minimizing the control efforts in a receding horizon manner. The motion planner plans the trajectory for finite time horizons considering the kinodynamic constraints and the static and dynamic obstacles. The motion planner jointly plans for the mobile bases and the arms to utilize the locomotion capability of the mobile base and the manipulation capability of the arm efficiently. We have introduced a convex cone approach to avoid self-collision of the formation by modifying the MMR’s admissible state without imposing any additional constraints. Numerical simulations and hardware experiments showcase the efficiency of the proposed approach.
\end{abstract}
\vspace{0.35cm}

  \end{@twocolumnfalse} 
] 



\section{Introduction}
Robotic systems became ingrained in automated manufacturing, unmanned exploration, and other areas that require risky, tiresome, and hazardous tasks. Sometimes, it becomes impossible for a single robot to accomplish the tasks like transporting oversized or heavy objects \cite{2025_Li} and fixture-less multipart assembly \cite{2018_Venkatesan} without becoming excessively large and costly. Such tasks include flexible manufacturing systems for heavy automotive and aerospace industries, reactor maintenance, construction, loading-unloading \cite{2023_Fan} and towing \cite{2008_Cheng} heavy object. A system of cooperative MMRs enhances the entire system's coverage, flexibility, and redundancy. As a downside, the complexity of robot coordination, communication, and motion planning increases. 

Object transportation with multiple MMRs leverages the mobile robots' locomotion \cite{2022_Shantanu} and manipulators' manipulation abilities  with coupled planning for the mobile base and the manipulators. The coupled planning provides advantages in workspace utilization and planning time \cite{2021_Spahn}. It also eliminates the need for coordination between two separate planners. The interaction between the mobile base and the manipulator's arm entails an integrated kinodynamic modeling method. Besides, the rigid grasping of the object by the end effector (EE) of the MMRs introduces a set of constraints and degrades the Degrees of Freedom (DoF). A motion planner must synchronize the motion of the mobile base and the manipulator for all the MMRs participating in the object transportation to maintain a stable formation. Obtaining a feasible collision-free motion plan is challenging in the presence of static and dynamic obstacles. We address the challenges of motion planning and trajectory generation problems for object transportation by multiple MMRs in an environment with static and dynamic obstacles.

In prior work on motion planning for cooperative multiple MMRs, the geometric path planning approaches \cite{2018_Cao}, hierarchical path planning \cite{2025_zhang,2022_Vlantis} and distributive trajectory planning algorithms \cite{2020_Shorinwa,2021_Wu} plan motion in environments with static obstacles. The motion planning algorithm for dynamic obstacle avoidance by \textit{Tallamraju et al.} \cite{2019_Tallamraju} conservatively approximates the obstacles as uniform cylinders that is highly limiting for navigation in a tight space. Constrained optimization based method \cite{2017_AlonsoMora} for the trajectory planning in dynamic environments uses obstacle-free regions around the formation in the position-time space and optimize for the object's pose to contain the robots in that region.
This position-time obstacles free approach often could not define an obstacle-free convex region in an environment with tightly spaced obstacles, hence the planner is unable to compute a collision-free plan in such situations. Also, the above technique did not incorporate the individual MMR's system dynamics. The computed plan may not be accomplished due to the system's kinodynamic limitations. Kinodynamic planning algorithms \cite{1993_Donald} eradicate such infeasibilities. Our proposed method differs from the typical cell decomposition technique \cite{2006_Lavalle} but is similar to the method by \textit{Deits and Tedrake} \cite{2015_Deits}, where the authors employed intersecting convex polytopes for computing trajectories for quad-rotors. The random sampling method proposed by \textit{Deits and Tedrake} \cite{2015_Deits} for obstacle-free region generation could not define narrow door like region.

We propose an end to end two-step motion planning framework for collaborative MMRs that define more accurate obstacle-free region and compute kinodynamic feasible plans using nonlinear optimization in finite receding horizons. Our proposed technique first defines the narrow static obstacle free convex polygons using targeted seeding in the narrow areas like doors and corridors. It improves the chances of finding a global path. Then the remaining obstacle-free region is defined by convex polygons originating from the uniform random seeding. At the intersections of the convex polygons, we compute the feasible formations using nonlinear constraint optimizations. The shortest path through these feasible formations provides a global path for the collaborative-MMRs system and hence the proposed approach differs significantly from the Probabilistic Road Maps (PRM) and pure sampling-based methods like Rapidly exploring Random Tree (RRT). Next, a local online motion planner computes an optimized, kinodynamically feasible trajectory of the MMRs in dynamic environments using the global path as an initial guess. The proposed motion planning technique removes the shortcomings in the literature. The major contributions of this work are in the following:
\begin{enumerate}
	\item \textit{Local optimal motion planning:} We propose an online kinodynamic motion planning algorithm for environments with static and dynamic obstacles using a nonlinear model predictive control (NMPC) scheme. The proposed planner jointly plans for the mobile bases and the arms to optimally utilize the mobile base's locomotion capability and the arm's manipulation capability. This removes the chances of failure in obtaining a feasible motion for the mobile base and arm.
	\item \textit{Global path planner for navigating through narrow regions:} Our proposed algorithm selectively picks up samples to ensure obstacle-free polygons in narrow areas that are difficult to define using random sampling. It improves path planning feasibility in tight and narrow areas e.g. door, corridor.
	\item \textit{Self-collision avoidance:} We incorporate a convex cone based approach for avoiding collisions among the MMRs and with the object without imposing additional constraints in the motion planner.
\end{enumerate}
\section{Related Work}
Early work on collaborative object transportation \cite{1996_khatib} was started in 1996. A virtual linkage model represents the collaborative manipulation systems. It generates closed-chain constraints \cite{2023_Xu} between the object and the manipulators for motion synchronization and coordination. Multi-MMR cooperative manipulation and transportation of an object comprises the centralized \cite{2013_Erhart}, decentralized \cite{2016_Petitti,2018_Culbertson,2018_Verginis}, and distributed \cite{2017_Dai,2018_Marino,2020_Ren} cooperative control scheme.

Calculus of variation-based navigation method \cite{1997_Desai} demonstrated collision-free motion planning for static obstacles avoidance for a two MMRs system. The variational-based method does not scale well with more numbers of MMRs. Dipolar Inverse Lyapunov Functions, combined with the potential field-based navigation function \cite{2003_Tanner}, find collision-free motion in static environments for deformable material transportation by multiple MMRs with meager control over the formation. The non-holonomic passive decomposition method \cite{2013_Yang} splits robots' motion into formation shape, object transportation, and internal motion. It utilizes internal motions for static obstacle avoidance. The method does not incorporate the kinematic and dynamic constraints of the MMRs.

Constrained optimization-based methods \cite{2017_AlonsoMora} for trajectory planning for dynamic environments use obstacle-free regions around the formation in the position-time space and optimize the object's pose to contain the robots in that region. These methods could only plan a collision-free trajectory when the planning horizon is adequately long, and the poses of the robots in the current or the target formation are inside the position-time embedded obstacle-free convex region. A geometric path planning approach was proposed \cite{2018_Cao, 2017_Jiao} for multiple MMRs transporting an object for static obstacle avoidance. A rectangular passageway-based approach \cite{2018_Cao} is used to find the optimal system width and moving direction in the static obstacle-free area. However, these above methods do not incorporate motion constraints and optimality guarantees.

In hierarchical planning technique \cite{2025_zhang}, a centralized layer plans the object motion offline and a decentralized layer explores the redundancy of each robot online in a static environment without considering the motion aspects of the mobile manipulator. A kinematic planning algorithm \cite{2019_Tallamraju} cooperatively manipulates spatial payload. The algorithm utilizes a hierarchical approach that conservatively approximates the obstacles as uniform cylinders and generates a real-time collision-free motion. This approach is highly restrictive for navigation in tight spaces with large heights and polygonal obstacles with high aspect ratios. MPC-based motion planning techniques for static obstacle avoidance have been presented \cite{2017_Nikou,2024_Kennel}. An alternating direction method of multipliers(ADMM) based distributed trajectory planning algorithm \cite{2020_Shorinwa} plans trajectory for the object transportation in a static environment. A distributed formation control technique \cite{2021_Wu} utilizes constrained optimization for object transportation in a static environment. The formation moves along a predefined reference trajectory and only avoids collision with static obstacles. Motion planning for the deformable object transportation by multiple robots in a static environment using optimization was proposed by \textit{Hu et al.} and \textit{Pei et al.} \cite{2022_Hu,2024_Pei}. Hierarchical quadratic programming approach \cite{2021_Koung} and factor graph optimization technique \cite{2024_Jaafar} plan trajectory for the cooperative object transportation. \textit{Tuci et al.} \cite{2018_Tuci} presented a detailed literature for object transportation. A review of cooperative robotic manipulation is presented by \textit{Feng et al.} \cite{2020_Feng} and \textit{Ramalepa et al.} \cite{2021_Ramalepa}. Lack in complete motion planning framework for collaborative MMRs could not drive the application in industries.

Motion planning for generic robotic systems also uses sampling-based motion planners like PRM \cite{1996_Kavraki} and RRT \cite{1998_Lavalle,2001_Lavalle} which can plan for higher DoF configurations. Considering fixed formation configuration, some temporal logic-based multi-robot planning frameworks have also been addressed \cite{2014_Saha,2013_Ulusoy,2022_Dhaval}. These techniques plan for straight-line and rectangular formation specifications and cannot manage the flexibility of formation specifications. A graph-based method like cell discretization, is proposed by \textit{Liu et al.} \cite{2024_Liu} and flexible formation RRT* is proposed by \textit{Borate et al.} \cite{2024_Borate}. These techniques work well for the static environment with global guarantees, but complete re-planning for dynamic environments and poor scalability with the number of robots remains challenging. The random sampling-based approach \cite{2012_Krontiris} computes PRM from a set of formations. 

Artificial Potential Field (APF) and modified APF techniques \cite{2018_Wen,2020_Liu,2022_Pan} plan obstacle-free motion locally in static and dynamic environments. Behavior-based navigation techniques \cite{2024_Liu} plan for static obstacle avoidance. A planning algorithm using abstraction to bind the formation inside a shape has been demonstrated \cite{2004_Belta,2008_Michael}. Rigid formation control \cite{2001_Egerstedt} avoids static obstacles with reactive behavior. Constrained optimization-based static collision avoidance for a single robot was presented \cite{2021_Zhang,2016_Faulwasser}. A reciprocal collision avoidance algorithm \cite{2020_Mao} combined with MPC for multiple robots does not maintain any formation. These generic planning algorithms cannot be used directly as they don't maintain the rigid formation required for collaborative MMRs.

\section{Problem Formulation}
\begin{figure}[htbp]
    \centerline{\includegraphics[width = 190px]{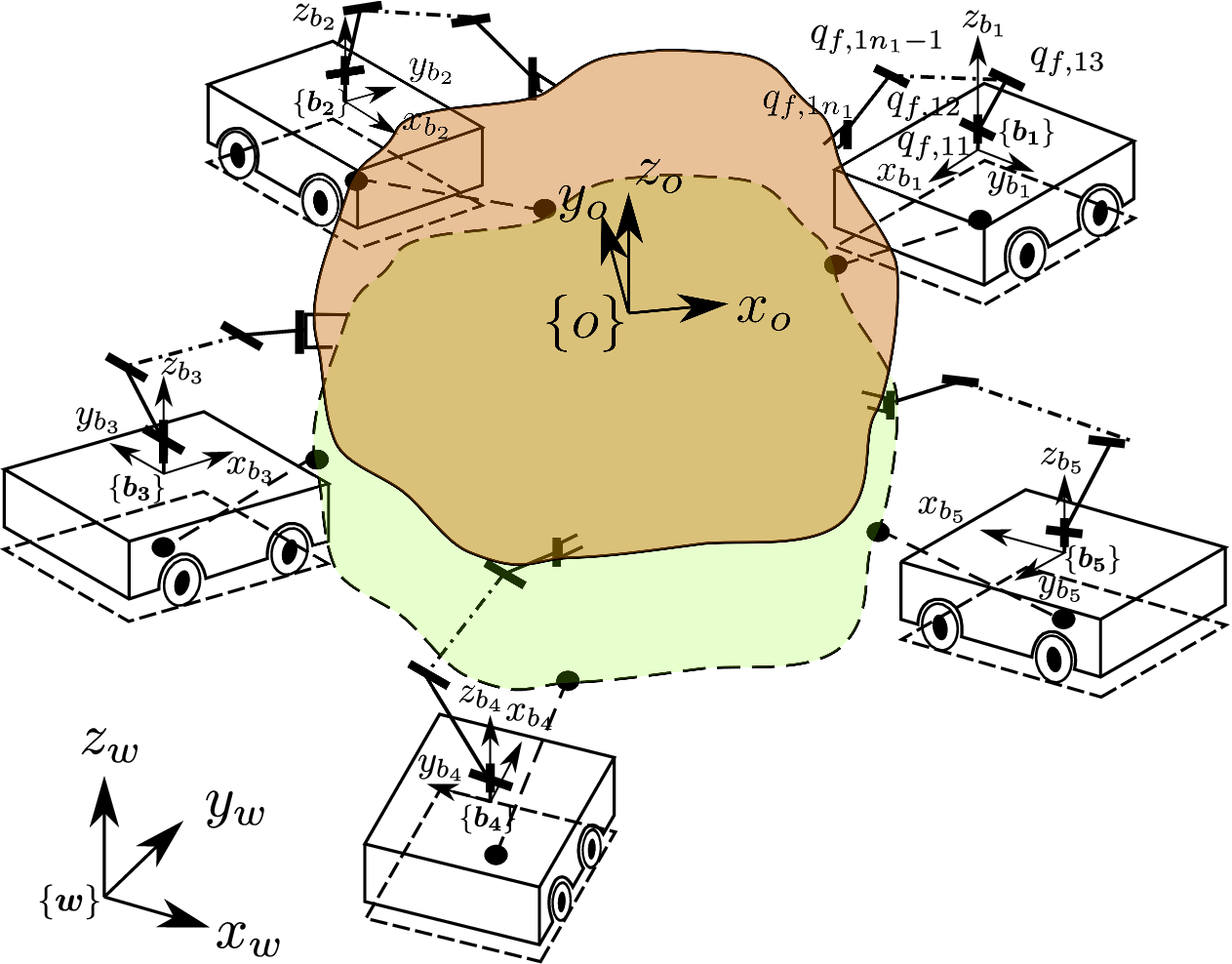}}
    \caption{Formation definition in 3D. The uniform dashed shapes are the projection of the formation in the ground plane.}
    \label{fig:1a}
    \end{figure}

Consider a collaborative object transportation system of $n$ MMRs. Each manipulator arm has $n_i$ no of joints $(\forall i\in[1, n])$. The MMRs grasp an object in the periphery, as shown in Fig. \ref{fig:1a}, to transport it collaboratively from one location to another. The frame of reference $\{\boldsymbol{w}\}$ defines the world fixed frame. An object coordinate frame $\{\boldsymbol{o}\}$ is attached to the object's center of mass (CoM), and each MMR has its own body coordinates $\{\boldsymbol{b}_i\}$ attached to the center of its mobile base. Without specific mention, all the quantities are defined in the world fixed frame $\{\boldsymbol{w}\}$. The collaborative manipulation system is defined in the following subsections.

\subsection{Collaborative MMRs Formation}
\begin{figure}[htbp]
\centerline{\includegraphics[width = 150px]{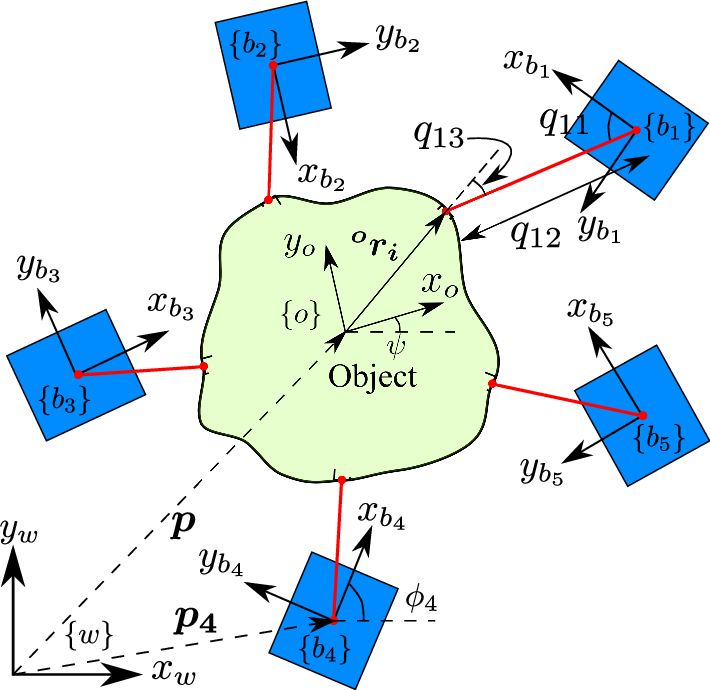}}
\caption{Formation definition in 2D. The blue color rectangles represent MMR bases with their body coordinate. The red line indicates the manipulators. The light green area represents the object to be carried by the MMRs.}
\label{fig:1b}
\end{figure}

The multi-MMRs formation $\mathcal{F}$ shown in Fig. \ref{fig:1a} comprises $n$ MMRs, collaboratively transporting a rigid object. We convert the three-dimensional (3D) object transportation problem into a two-dimensional (2D) problem by projecting the formation into the ground plane (shown with dashed boundaries in Fig. \ref{fig:1a}) with the assumption that the object is at a constant height. Fig. \ref{fig:1b} shows the formation in 2D.
The mobile base of $i$-th MMR is defined with pose $q_{m,i}=[p^T_i,\phi_i]^T$ where $p_i\in\mathbb{R}^2, \phi_i\in\mathbb{R}$ are the position and orientation of the mobile base in $\{w\}$. The body of the mobile base is contained within an area defined by a set of vertices $\mathscr{V}_i$.

The base of $i$-th MMR's arm is at $p_i$. The manipulators are projected in 2D. The projected manipulator is represented with three joints, two revolute and a prismatic. Details are explained in Appendix \ref{append:1}. The joint position is represented by $q_{a,i}=[q_{i1},q_{i2},q_{i3}]^T$ (Fig. \ref{fig:1b}) and the EE's position by $p_{ee,i}\in\mathbb{R}^2$. The EE of $i$ th MMR is grasping the object at position $^or_i$, with orientation $\gamma_i$ defined in
$\boldsymbol{\{o\}}$.

The position and orientation of the object CoM are defined by $p\in\mathbb{R}^2$ and $\psi\in\mathbb{R}$ respectively. The object is within a minimum area defined by a set of vertices $\mathscr{V}_0$. The configuration of $i$ th  MMR is represented by $q_i=[q_{m,i}^T,q_{a,i}^T]^T$. The formation configuration of the collaborative MMRs in the projected 2D plane is defined by $\mathcal{X}=[p^T,\psi,Q^T]^T$, where $Q=[q_1^T,q_2^T,\cdots,q_i,\cdots,q_n^T]^T$ is the combined configuration of $n$ MMRs. The region occupied by the whole formation is defined as a set $\mathcal{B}(\mathcal{X})$.

\subsection{Mobile Manipulator's Model} \label{MMM}
We consider a holonomic MMR. The first-order dynamics model for the mobile base and the arm is considered as $\dot{q}_i = [u^T_{m,i}, u^T_{a,i}]^T$. $u_{m,i}$ and $u_{a,i}$ are the $i$-th MMR's control inputs to the mobile base and manipulator arm joint. Therefore $u_i = [u_{m,i},u_{a,i}]$.

The discrete system representation of the MMR model is defined in Eqn. \eqref{eqn:stf}. The discrete state transition function is computed using a numerical integration scheme.

\begin{equation}\label{eqn:stf}
    q^{k+1}_i =f(q^k_i ,u^k_i)
\end{equation}
$k$ is the discrete time step.
\begin{subequations}\label{eqn:AdmissiableLimits}
        \begin{align}
            \quad & \underline{q}_{a,i}\leq q_{a,i}\leq\overline{q}_{a,i} \label{jsl}\\
            \quad & \underline{u}_{m,i}\leq u_{m,i}\leq\overline{u}_{m,i},
            \underline{u}_{a,i}\leq u_{a,i}\leq\overline{u}_{a,i}
            \label{ajrl}
        \end{align}
\end{subequations}
where $\forall i \in [1,\ n]$, $\underline{q}_{a,i}$ and $\overline{q}_{a,i}$, $\underline{u}_{m,i}$ and $\overline{u}_{m,i}$ and, $\underline{u}_{a,i}$ and $\overline{u}_{a,i}$ are the minimum and maximum joint position of manipulators, mobile base velocity limits and joint velocity vectors' limit respectively. $\mathcal{Q}_i = [\underline{q}_{a,i},\overline{q}_{a,i}]$ and $\mathcal{U}_i = [\underline{u}_i, \overline{u}_i]$ represent the admissible sets of state and control commands respectively. The set of admissible states and control inputs $\mathcal{Q}_i$ and $\mathcal{U}_i$ are indicated by joint position and velocity vectors' limit (Eqn. \eqref{eqn:AdmissiableLimits}).


\subsection{Environment}\label{subsec:Env}
The environment is structured and bounded, having static and dynamic obstacles, and is defined as $W$. The static obstacles are known as apriori and are convex. The start position $p_s$ and the goal position $p_g$ are in the obstacle-free space $W_{free}$ of $W$. An environment is shown in Fig \ref{gor:fig}. The MMRs (green) need to transport an object (blue) cooperatively from the start (bottom-left) to the goal (top-right). Define a set of static convex obstacles $\mathcal{O}\in\mathbb{R}^2$ in $W$. The static obstacle-free workspace is defined by 
\begin{equation}\label{eqn11}
    W_{free} = W\setminus\mathcal{O}\in\mathbb{R}^2
\end{equation}

The set of dynamic obstacles within the sensing zone of the MMRs is defined as $\mathcal{O}_{dyn}$. 
\section{Motion Planning}
\label{MotionPlanning}
\begin{figure*}[htbp]
	\centerline{\includegraphics[width = 460px]{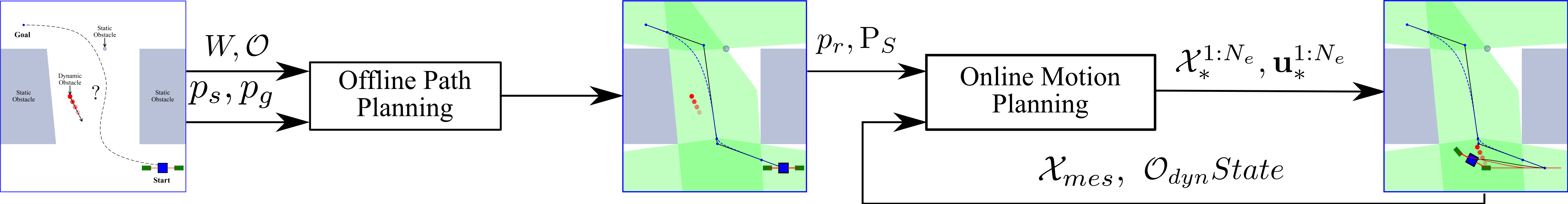}}
	\caption{Motion Planning overview.}
	\label{fig:PlanningSchematic}
\end{figure*}

We solve the motion planning problems for a multi-MMR cooperative object transportation system in a two-step process (Fig. \ref{fig:PlanningSchematic}). In the first step, an offline path planner (Section \ref{Global Path Planning}) computes a feasible linear piece-wise path $(S)$ from the start to the goal location, considering the static obstacles. Then, it computes a continuous reference trajectory $p_r$ from the piece-wise linear path $(S)$ utilizing a cubic Bézier curve with normalized time parameter $c_t \in [0,1]$. In the next step, an online motion planner (Section \ref{OMP}) plans the motion for the MMRs in a receding time horizon. It generates a kinodynamically feasible collision-free trajectory in the dynamic environment using the reference trajectory $p_r$ as an initial guess.

\begin{figure*}[htbp]
	\centerline{\includegraphics[width = 460px]{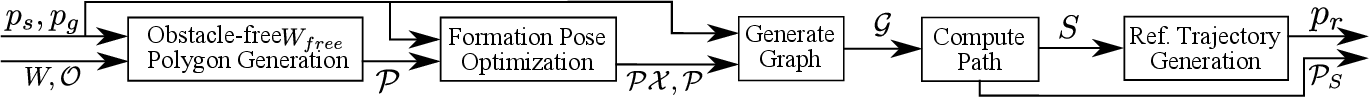}}
	\caption{Path planning process.}
	\label{fig2}
\end{figure*}

\subsection{Offline Path Planning}\label{Global Path Planning}
We define the static obstacle-free region $W_{free}$ by a set of convex polygon $\mathcal{P}$. The method then computes formation configurations $\mathcal{X}$ at each intersection of the convex polygons $\mathrm{P} \in \mathcal{P}$ (Fig. \ref{gor:fig}) utilizing the formation pose optimization (\ref{Formation Optimization}). The feasible configurations $\mathcal{X}$ are added as nodes $\mathcal{V}$ to a graph $\mathcal{G} = (\mathcal{V},\mathcal{E},\mathcal{W})$.  An edge $\mathcal{E}$ between two nodes or configuration exists if both the nodes belong to the same convex polygon. The Euclidean distance between the nodes defines the weight $\mathcal{W}$ of the connecting edge. A graph search method finds the shortest path ($S$) from the start to the goal. The offline path planning technique has been illustrated in the following subsections.

\subsubsection{Static Obstacle-free Region}
We compute a set of convex polygon $\mathcal{P}$ in $W_{free}$ through a sequence of convex optimization. The convex polygon $\mathrm{P} \in \mathcal{P}$ is grown from a seed point utilizing the Iterative Region Inflation by Semi-definite programming (IRIS) method in \cite{2015_Deits}. The path planning algorithm first computes the convex polygons in the narrow regions, like doors and corridors, utilizing targeted seed points and then in the remaining area using uniform random seed points placed in the unexplored areas in $W_{free}$. The targeted seed points for the narrow region are obtained from the midpoint of shortest line segments connecting two static obstacles using Algo. \ref{algo:seeding}.

\begin{algorithm}
	\caption{Seed point generation}\label{algo:seeding}
	\begin{algorithmic}[1]
		\Statex \textbf{Input:} Static Obstacles $\mathcal{O}$
		\Statex \textbf{Output:} Seed Points $\mathcal{L}_{p,seed}$
		\Statex\hrulefill
		\State $\mathcal{L}_{edge} = \emptyset$ \Comment{Initialized an empty list of edges}
		\State $\mathcal{E}_c = \emptyset$ \Comment{ List of connected edges}
		\Statex $\#$ Obtaining the closest point pair between the pair of obstacles
		\For{$\mathcal{O}_i \in \mathcal{O}$}
		\For{$\mathcal{O}_j \in \mathcal{O}$}
		\State $p_{i}, p_{j} = ClosestPoint(\mathcal{O}_i,\mathcal{O}_j),\ \forall i\neq j$
		\State $\mathcal{L}_{edge} \gets \{i, j, p_i, p_j, dist(p_i, p_j)\}$
		\EndFor
		\EndFor
		\Statex $\#$ Finding a list of the shortest edges and vertices connecting all the obstacles
		\State $\mathcal{E}_c =$ ConnectStaticObstacles$(\mathcal{O}, \mathcal{L}_{edge})$ \Comment{Details in Algo. \ref{algo:ConnectStaticObstacles}}
		\Statex $\#$ Sort the edges from min to max of length and find the mid-point of each edge.
		\State $\mathcal{L}_{p,seed} = MidPointSorted(\mathcal{E}_c)$
	\end{algorithmic}
\end{algorithm}

\begin{algorithm}
	\caption{Function: ConnectStaticObstacles$(\mathcal{O}, \mathcal{L}_{edge})$}\label{algo:ConnectStaticObstacles}
	\begin{algorithmic}[1]
		\Statex \textbf{Input:} Static Obstacles $\mathcal{O}$, List of connecting edge $\mathcal{L}_{edge}$
		\Statex \textbf{Output:} List of connecting edged $\mathcal{E}_c$
		\Statex Search the shortest edges connecting obstacles
		\Statex\hrulefill
		\State $\mathcal{O}_{add} = \emptyset, \mathcal{O}_{con} = \emptyset$
		\State $\mathcal{E}_c = \emptyset$
		\State $edge_c = ShortestEdge(\mathcal{L}_{edge})$\Comment{Find the shortest edge}
		\Statex $\#$ Find the obstacles connected by $edge_{c}$
		\State $\mathcal{O}_i, \mathcal{O}_j = ConnectingObstacles(edge_{c})$
		\State $\mathcal{O}_{add}\ =\ \{\mathcal{O}_i, \mathcal{O}_j\},\mathcal{O}_{con} \gets \{\mathcal{O}_i, \mathcal{O}_j\}, \mathcal{E}_c \gets edge_c$ 
		\State $\mathcal{L}_{edge} = \mathcal{L}_{edge}\setminus{edge}_c$ \Comment{Subtract the connecting edge}
		\While{All $\mathcal{O}_{add} \neq \emptyset$}
		\State $\mathcal{O}_{new} = \emptyset$
		\For{$\mathcal{O}_i\ \in \mathcal{O}_{add}$}
		\Statex $\# $ Get the edges from $\mathcal{L}_{edge}$ that are connected to $\mathcal{O}_i$ 
		\State $\mathcal{L}_{edge,\mathcal{O}_i} = ObstacleEdges (\mathcal{L}_{edge},\mathcal{O}_i)$
		\State ${edge}_c = ShortestEdge(\mathcal{L}_{edge,\mathcal{O}_i})$
		\State $\mathcal{O}_i, \mathcal{O}_j = ConnectingObstacles(edge_{c})$
		\State $\mathcal{E}_c \gets edge_c,\mathcal{O}_{new}\gets \mathcal{O}_j$
		\State $\mathcal{L}_{edge} = \mathcal{L}_{edge}\setminus{edge}_c$
		\EndFor
		\State $\mathcal{O}_{add} = \mathcal{O}_{new} \notin \mathcal{O}_{con}$ \Comment{Newly unconnected obstacles}
		\If{$\mathcal{O}_{add}\neq\ \emptyset $}
		\State $\mathcal{O}_{con} \gets \mathcal{O}_{add}$
		\ElsIf{$(\mathcal{O}\setminus\mathcal{O}_{con})\neq\ \emptyset $}
		\State Draw random sample $\mathcal{O}_i \in (\mathcal{O}\setminus\mathcal{O}_{con})$
		\State $\mathcal{O}_{con} \gets \mathcal{O}_i,\ \mathcal{O}_{add} = \mathcal{O}_i$
		\EndIf
		\EndWhile
		\State return $\mathcal{E}_c$
	\end{algorithmic}
\end{algorithm}

\begin{figure}[H]
	\centerline{\includegraphics[width = 120px]{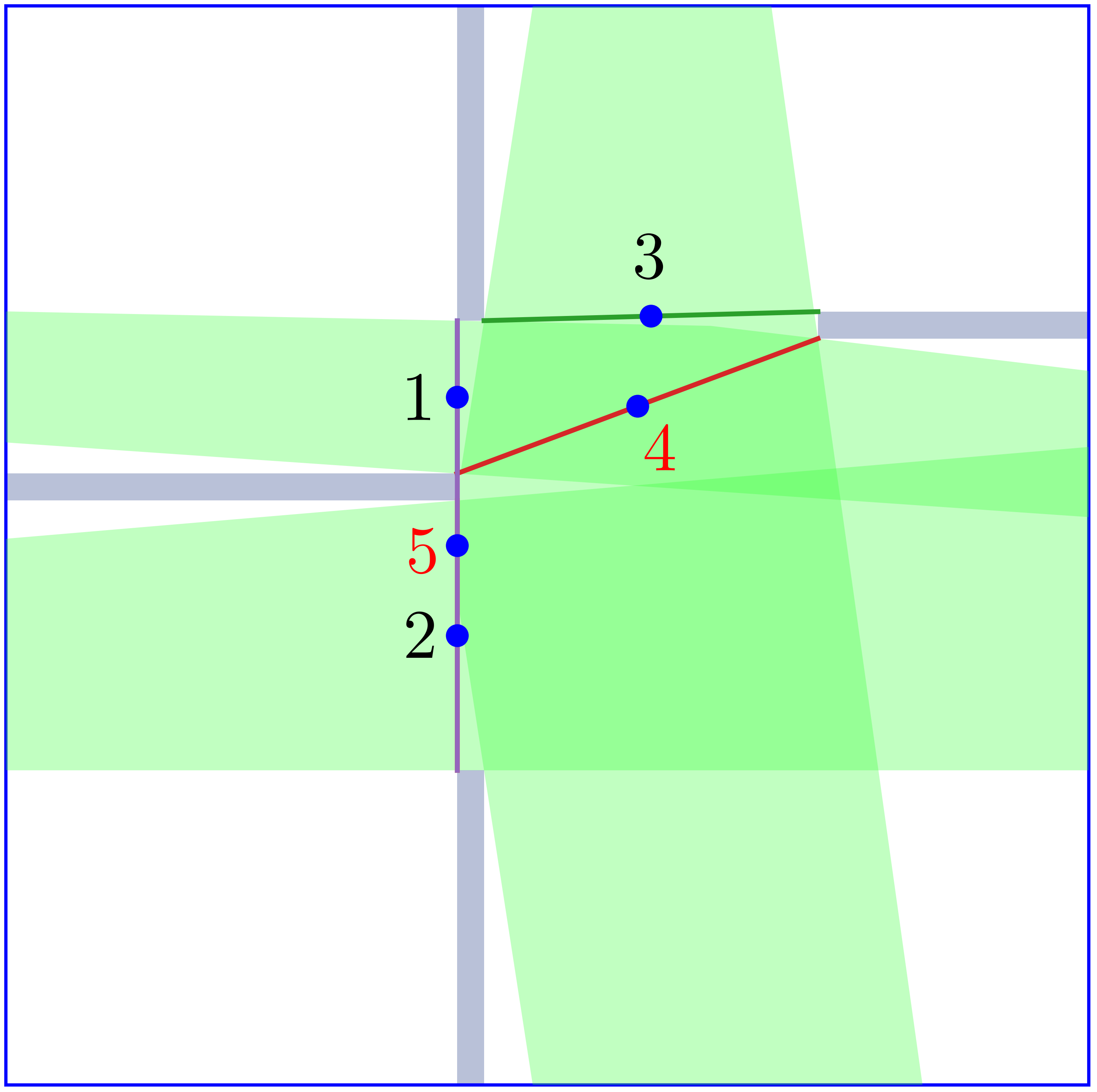}}
	\caption{Seeding Points for the polygon generation in narrow gaps between the obstacle. The points are sorted from 1-5 in accordance to their edge length. Polygon generation starts from point 1, and the points 4, 5 are discarded as they are inside the previously generated polygons}
	\label{fig:seeding}
\end{figure}

\begin{figure}[h]
	\centerline{\includegraphics[width = 200px]{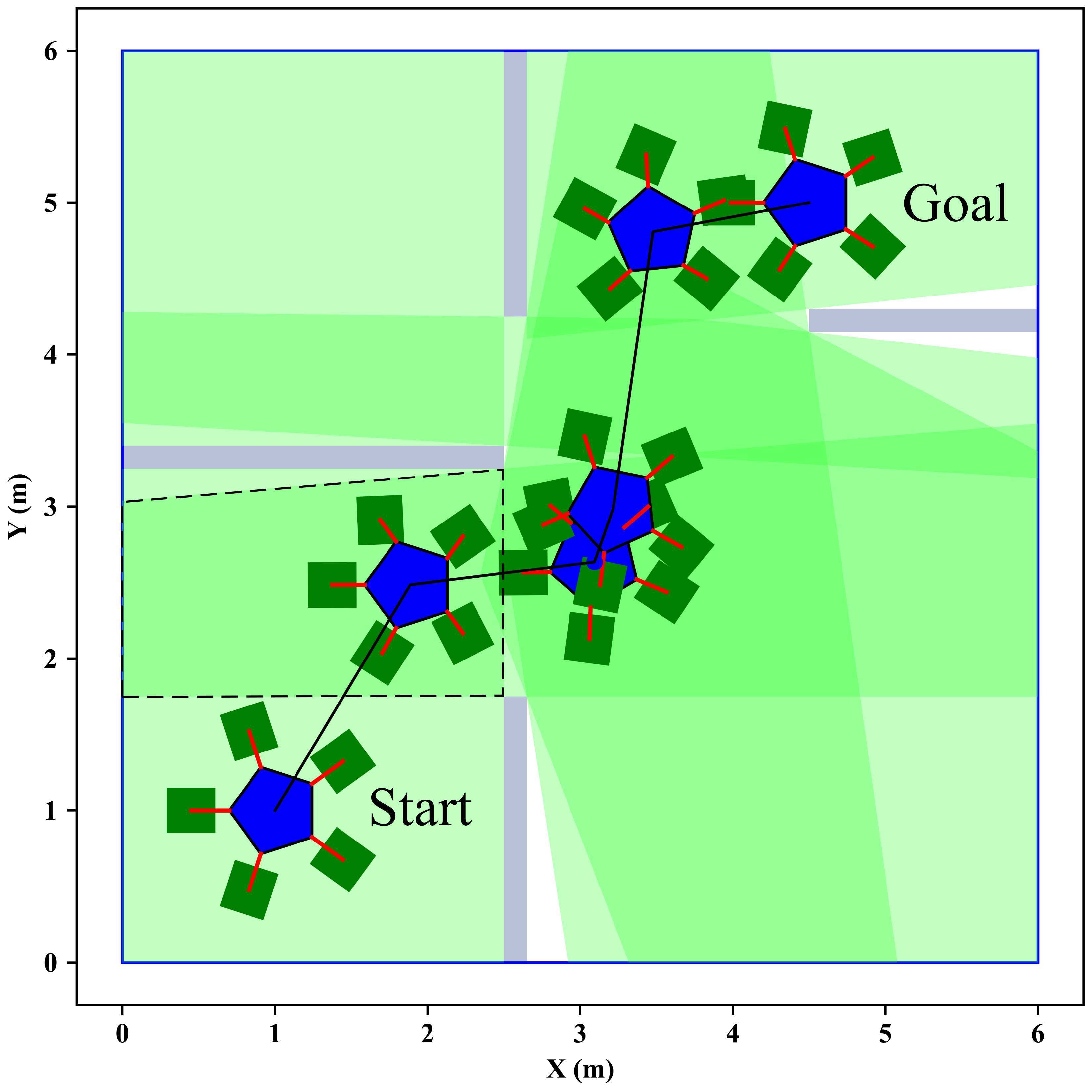}}
	\caption{The thin grey rectangles indicate the static obstacles. The set of green convex polygons define the static obstacle-free regions. The dashed-line area indicates one of the intersection areas of two polygons. The formation is shown at the node points on the resultant path. The black lines indicate the computed global path $(S)$ from the start (bottom-left) to the goal (top-right).}
	\label{gor:fig}
\end{figure}

Consider the static obstacle set $\mathcal{O}$ defined in subsection \ref{subsec:Env}, Algo. \ref{algo:seeding} computes the seed at the midpoints of the shortest line segments or edges connecting two obstacles. The Algo. \ref{algo:ConnectStaticObstacles} finds the shortest edges connecting all the obstacles. A list $\mathcal{L}_{edge}$ of shortest line segment between any two obstacles is computed in line $3-7$. The shortest edge from $\mathcal{L}_{edge}$ is picked to initialize the list of shortest edges $\mathcal{E}_c$ connecting a pair of static obstacles in line $2-5$. The obstacles pair $ \mathcal{O}_{add}\ =\ \{\mathcal{O}_i, \mathcal{O}_j\}$ connected by the edge $\mathcal{E}_c$, computed in line 4, initialize the set of connected obstacles $\mathcal{O}_{con}$ in line 5. The selected edge has been discarded from the edge list $\mathcal{L}_{edge}$ in line 6. In the next step, the shortest lines or edges connecting $\mathcal{O}_{add}$ with the neighbor obstacles $\mathcal{O}_{new}$ are obtained from the remaining edges in $\mathcal{L}_{edge}$ in line $9-15$. The newly found lines or edges are added to $\mathcal{E}_c$ and discarded from $\mathcal{L}_{edge}$ in line $13-14$. $\mathcal{O}_{add}$ is reset with the newly connected obstacles in line 16. The newly found obstacles $\mathcal{O}_{add}$ are added to $\mathcal{O}_{con}$ line $16-18$. The process in line $7-23$ continues until any unconnected obstacle left. During this process if $\mathcal{O}_{add}\ =\emptyset\ and\ \mathcal{O}\setminus\mathcal{O}_{con} \neq\ \emptyset\ $, $\mathcal{O}_{add}$ is drawn randomly from $\mathcal{O}\setminus\mathcal{O}_{con}$ in line $19-22$. While all the obstacles are connected, the $\mathcal{E}_c$ is returned to Algo. \ref{algo:seeding} line 9. In line 10, $\mathcal{E}_c$ is sorted from $\min$ to $\max$ by their length, and their midpoints are computed for defining specific seed points list $\mathcal{L}_{p,seed}$. The computed seed points are shown in Fig.\ref{fig:seeding}.

The convex polygon generation starts from the sorted seed points e.g., in Fig. \ref{fig:seeding} from points $1-5$, but points 4 and 5 are in already generated polygons and are discarded from the seed point list. The convex polygon generation method using only random seed point \cite{2015_Deits,2017_AlonsoMora} could not certainly define the narrow region in $W_{free}$. Our proposed targeted seeding technique in Algo.\ref{algo:seeding} defines the narrow region by convex polygons with a guarantee. It improves the chances of finding a path through narrow regions, like doors and corridors.

\subsubsection{Static Obstacle Avoidance} \label{soa}
The formation $\mathcal{F}$ remains collision-free from static obstacles when $\mathcal{B}(\mathcal{X})$ is inside of any convex polygon $\mathrm{P} \in \mathcal{P}$. It implies that the vertices of MMRs and the object should remain within $\mathrm{P}$. A set of linear inequality constraints represents the polygon $\mathrm{P}$.

\begin{equation}\label{eqn12}
    \begin{aligned}
        \quad& \mathrm{P}= \{\textrm{x}\in\mathbb{R}^2:\mathbf{A}\textrm{x}\leq \boldsymbol{b}, \mathbf{A}\in\mathbb{R}^{n_f\times 2}, \boldsymbol{b}\in\mathbb{R}^{n_f}\}\\
    \end{aligned}
\end{equation}
where $n_f$ is the number of faces of polygon $\mathrm{P}$ and $\textrm{x}$ is an interior point of the polygon. Here $\mathbf{A} = [\boldsymbol{a}^T_1,\boldsymbol{a}^T_2, \cdots, \boldsymbol{a}^T_{n_f}]^T,\ \mathbf{b} = [b_1,b_2, \cdots, b_{n_f}]^T, \boldsymbol{a}^T_i$ is the normal vector of $i$-th face of the polygon $\mathrm{P}$,\ $b_i$ is the shortest distance of $i$-th face of the polygon $\mathrm{P}$  from the origin of $\{w\}$. 

The vertex set $\mathscr{V}(\mathcal{X})$ contains the vertices of bounding polygons of the object and $n$ MMR base body at a formation configuration $\mathcal{X}$ and is defined by
\begin{equation}
    \mathscr{V}(\mathcal{X}) = \{\mathscr{V}_{0}, \mathscr{V}_{1},\cdots, \mathscr{V}_{n} \}
\end{equation}
All the vertices $\mathrm{v} \in \mathscr{V}(\mathcal{X})$ should remain within the polygon $\mathrm{P}$ and the corresponding constraints is defined in Eqn \eqref{eqn:StaticObstacleVerticesConstr}
\begin{equation}\label{eqn:StaticObstacleVerticesConstr}
    \begin{aligned}
        \mathbf{A}\mathrm{v} \leq \mathbf{b}, \quad & \forall \mathrm{v} \in \mathscr{V}(\mathcal{X})
    \end{aligned}
\end{equation}

To ensure $\mathscr{V} \subset \mathrm{P}$, the number of constraints in Eqn. \eqref{eqn:StaticObstacleVerticesConstr} increases significantly, which results in increased computational complexity. The constraints can be reduced if a bounding circle is considered instead of bounding polygons. The outer vertices of a rigid body are considered to remain within a circumscribing circle defined for the body. The implementation of circumscribing a circle from a polygonal shape is described in the rest of this subsection.

Consider the base of the MMRs to be within a circumscribing circle of radius $r_{base}$, and the center located at $p_i$. Similarly, the object is within a circumscribing circle of radius $r_{obj}$ located at $p$, its CoM. The dashed circles in Fig. \ref{sca:fig} show the enclosed region for one of the mobile bases and the object. A set of linear inequality constraints of a closed convex polygon imposed on the circumscribing circles of MMR's base and the object ensures static obstacles avoidance $\mathcal{B}(\mathcal{X}) \cap \mathcal{O} = \emptyset$ as follows
\begin{equation}\label{eqn:StaticObstacleConstr}
    \begin{aligned}
        \boldsymbol{a}_i^Tp_m \leq b_i - (r_m + d_{safe}), \forall m \in \{base,obj\} \forall i \in [1,n_f]
    \end{aligned}
\end{equation}
where $d_{safe}$ is the safety distance.
We know that if any two circles remain within a convex polygon, the region, defined by their two common external tangents and the circles, would be within the same polytope because of the convexity property. No static obstacle avoidance constraints for the MMR's arms would be required if the the MMR's arm is within the dashed tangent line of the two circumscribing circles shown in Fig. \ref{sca:fig} while the MMR's base and the object remain within the static obstacle-free polygons.

Note: The object and the MMR base can be approximated with other generalized shapes, giving a close bound. It will increase the number of constraints. Hence, we restricted to circles only. The same formulation works for other bounding shapes as well.

\subsubsection{Self Collision Avoidance} \label{sca}
\begin{figure}[htbp]
	\centerline{\includegraphics[width = 180px]{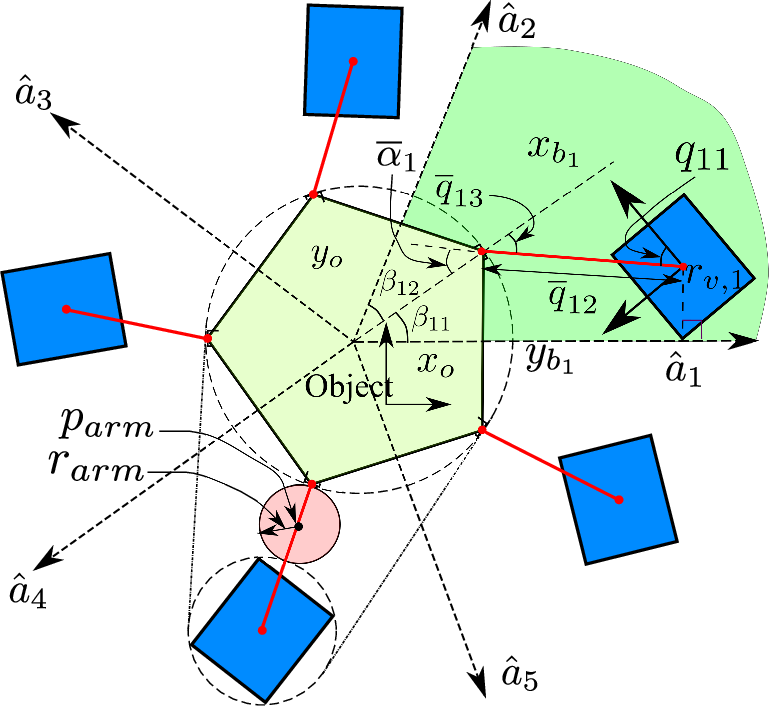}}
	\caption{The convex cone defined by the unit vector $\hat{a}_1$ and $\hat{a}_2$. The dotted enclosing circles for the object, MMR base and light red circle for the manipulator arm are utilized for collision avoidance.}
	\label{sca:fig}
\end{figure}

The MMRs employed in cooperative object transportation must not collide with the neighbor MMRs and the object while transporting and manipulating the object. To avoid such collision, we modify the admissible states of the manipulators of the MMRs based on convex cone geometry that eliminates any additional constraints on the motion planner. We define 2D convex cone $K(\mathbf{E_i},p)$ vertex at $p$ bounded by the unit vectors $\hat{a}_i$ and $\hat{a}_{(i+1)\%n}$ for the $i$ th MMR as shown in Fig. \ref{sca:fig}. The unit vectors $\hat{a}_i$ and $\hat{a}_{(i+1)\%n}$ specify the region of motion for $i$-th MMR free from movements of the object and the other MMRs. The $\hat{a}_i$ and $\hat{a}_{(i+1)\%n}$ is decided based on the relative workspace of MMRs' base fixing the object and the admissible motions of the MMRs respectively. Here, we have divided the region around the periphery of the object into equal convex cones, apex at the CoM of the object for each MMR, as the grasping of the MMR is equispaced and each MMR has same workspace. The 2D convex cone $K(\mathbf{E_i},p)$ vertex at $p$ is defined as

\begin{equation}
	\begin{aligned}
		\quad & K(\mathbf{E_i},p) = \{x\in \mathbb{R}^2: \mathbf{E_i}(x-p) \leq \boldsymbol{0}_2,\ \mathbf{E_i} \in\mathbb{R}^{2\times2}\}\\
		\quad & \mathbf{E_i} = [(\hat{a}^{\bot}_{i+1})^T,
		(-\ \hat{a}^{\bot}_{i})^T]^T
	\end{aligned}
\end{equation}

where $\mathbf{E_i}$ is an affine matrix.

All vertices of the $i$-th MMR must be inside the green region of the $i$-th convex cone in Fig. \ref{sca:fig}. Since the EE is always fixed at a grasp point inside the cone, the base of $i$-th manipulator's arm would also be inside the cone when the vertices of the mobile base remain within the cone.
\begin{equation}
\mathbf{E_i}(\mathrm{v}_{ij}-p) \leq \boldsymbol{0}_2,\quad \mathrm{v}_{ij}\in \mathscr{V}_i\ \forall j \in [i,n_{v,i}]
\label{eqn:convexcone}
\end{equation}
We implement the self-collision avoidance constraints in Eqn. \eqref{eqn:convexcone} by modifying the admissible joint state $q_{a,i}$ of the manipulators. Here, the modified admissible joint limits of $q_{i3}$ satisfy the convex cone constraints. Fig. \ref{sca:fig} shows one of the two extreme configurations of the MMR 1 within its convex cone shown in green. The other MMRs are at any admissible state similar to Fig. \ref{fig:1b}. In the extreme configuration (Fig. \ref{sca:fig}) of MMR 1, the MMR base's furthest vertex touches one side of the cone ($\hat{a}_1$) while the manipulator's arm at its maximum extended position, i.e., the joint 2, reaches to $\overline{q}_{i2}$. The other extreme configuration would be when the MMR touches the other side of the cone ($\hat{a}_2$). Similarly, the extremum of $q_{i3}$ is computed (Eqn. \eqref{eqn:SCAextremum}) from the two extreme conditions. Fig. \ref{sca:fig} shows the upper limit $\overline{\alpha}_1$ of $q_{i3}$ for the MMR 1. The extremum values $\underline{\alpha}_{i}$ and $\overline{\alpha}_{i}$, lower and upper bound for $q_{i3}$ for self-collision avoidance, have been computed utilizing the Eqn \eqref{eqn:SCAextremum}.
\begin{equation}
\begin{aligned}
    \quad & \overline{\alpha}_{i} = \frac{\pi}{2} + \beta_{i1} - \cos^{-1} \left( {\frac{^or_1\sin{\beta_{i1}} - r_{v,i}}{\overline{q}_{i2}}} \right)\\
    \quad & \underline{\alpha}_{i} = - \left[\frac{\pi}{2} + \beta_{i2} - \cos^{-1} \left( {\frac{^or_1\sin{\beta_{i2}} - r_{v,i}}{\overline{q}_{i2}}} \right) \right]
\end{aligned}
\label{eqn:SCAextremum}
\end{equation}
where $\beta_{i1}$ is the angle between the grasp vector $^or_i$ and $\hat{a}_i$, $\beta_{i1}$ is the angle between $^or_i$ and $\hat{a}_{(i+1)\%n}$ and $r_{v,i}$ is the distance between the furthest vertex of the mobile base and mounting point of the manipulator's base. $\overline{\alpha}_{i}$ and $\underline{\alpha}_{i}$ are the angle between $\overline{q}_{i2}$ and $^or_i$ measured at both side of $^or_i$ in the two extreme scenarios mentioned above.
The values of $\underline{\alpha}_{i}$ and $\overline{\alpha}_{i}$ ensure that the MMR remains inside the green region (Fig. \ref{sca:fig}) of the convex cone $K(\mathbf{E_i},p)$. The admissible joint state is updated as $\overline{q}_{i3} = min(\overline{q}_{i3},\overline{\alpha}_{i})$ and $\underline{q}_{i3} = max(\underline{q}_{i3},\underline{\alpha}_{i})$.
The admissible joint movement $\mathcal{Q}_i,( \forall i \in [1,n])$ is adjusted, so the MMRs are in the collision-free zone and follow the physical joint state limits of the manipulator's arm defined in Eqn. \eqref{jsl}.\\

\subsubsection{Grasp Constraints} \label{gc}
The grasping point of EE on the object remain in the same pose and maintain stable formation during collaborative object transportation. The EE's position and orientation must match with the grasp point on the object while the entire formation moves. The grasp constraint for $i$-th MMR is defined in Eqn. \eqref{gpco}.
\begin{subequations}
        \begin{align}
            \quad &  p_{ee,i} = p +{}^w_o\textrm{R}(\psi)^or_i\label{gpc}\\
            \quad &  \phi_i + q_{i1} + q_{i3} = \psi + \gamma_i + \pi
            \label{gpo}
        \end{align}
        \label{gpco}
\end{subequations}
where $p_{ee,i}$ is $i$-th MMR's EE position, ${}^w_o\textrm{R}$ is the rotation matrix of the object frame $\{o\}$ about the world frame $\{w\}$. Eqn. \eqref{gpco} is simplified to $g(\mathcal{X})$ in Eqn. \eqref{go4} and \eqref{lo7}.

\subsubsection{Formation Pose Optimization}\label{Formation Optimization}
We consider the intersection region of the convex obstacle-free polygons to find a formation $\mathcal{F}$ configuration $\mathcal{X}_*$ using a constrained optimization. These formation configurations constitute the intermediate points to form the nodes of a graph $\mathcal{G}$. The cost function of the formation optimization is defined as $J( \mathcal{X})=\lVert p_g-p\rVert^2+\lVert p_s-p\rVert^2$, the total distance from the CoM $p$ of the manipulated object to the start $p_s$ and goal $p_g$ location. It minimizes the deviation of the object's CoM from the straight line connecting the start $(p_s)$ and the goal $(p_g)$ to navigate the formation toward the goal with the minimum path length.

\begin{subequations}\label{go}
        \begin{align}
            \mathcal{X}_*=\quad & \arg\min_{\mathcal{X}} J( \mathcal{X}) \label{go1}\\ 
            \textrm{s.t.} \quad & \mathcal{B}(\mathcal{X})\subset W_{free} \label{go2}\\ 
            \quad & q_{a,i} \in \mathcal{Q}_i , \forall i\in [1,n] \label{go3}\\ 
            \quad & g(\mathcal{X}) = 0 \label{go4}
        \end{align}
\end{subequations}

The static obstacle avoidance constraints in Eqn. \eqref{go2} is defined in Section \ref{soa}. It ensure the formation area $\mathcal{B}(\mathcal{X})$ remains within the $W_{free}$. The constraints in Eqn. \eqref{go3} implements the self-collision avoidance described in Section \ref{sca} and MMR's joint limits. Eqn. \eqref{go4} ensures a stable grasp between the EEs and the object defined in Section \ref{gc} and the kinematic constraints of the manipulator's arm.

\subsubsection{Path Finding}
The algorithm starts with feasible formations computation defined in Section \ref{Formation Optimization}. Along with the start and goal, the computed formations' configuration is stored as the nodes $\mathcal{V}$ of the graph $\mathcal{G}$. The edges $\mathcal{E}$ between these nodes are added if both the nodes are in the same polygon $\mathcal{P}$. Since the polygon is convex, the edges remain within the polygon, thus ensuring the formation remains in $W_{free}$. The Euclidean distance between the nodes defines the weight $\mathcal{W}$ of the edge connecting them. A graph search computes the shortest path $S$ connecting the start and the goal. The computed path is piece-wise linear and static obstacle-free since each edge is inside an obstacle-free convex polygon. A smooth reference path $p_r$ is generated from the piece-wise linear path $S$ utilizing a cubic Bézier curve for the online motion planner.

\subsection{Online Motion Planning} \label{OMP}
The global path planner described in Section \ref{Global Path Planning} ensures a feasible path between the start and goal considering the static environment. It excludes motion planning aspects, MMR dynamics, and dynamic obstacles. Here, we introduce the kinodynamic motion planning technique for the collaborative manipulation system as a non-linear optimization problem using the smoothed version $p_r$ of $S$ as the initial guess. We specify the collision avoidance and kinodynamic constraints of the first-order system dynamics of the MMRs (Eqn. \eqref{eqn:stf}) to compute near-optimal control inputs within the admissible control limits for all the MMRs, enabling the formation to reach the goal successfully.
\begin{figure}[htbp]
\centerline{\includegraphics[width = 220px]{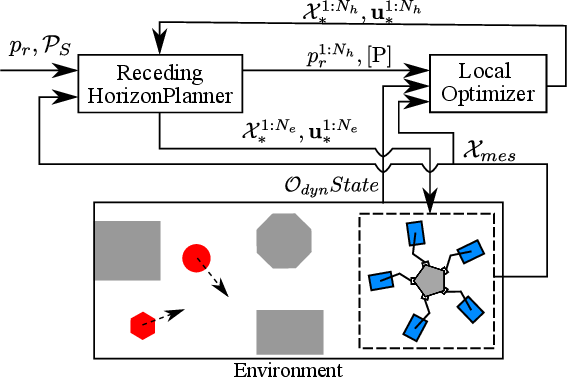}}
\caption{Online motion planning process.}
\label{fig:OMP}
\end{figure}
\subsubsection{Local Optimization:}
The optimization problem for the online motion planning algorithm is formulated in Eqn. \eqref{lo}.

\begin{subequations}\label{lo}
        \begin{align}
            \mathcal{X}^{1:N_h}_*,\mathbf{u}^{1:N_h}_*=\quad & \arg\min_{\mathcal{X},\mathbf{u}} \sum_{k=1}^{N_h-1} J( \mathcal{X}^k,u^k) + J_{N_h} \label{lo1}\\
            \textrm{s.t.} \quad & q^{k+1}=f(q^k,u^k) \label{lo2}\\
            \quad & \mathcal{B}(\mathcal{X})\subset W_{free} \label{lo3}\\
            \quad & \mathcal{B}(\mathcal{X}) \cap \mathcal{O}_{dyn} = \emptyset  \label{lo4} \\
            \quad & q^k_{a,i} \in \mathcal{Q}_i,u^k_i \in \mathcal{U}_i \forall i\in [1, n] \label{lo5}\\
            \quad & g_i(\mathcal{X}) = 0, \forall i\in [1, n]\label{lo7}\\ 
            \quad & \mathcal{X}^0 = \mathcal{X}(0) \label{lo8}
        \end{align}
\end{subequations}
where the superscript $k$ refers to the discrete time steps. In the formulation, Eqn. \eqref{lo1} is the cost function (Section \ref{lcf}), Eqn. \eqref{lo2} represents the first-order system dynamic described in Eqn. \eqref{eqn:stf}. Eqn. \eqref{lo3} and \eqref{lo4} account for the collision avoidance with static (Section \ref{soa}) and dynamic (Section \ref{subsec:doa}) obstacles. The set of admissible states and control inputs (Eqn. \eqref{eqn:AdmissiableLimits}) are defined by Eqn. \eqref{lo5}. The grasp constraints and the manipulator arm's kinematic constraints given in Eqn. \eqref{lo7}. Eqn. \eqref{lo8} initializes the state of the formation.

Once the reference path $p_r$ is obtained from the offline path planner, the online motion planner generates the collision-free trajectory for the formation $\mathcal{F}$ and the motion plans for the MMRs. The online planner solves the optimization problem over a horizon $N_h$ segment with time $T_h$ to reduce the computational burden. The MMRs execute the plan for time $T_e$ $(T_e < T_h)$, and the next re-planning cycle starts after time $T_e$. We choose the execution time $T_e$ so that the computation time for a motion planning horizon is guaranteed to be less than $T_e$ so that the formation can start the subsequent trajectory once it finishes executing the current trajectory.

\subsubsection{Cost Function} \label{lcf}
The offline path planner generates a continuous time normalized reference trajectory $p_r(c_t),\ c_t \in [0,1]$ for the online motion planner. We select the time for traveling the path considering the formation's operational velocity $v_{op}$. The trajectory time has been fused with $p_r(c_t)$ to obtain reference trajectory $p_r(t)$. The predicted and reference position for the CoM of the object is denoted as $p^k$ and $p_r^k(t) = [x,y]_r^k$ for the time step $k$. The tracking error vector $e_k$ is defined as $e^k = p^k - p_{r}^k$.

The cost function for the optimization in Eqn. \eqref{tcf} comprises two components: one is for quadratic reference trajectory tracking error with a weight matrix $(\mathbf{W_e})$ and the other is quadratic control input minimization with a weight matrix $(\mathbf{W_u})$. The weight $(\mathbf{W_e})$ for the tracking is kept relatively low to ensure flexibility for the generated trajectory during dynamic obstacle avoidance. 
The diagonal elements of $\mathbf{W_u}$ for the mobile base and the manipulator arm differ. A lower weightage to the mobile base prioritizes the base motion as it has no restriction on displacement, unlike the manipulator's joints. Assigning lower weight to the base control input provides unrestricted utilization of its faster motion capability and hence provides faster maneuverability. 
Tuning of weight for an MMR system in the simulation provides optimal performance. A change in weight ratio has an impact on solution time and feasibility. Assigning a higher weight ratio to the trajectory tracking error limits the dynamic obstacle avoidance performance and may lead to an infeasible solution. Higher weight to base reduces the maneuverability of the MMRs.
\begin{equation}\label{tcf}
    J( \mathcal{X}_k,u_k)= {u^k}^T\mathbf{W_u}u^k + {e^k}^T\mathbf{W_e}e^k
\end{equation}

The terminal cost $J_{N_h}$ is defined in Eqn. \eqref{tlcf} similar to the tracking error but with a higher weightage. $(\mathbf{W}_{n_h})$.
\begin{equation}\label{tlcf}
    J_{N_h}={e^{N_h}}^T\mathbf{W_{N_h}}e^{N_h}
\end{equation}

\subsubsection{Dynamic Obstacle Avoidance} \label{subsec:doa} 
Dynamic obstacles in any environment can have different shapes, speeds, and paths. In a factory or warehouse-like environment, human and material handling machinery usually move in a linear motion. In this problem, we estimate the position and velocity of the obstacle at the beginning of a prediction horizon $T_h$ and assume the velocity remains the same during $T_h$. Let $d=[1, n_{dyn}]$  denotes the index of the dynamic obstacles $\mathcal{O}_{dyn}$ within the sensing zone.
The shape of a dynamic obstacle is approximated as a circle of radius $r_{dyn,d}, \forall d$.The motion planer takes the predicted future position of the dynamic obstacles.The position of dynamic obstacles is defined as $p^k_{dyn,d}, \forall k \in [1,N_h]$.

The future positions $p^k_{dyn,d}=p_{dyn,d}+v_{dyn,d}k T_c$ are computed employing constant obstacle velocity $v_{dyn,d}$ in planning horizon $N_h$. $p_{dyn,d}$. $v_{dyn,d}$ are estimated at the beginning of each horizon. Any state estimation model for dynamic obstacles can be used without any change in the motion planner. The solution convergence and the solution time might be impacted by abruptly changing motion patterns of the dynamic obstacles. The dynamic obstacles cut the convex solution space, and the abrupt change creates more complexity for the solver, which may add to the solution time and sometimes lead to infeasibility. The possible way forward is to use a higher-order polynomial state estimation model, e.g., constant acceleration, linearly varying acceleration (constant jerk) model, or a constant velocity prediction model using a higher re-planning rate.

The same circumscribing circles defined for static obstacle avoidance in Section \ref{soa} for the base of MMRs and the object are utilized here. For the manipulator's arm, a circle (red circle in Fig. \ref{sca:fig}) of radius $r_{arm,i}$ at the middle of the manipulator is shifted by half radius of the collision circle for the mobile base and is simplified as $p_{arm,i} = p_i + 0.5(1 + r_{base,i}/\ \overline{q}_{i2})(p_{ee,i} - p_i)$. The radius $r_{arm,i} = 0.5(\overline{q}_{i2} - r_{base,i})$ is defined to contain the manipulator arm for all instances. A circle approximating the arm's collision geometry can be conservative in some tightly spaced environments. In such scenarios, multiple circles should be defined.

The nonlinear constraints defined in Eqn. \eqref{eqn:doa} ensures dynamic obstacles avoidance. An additional safety margin $d_{safe,dyn}$ has been introduced to mitigate any uncertainty in the dynamic obstacle's trajectory.
\begin{equation}
    \begin{aligned}
        ||p^k_{dyn,d}-p^k_{m}|| \geq r_{dyn,d} + r_m + d_{safe,dyn}\\
        \forall m \in \{base,obj,arm\}, \forall d \in [1,n_{dyn}],k = [1,N_h]
    \end{aligned}
    \label{eqn:doa}
\end{equation}

\section{Results}
We evaluate the proposed motion planning algorithm in multiple dynamic environments with a simulation study for collaborative object transportation by five MMRs and a hardware experiment with two in-house developed MMRs that accept velocity command $u$ as the control input. Comparisons of the proposed planning algorithm with a similar state-of-the-art motion planning algorithm by \cite{2017_AlonsoMora} and \cite{2025_zhang} have been incorporated to showcase the advantage of the proposed algorithm over the other.

The MMR system dynamics are approximated using the fourth-order Runge-Kutta method to calculate the state transition function mentioned in Eqn. \eqref{eqn:stf}. The MPC problem of the local motion planning and the nonlinear optimization of global planning is solved using the CasADi framework \cite{2019_Andersson} with an Interior point optimization (Ipopt) method.

\subsection{Simulation Setup}
All the MMR bases employed here have the same forwarding and reversing capabilities. The manipulator arm's Denavit–Hartenberg (DH) parameters are mentioned in Table \ref{Tab:0}.

In this study, five MMRs grasp an object at the vertices of the pentagonal object, as shown in Fig. (\ref{fig:1a}), and do not change their initial grasp position during the task execution.

\begin{table}[htbp]
\caption{DH Parameters Value for the MMR used in Simulations}
\label{Tab:0}
\begin{center}
\begin{tabular}{|c|c|c|c|c|}
\hline
\textbf{Joint} & $d\ (m)$ & $a\ (m)$ & $\alpha\ (rad)$& $\theta\ (rad)$\\
\hline
Joint 1 & 0.070 & 0 & 0 & $q_{f,1}$\\
\hline
Joint 2 & 0 & 0 & $0.5\ \pi$ & $q_{f,2}$\\
\hline
Joint 3 & 0.100 & 0 & $- \pi$ & $q_{f,3}$\\
\hline
Joint 4 & 0.125 & 0 & $ \pi$ & $q_{f,4}$\\
\hline
Joint 5 & 0 & 0.120 & $-0.5\ \pi$ & $q_{f,5}$\\
\hline
Gripper & 0 & 0 & 0 & 0\\
\hline
\end{tabular}
\end{center}
\end{table}

We select the operational velocity of the formation $v_{op} = 0.15\ m/s$ and use prediction horizon time $T_h = 6\ s$, trajectory execution time $T_e = 2\ s$ and the discretization time step $T_c = 0.25\ s$. The safety margins  $d_{safe} = 0.05\ m$ and $d_{safe,dyn} = 0.1\ m$ for static and dynamic obstacle avoidance keep the formation safe during object transportation. A higher value restricts the formation from approaching the obstacle thus reducing the obstacle-free space. The optimization weights are  $\mathbf{W_u} = diag(repeat([0.05,0.05,0.25,2.5,2.5,2.5],5))$, $\mathbf{W_e} = diag([0.01,0.01])$ and $\mathbf{W_{N_h}} = 10^3$.

\begin{figure}[H]
 	\begin{subfigure}{0.24\textwidth}
 		\centering
      	\includegraphics[width=115px]{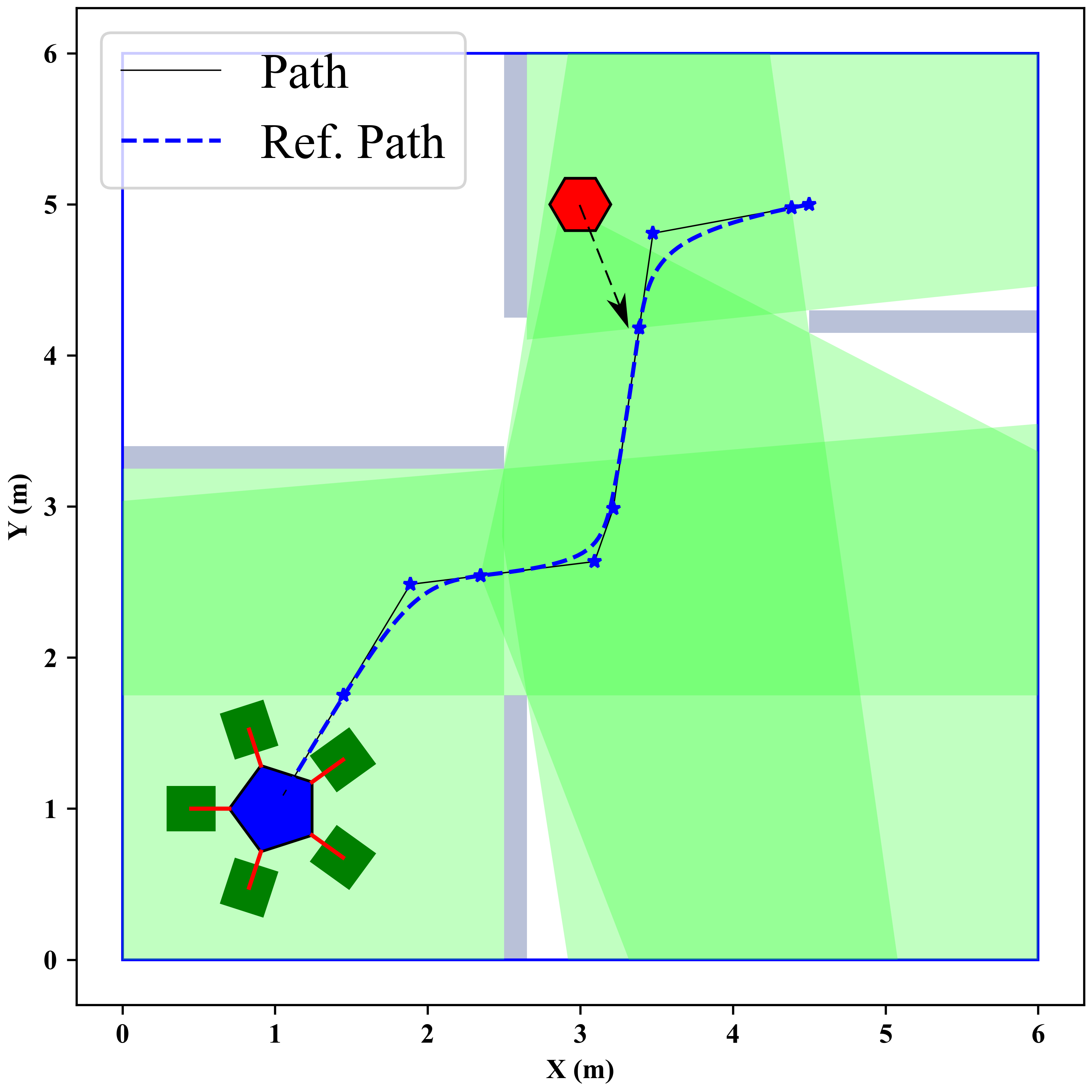}
		\subcaption{\label{fig:5a} $t = 0\ s$}
    \end{subfigure}
    \begin{subfigure}{0.24\textwidth}
    	\centering
    	\includegraphics[width=115px]{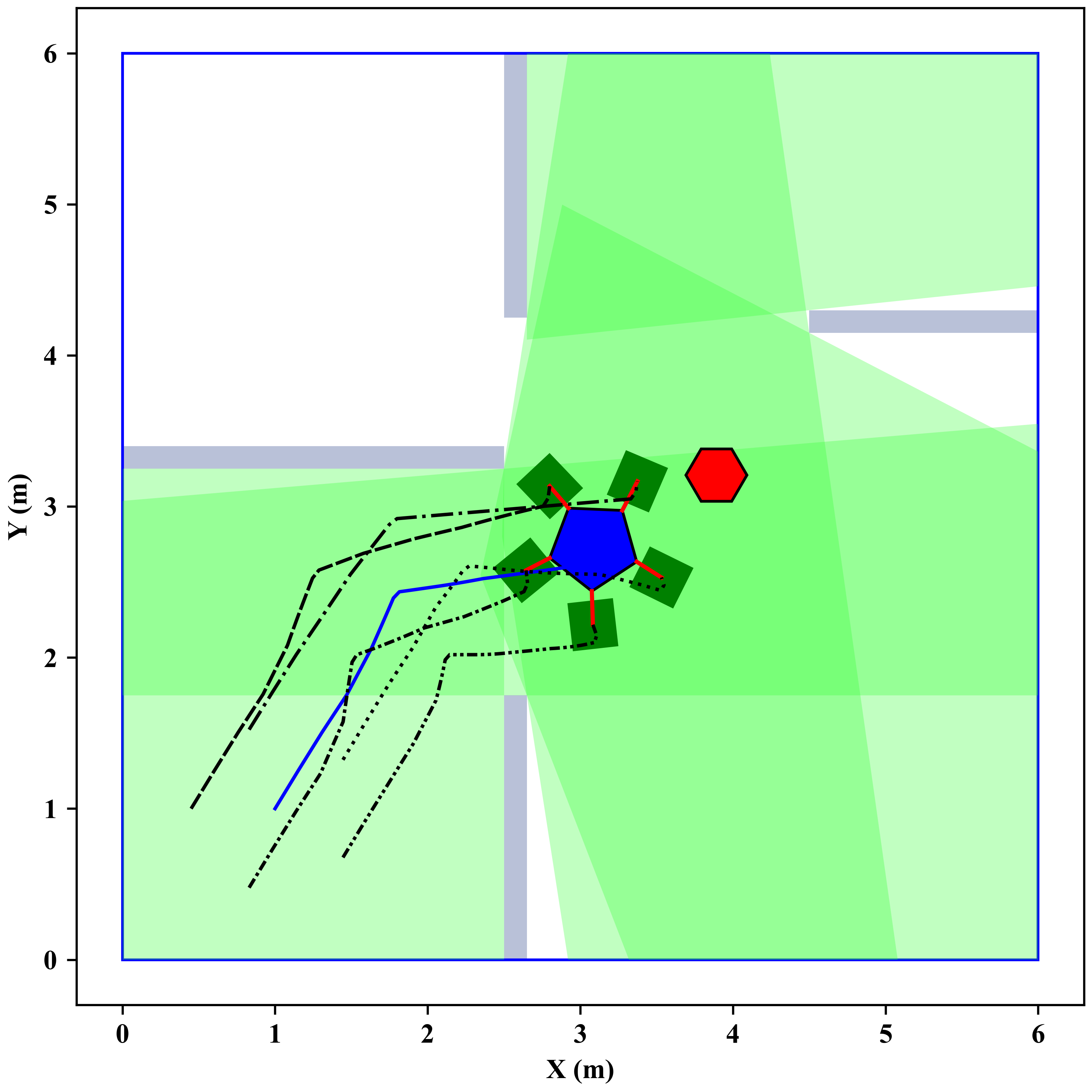}   
    	\subcaption{\label{fig:5c}  $t = 20\ s$}
	\end{subfigure}
	
    \begin{subfigure}{0.24\textwidth}
    	\centering
        \includegraphics[width=115px]{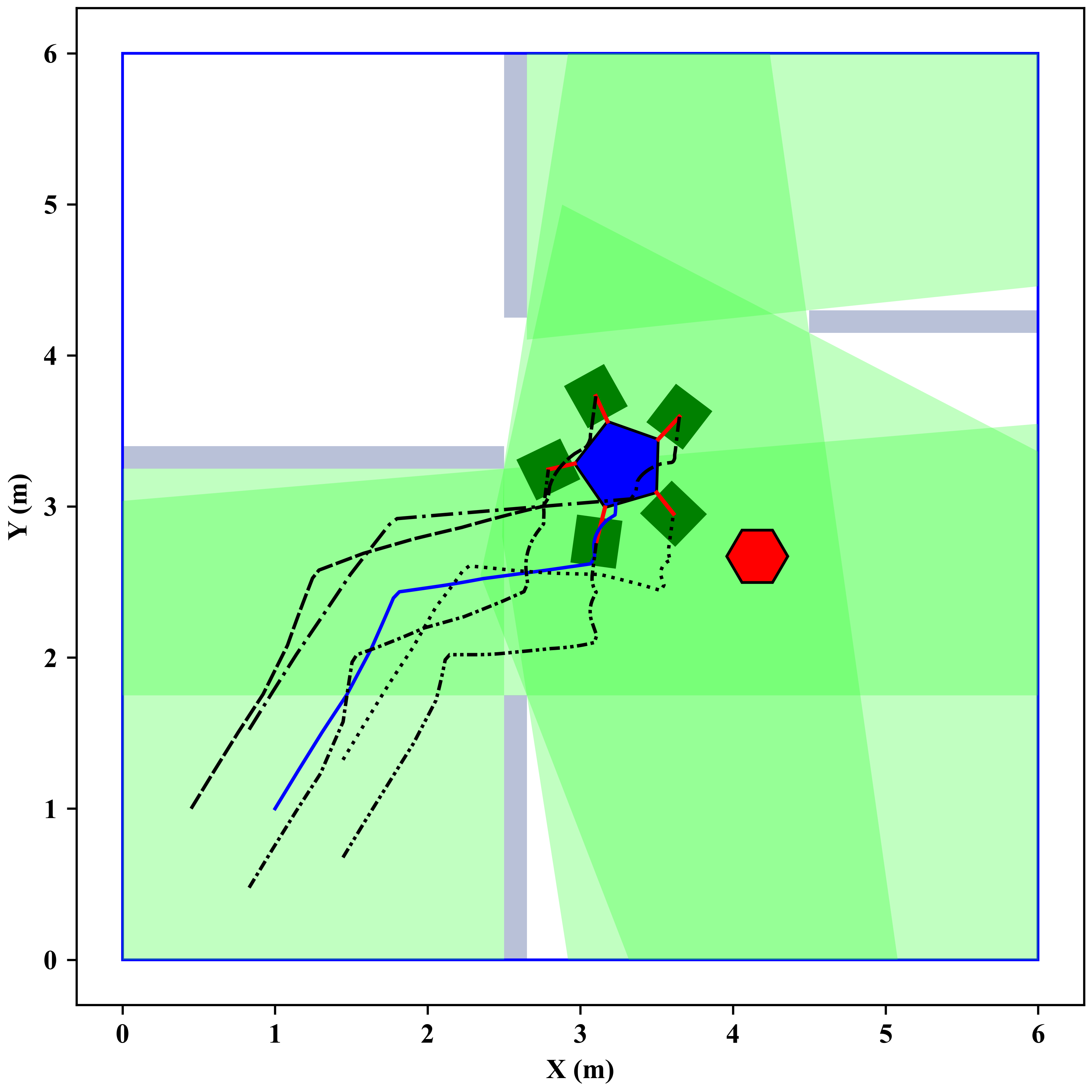}  
    	\subcaption{\label{fig:5f}  $t = 26\ s$}
	\end{subfigure}
    \begin{subfigure}{0.24\textwidth}
    	\centering
        \includegraphics[width=115px]{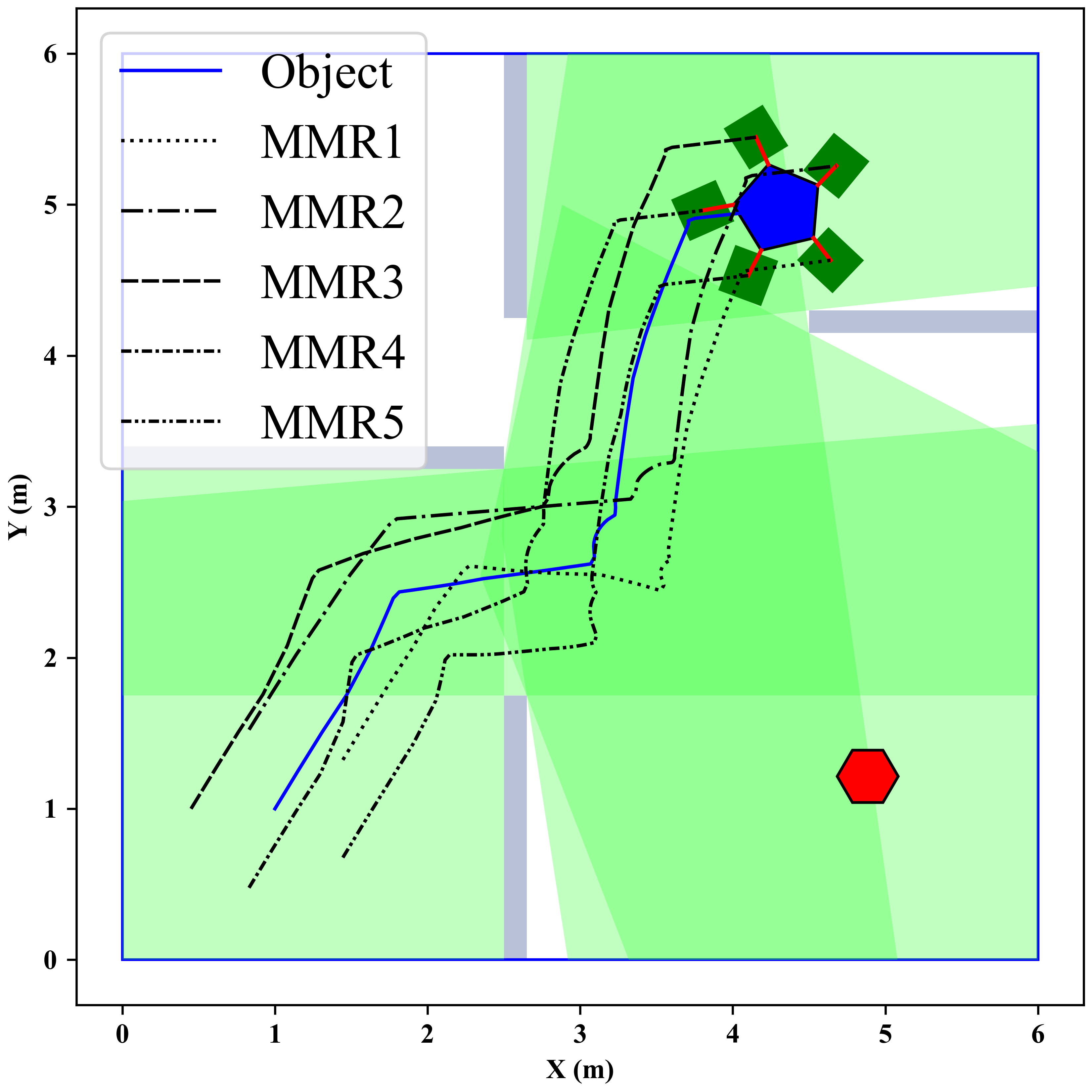}
    	\subcaption{\label{fig:5h}  $t = 42.25\ s$}
	\end{subfigure}
	
	\begin{subfigure}{0.49\textwidth}
		\centering
		\includegraphics[width = 210px]{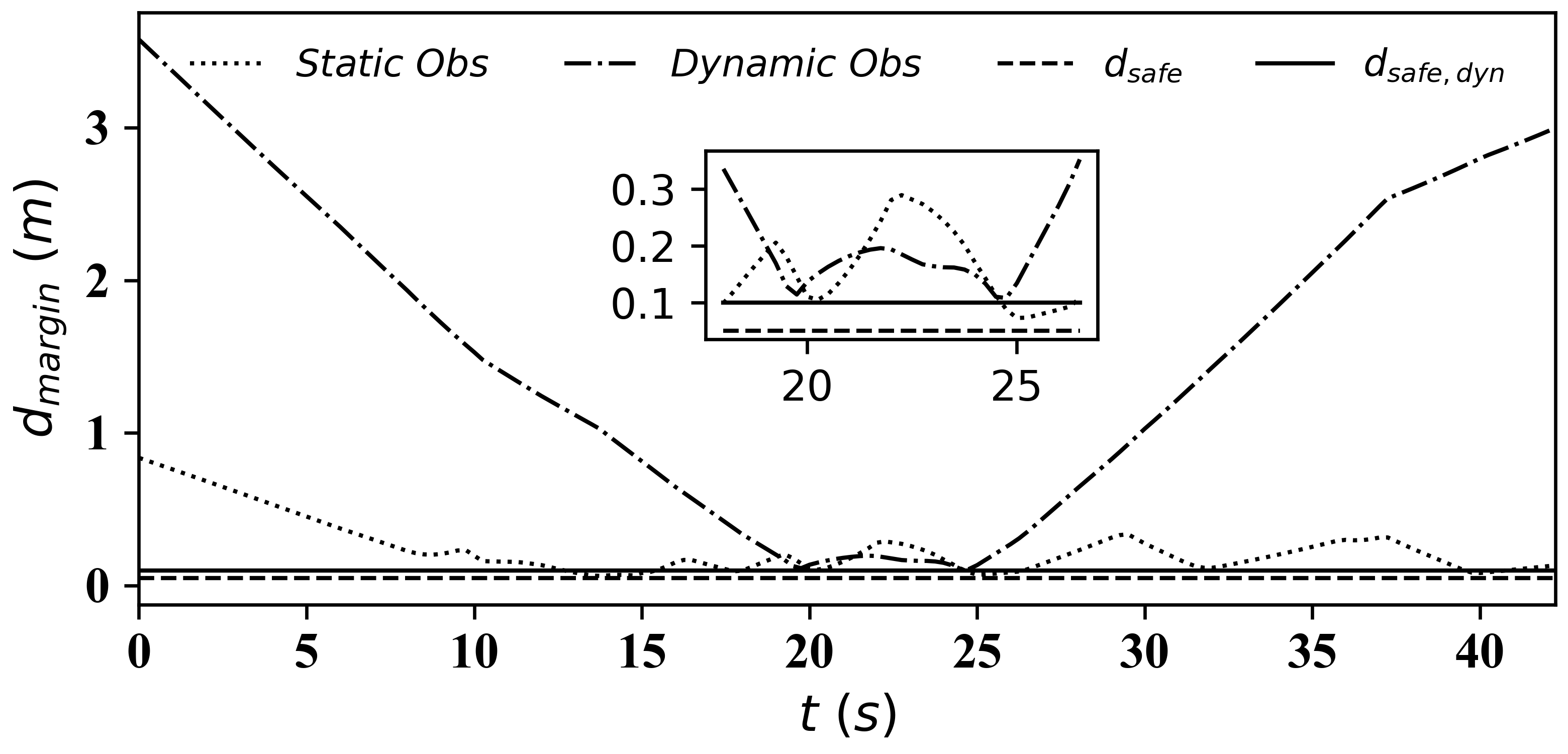}
		\subcaption{\label{fig:safety_margin}}
	\end{subfigure}
	
    \caption{The snapshots of cooperative object transportation using the proposed online motion planner. The object is being transported from the bottom-left to the top-right corner. The red hexagon is a dynamic obstacle in its current state. The green polygons represent the static obstacle free convex region around the path. Fig. \ref{fig:5a} shows the starting of the transportation. Fig. \ref{fig:5c} shows the passing through the door. The dynamic obstacle avoidance maneuver has been captured in Fig. \ref{fig:5c}. Fig. \ref{fig:5h} captures the successful completion of the goal. Fig. \ref{fig:safety_margin} plots the safety margin during object transportation through the narrow passages. The horizontal lines plots safety margins $d_{safe }= 0.05\ m$ and $\ d_{safe,dyn} = 0.1\ m$ is for static and dynamic obstacles.}
    \label{fig:5}
\end{figure}

\subsection{Object Transportation in Simulations} \label{OTTNP}
We employ the proposed motion planning algorithm in three different cases of object transportation,:
\begin{enumerate}
	\item Case 1: through a narrow region.
	\item Case 2: in an environment with dynamic obstacles with curvilinear motion.
	\item Case 3: in an environment with multiple dynamic obstacle avoidance.
\end{enumerate}

In Case 1, we analyse the planning efficacy in narrow regions. Case 2 focuses on the dynamic obstacle state prediction error, and Case 3 test the algorithm for simultaneous multiple dynamic obstacle avoidance.

\subsubsection{Case 1}
The proposed algorithm plans the motion for cooperative object transportation in a warehouse-like environment as illustrated in Fig. \ref{fig:5a}. The environment with a dynamic obstacle represented in red-hexagonal shape with initial position $p_{dyn} = [3, 5]^T\ m$ and velocity $v_{dyn} = [0.045, 0.090]^T\ m/s$. The cooperative MMRs need to transport the object from the bottom left corner to the top right corner of the environment, navigating through the narrow doorways with opening $1.5\ m$ and $1.85\ m$.

The offline path planner, described in Section \ref{Global Path Planning}, generates intermediate feasible formations as shown in Fig. \ref{gor:fig} and computes the shortest path from the start to the goal through these intermediate points. The online local motion planner in Section \ref{MotionPlanning} generates a near-optimal collision-free trajectory in a receding horizon manner.

Fig. \ref{fig:5} includes the snapshots of the object transportation process. The green convex polygons in Fig. \ref{fig:5} contain the obstacle-free shortest distance path (S). The collaborative MMRs start moving from their start location (Fig. \ref{fig:5a}) to the first door, adhering close to the reference trajectory. Just after navigating through the first door (Fig. \ref{fig:5c}), the MMRs encounter a dynamic obstacle (Fig. \ref{fig:5c}) and traverse safely without collision (Fig. \ref{fig:5f}) to the goal (Fig. {\ref{fig:5h}}). Fig. \ref{fig:safety_margin} plots the distance margin $d_{margin}$, i.e., the minimum distance between the formation and the static and dynamic obstacles. The minimum safety margin for the static and dynamic obstacles value $0.05\ m$ and $0.1\ m$, indicates the proposed algorithm's efficacy in obstacle avoidance. The simulation video is available in the link \url{https://youtu.be/LhE_HcK4g-s}.

\begin{figure}[htbp]
	\begin{subfigure}[t]{0.24\textwidth}
	\centering
	\includegraphics[width=110px]{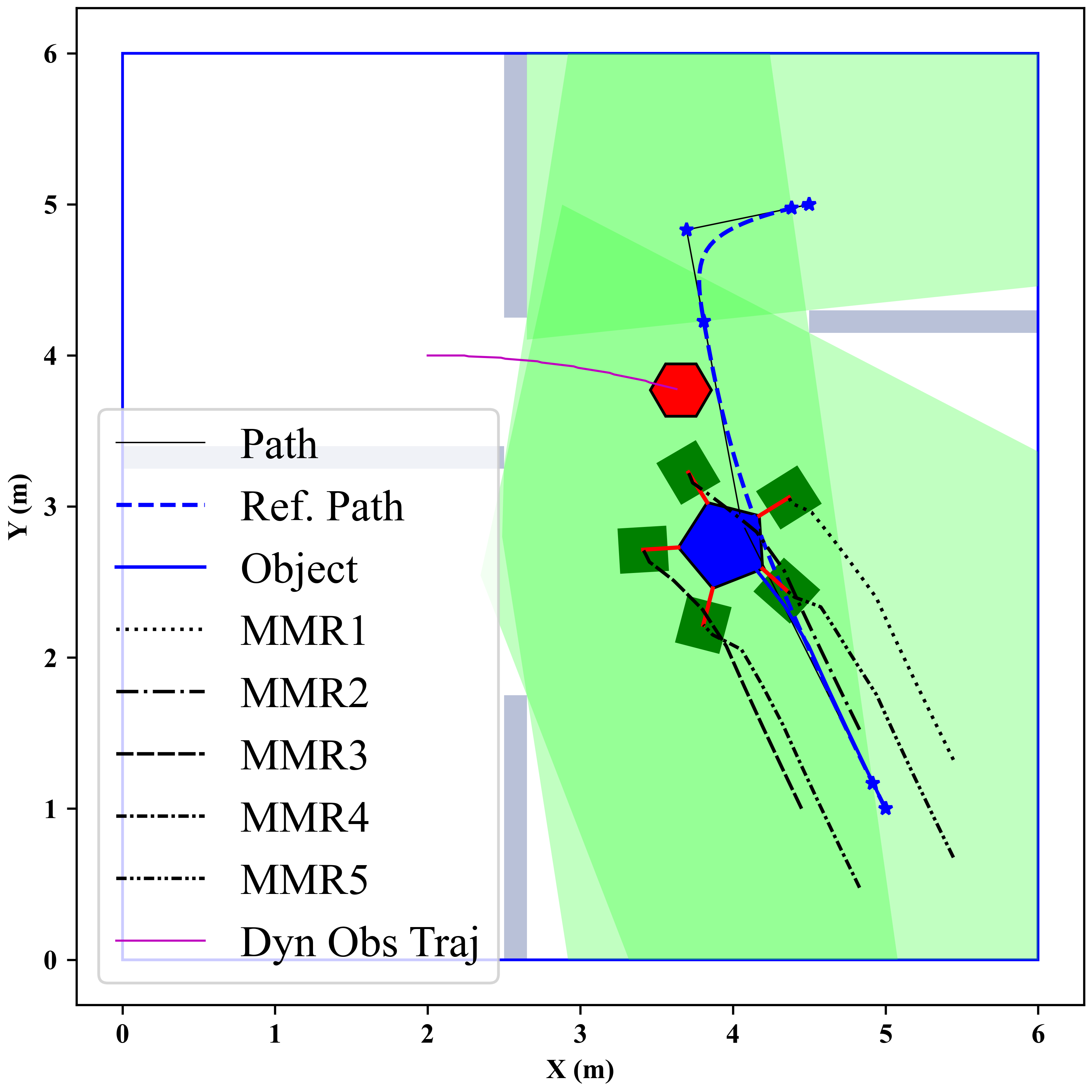}
	\subcaption{\label{fig:DOS1} $t = 14\ s$}	
	\end{subfigure}
	\begin{subfigure}[t]{0.24\textwidth}
	\centering
	\includegraphics[width=110px]{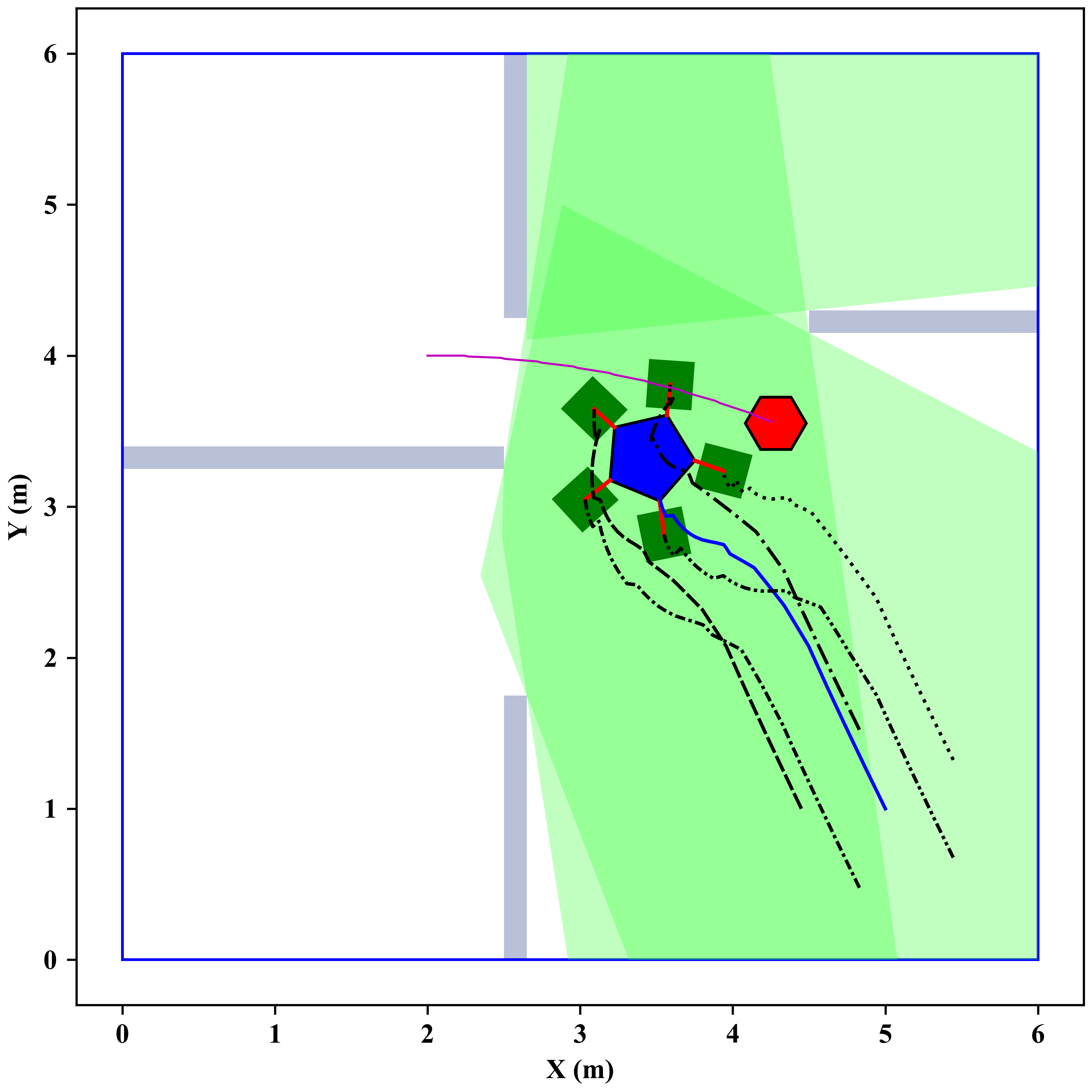}
	\subcaption{\label{fig:DOS2} $t = 19.5\ s$}
	\end{subfigure}
	\begin{subfigure}[t]{0.48\textwidth}
	\centering
	\includegraphics[width = 200px]{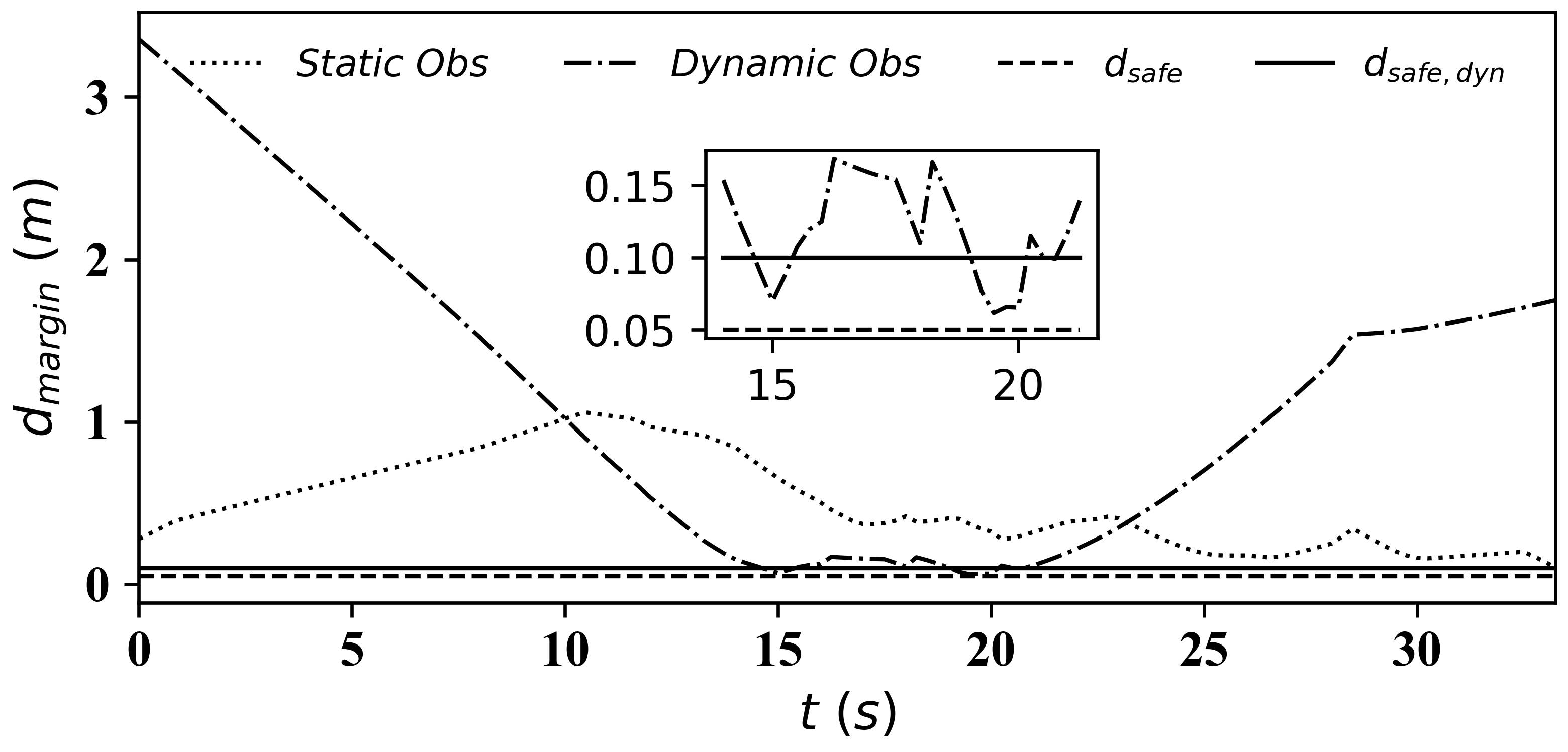}
	\subcaption{\label{fig:DOSSM} Safety Margin}

	\end{subfigure}
	\caption{Object transportation in an environment with dynamic obstacles with a curvilinear trajectory. Fig. \ref{fig:DOS1} shows the snapshot when the dynamic obstacle is close to the formation. Fig. \ref{fig:DOS2} indicates a successful avoidance maneuver. Fig. \ref{fig:DOSSM} shows the safety margin for obstacle avoidance.}
	\label{fig:DynamicObsSpecial}
\end{figure}

\subsubsection{Case 2} We evaluate the proposed motion planner's capability to handle dynamic obstacle state prediction errors using curvilinear motion. The positions of dynamic obstacle have been predicted with a constant velocity model updated at the beginning of each horizon. The dynamic obstacle starts at an initial position $p_{dyn} = [2, 4]^T\ m$ and moves with a velocity of $v_{dyn} = [0.12\cos(0.02t), -0.12\sin(0.02t) ]^T\ m/s$. The cooperative MMRs encounter the dynamic obstacle (Fig. \ref{fig:DOS1}) and start avoidance maneuver. Fig. \ref{fig:DOS2} captures at $t=19.5\ s$ when the dynamic obstacle comes closest to the formation. Despite the constant velocity state estimation model the MMRs successfully transport the obstacles to the goal . Fig. \ref{fig:DOSSM} shows the safety distance maintained during the object transportation. The safety margin $0.06\ m$ at $t\ =\ 20\ s$, which is lower than the applied dynamic safety margin ($0.1\ m$). This discrepancy is because of the constant velocity trajectory estimation model. The safety margin $d_{safe,dyn}$ considers the error in dynamic obstacle position prediction. The formation may collide with the dynamic obstacle in the event of a sudden change in its movement.

\subsubsection{Case 3}
\begin{figure}[htbp]
	\begin{subfigure}{0.24\textwidth}
		\centering
		\includegraphics[width=110px]{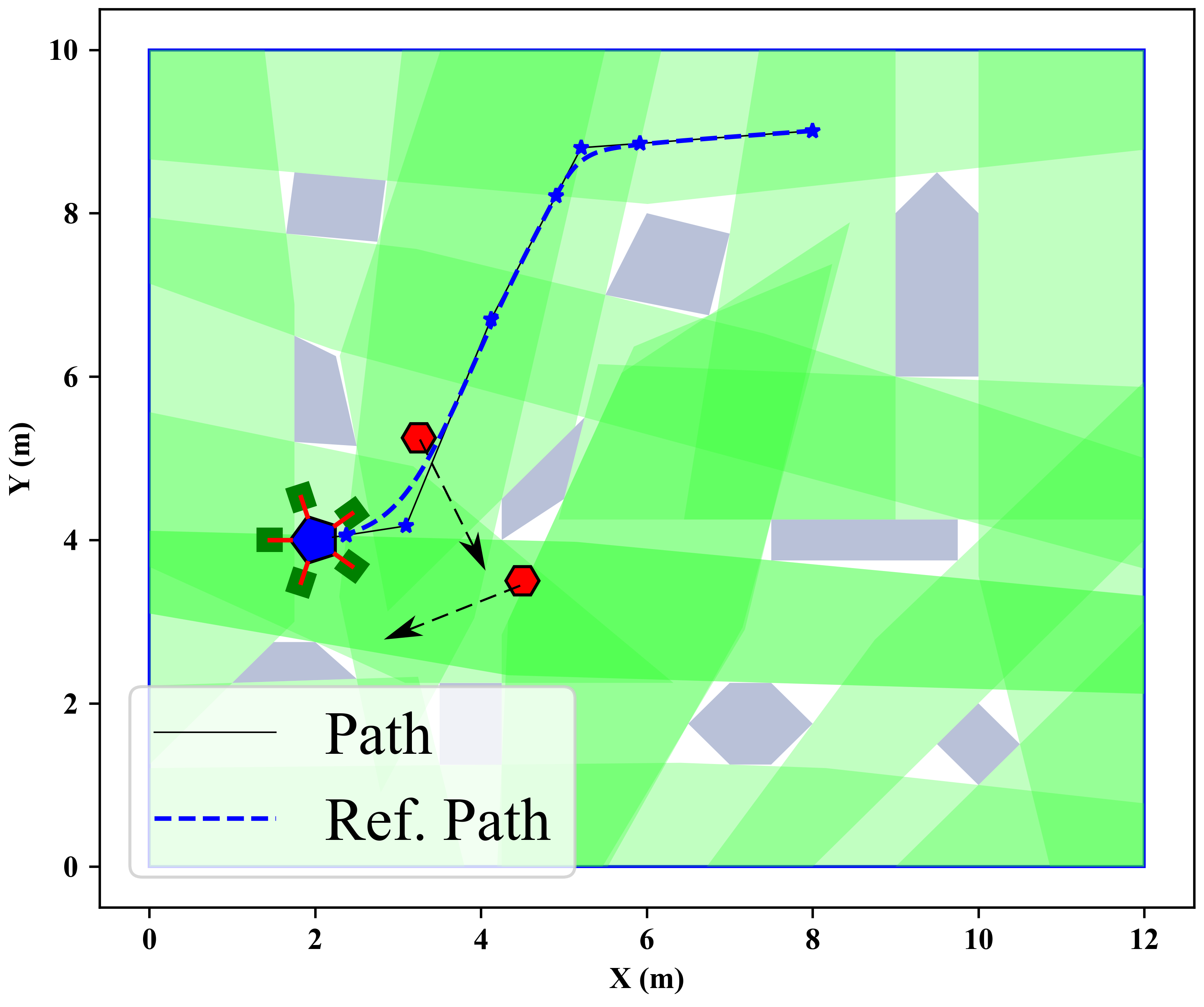} 
		\subcaption{\label{fig:6a} $t = 0\ s$}
	\end{subfigure}
	\begin{subfigure}{0.24\textwidth}
		\centering
		\includegraphics[width=110px]{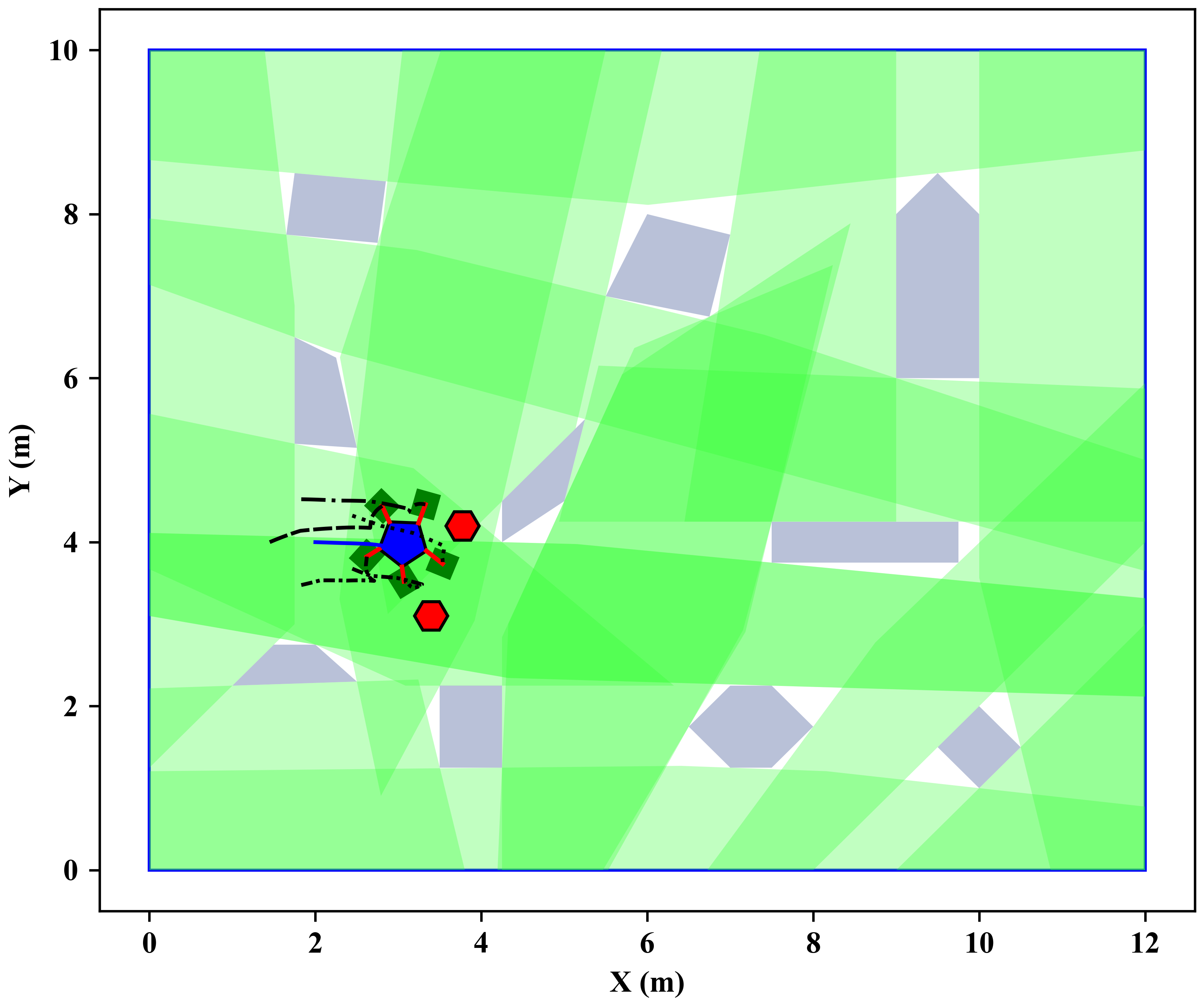}
		\subcaption{\label{fig:6b} $t = 11.75\ s$}
    \end{subfigure}
	\begin{subfigure}{0.24\textwidth}
		\centering
		\includegraphics[width=110px]{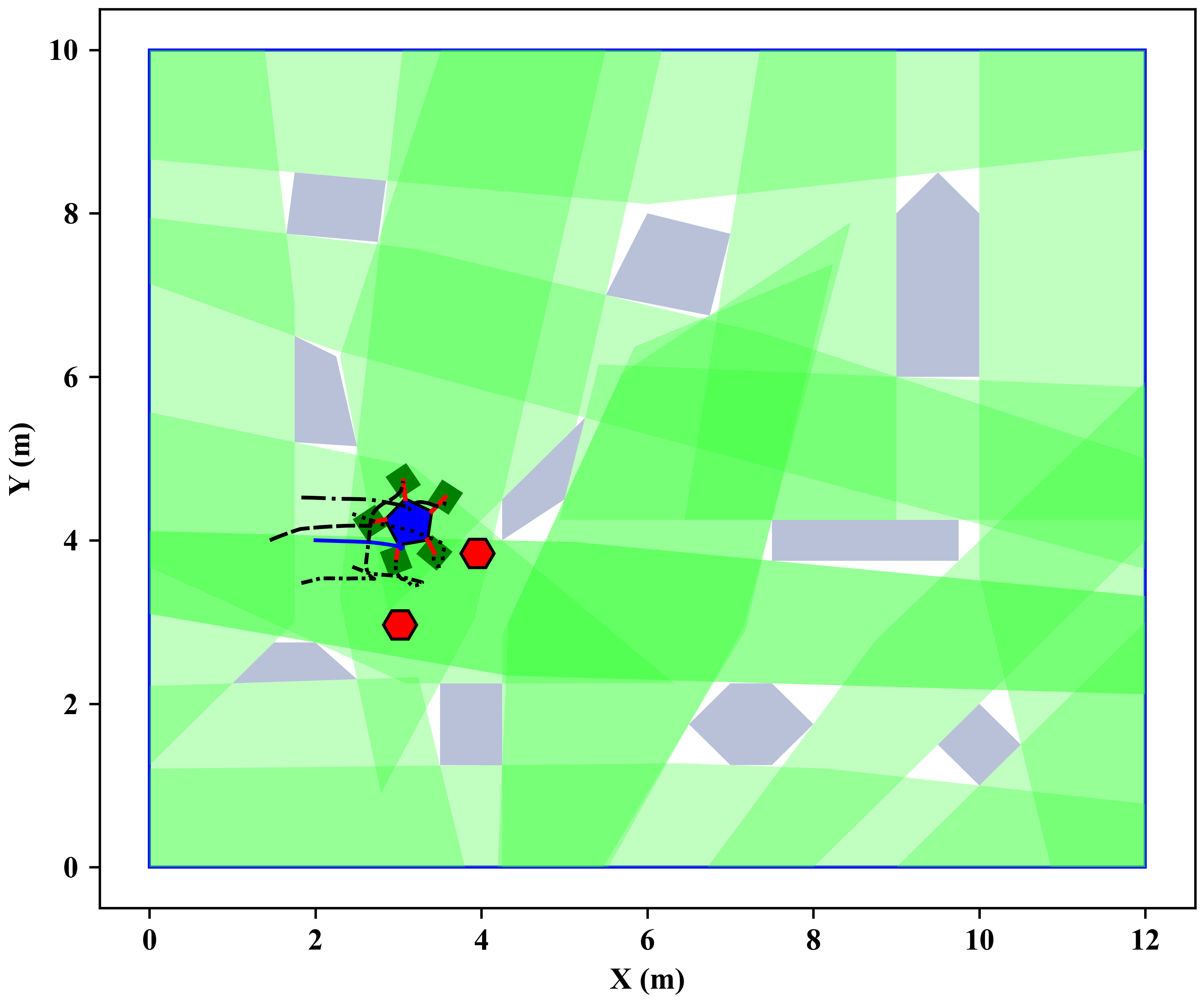}
		\subcaption{\label{fig:6c} $t = 15.75\ s$}
	\end{subfigure}
	\begin{subfigure}{0.24\textwidth}
	    \centering
        \includegraphics[width=110px]{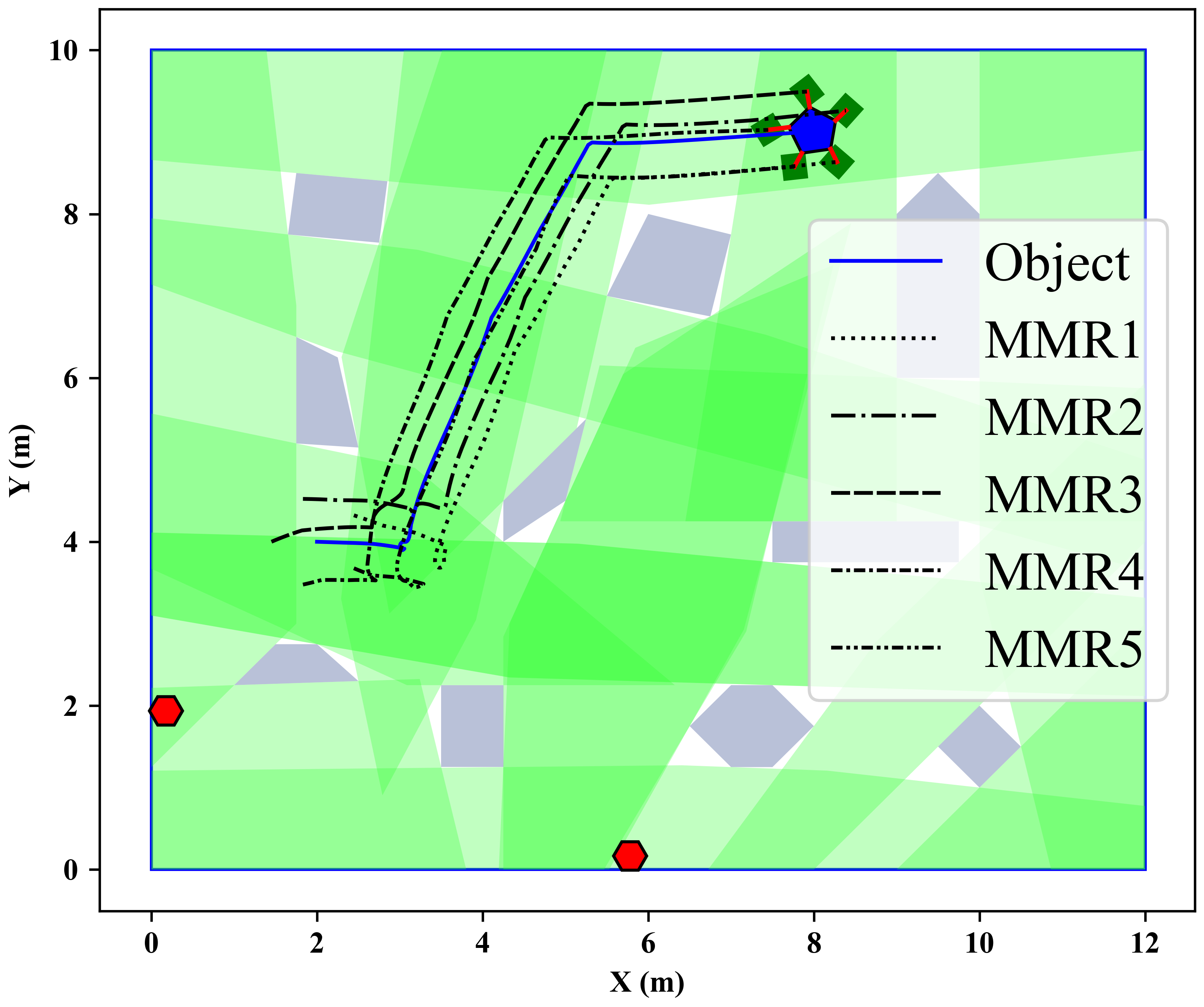}
        \subcaption{\label{fig:6d} $t = 67.75\ s$}
    \end{subfigure}
    
    \begin{subfigure}{0.49\textwidth}
    	\centering
    	\includegraphics[width = 220px]{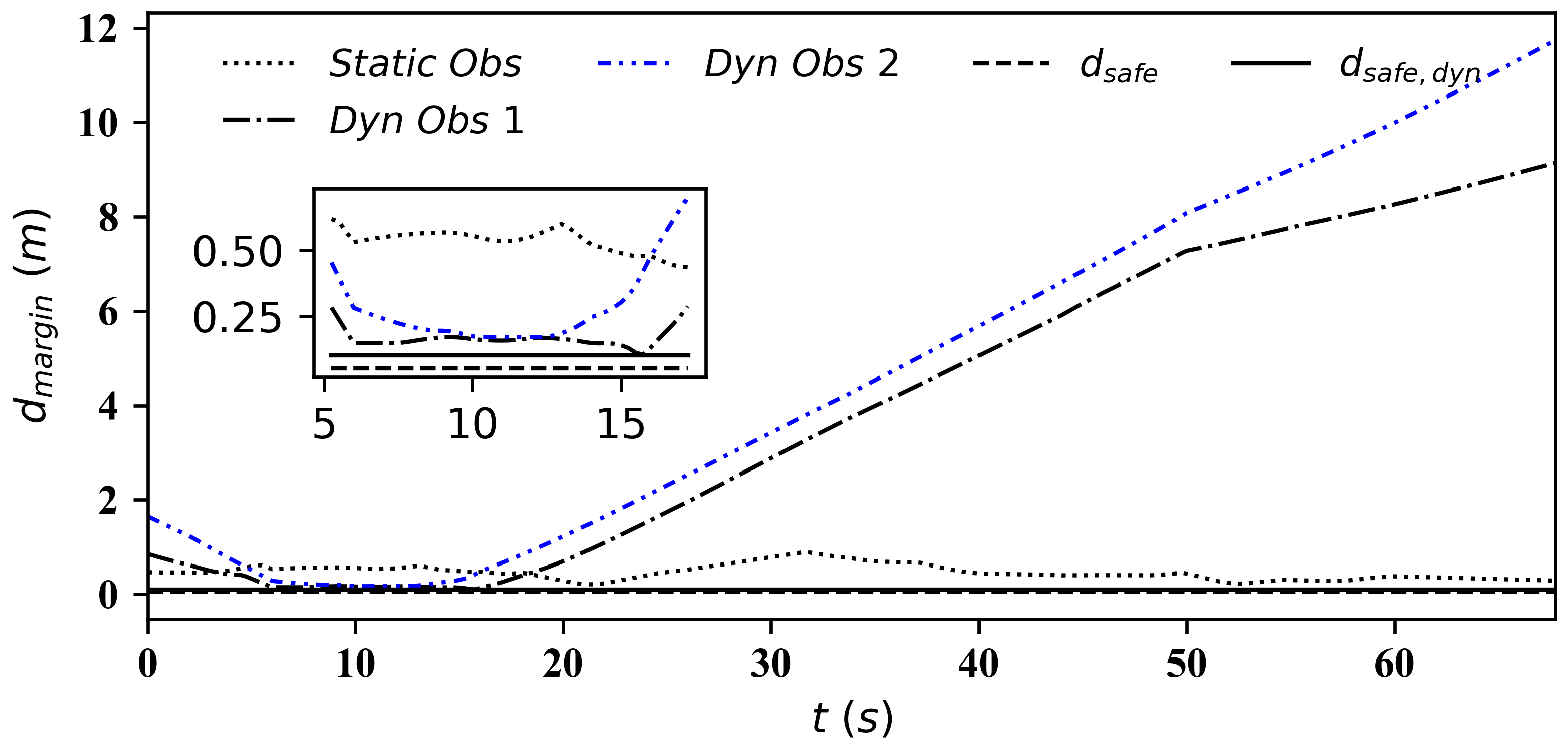}
    	\subcaption{\label{fig:safety_marginDO}}
    \end{subfigure}
    \caption{Dual dynamic obstacle avoidance. Fig. \ref{fig:6a} and \ref{fig:6d} show the start and goal positions. Fig. \ref{fig:6b} and \ref{fig:6c} show the avoidance maneuver while encountering dynamic obstacles one and two. \ref{fig:safety_marginDO} plots the safety margin in simultaneous dual dynamic obstacle avoidance. The horizontal lines plots safety margins $d_{safe }= 0.05\ m$ and $\ d_{safe,dyn} = 0.1\ m$ is for static and dynamic obstacles.}
    \label{fig:6}
\end{figure}

To evaluate the proposed algorithm for a cluttered environment, we have employed the algorithm for motion planning in an environment in Fig. \ref{fig:6a} with two dynamic obstacles with initial positions $p_{dyn,1} = [3.25, 5.25]^T\ m,\ p_{dyn,2} = [4.5, 3.5]^T\ m$ and velocities $v_{dyn,1} = [0.045, -0.090]^T\ m/s,\ v_{dyn,2} = [-0.094, -0.034]^T\ m/s$. Fig. \ref{fig:6b} and \ref{fig:6c} show the snapshots of the dynamic obstacle avoidance maneuvers while they are closest to the obstacles. The formation encounters both dynamic obstacles simultaneously. The multi-MMRs successfully transport the object to the goal location (Fig. \ref{fig:6d}) without collision with nearly $7 \%$ state prediction error in magnitude. It demonstrates the motion planner's capability in random dynamic environment scenarios with multiple dynamic obstacles.
Fig. \ref{fig:safety_marginDO} shows that the minimum distance from the MMRs to the static and dynamic obstacles remains positive, indicating no collision.

\subsubsection{Computational Time Analysis}

	\begin{table}[h]
		\centering
		\caption{Computation Time (in second) for three planning cases}
		\begin{tabular}{@{}llllll@{}}
			\hline
			& Min& Mean & Max & SD\\
			\hline
			Case 1 &$0.71$&$1.27$ & $1.55$& $0.21$ \\
			\hline
			Case 2 &$0.73$&$1.31$ & $1.80$& $0.30$ \\
			\hline
			Case 3 &$0.61$&$1.11$ & $1.36$& $0.15$ \\
			\hline
		\end{tabular}
		\label{tab:CompTime}
	\end{table}

We present the computation time in Python implementation for a single online planning horizon of $T_h = 6\ s$ with the execution horizon $T_e = 2\ s$ for three different cases. The motion planning algorithm was run on a Laptop equipped with AMD Rayzen 5800H CPU and 16 GB RAM with Windows 11 OS. The maximum value of computation time of all three cases is $1.8\ s$, which is well below  $T_e$ ($1.8\leq2$), demonstrate the real-time applicability. However, the computation time may increase for planning in narrow regions or due to dynamic obstacle state prediction.

\subsection{Hardware Validation with two MMRs}
\begin{figure}[htbp]
\centerline{\includegraphics[width = 220px]{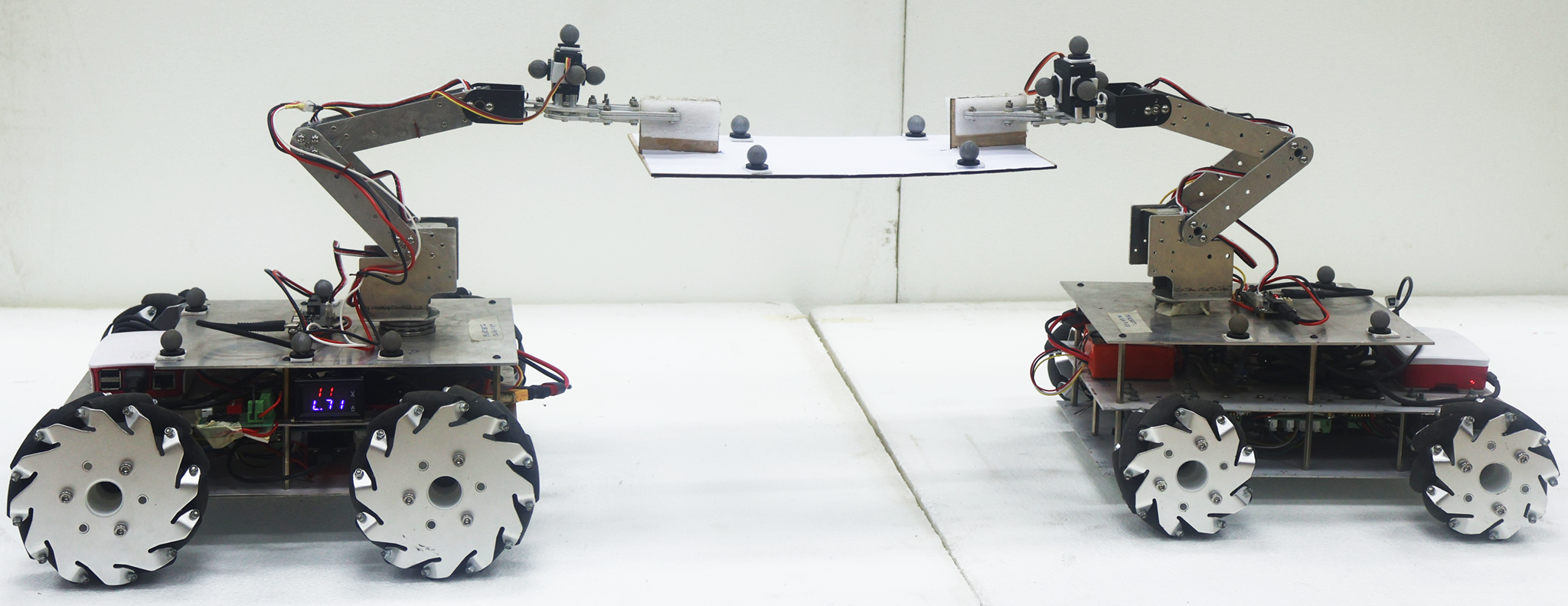}}
\caption{Experimental Setup of two MMRs. The left and right MMR's wheel diameters are $0.127m$ and $0.100m$, respectively. The height of the top plate, where the manipulator base has been mounted, is adjusted to the same.}
\label{fig:HoloSetup}
\end{figure}

\begin{figure}[H]
	\begin{subfigure}[t]{0.24\textwidth}
		\includegraphics[width=110px]{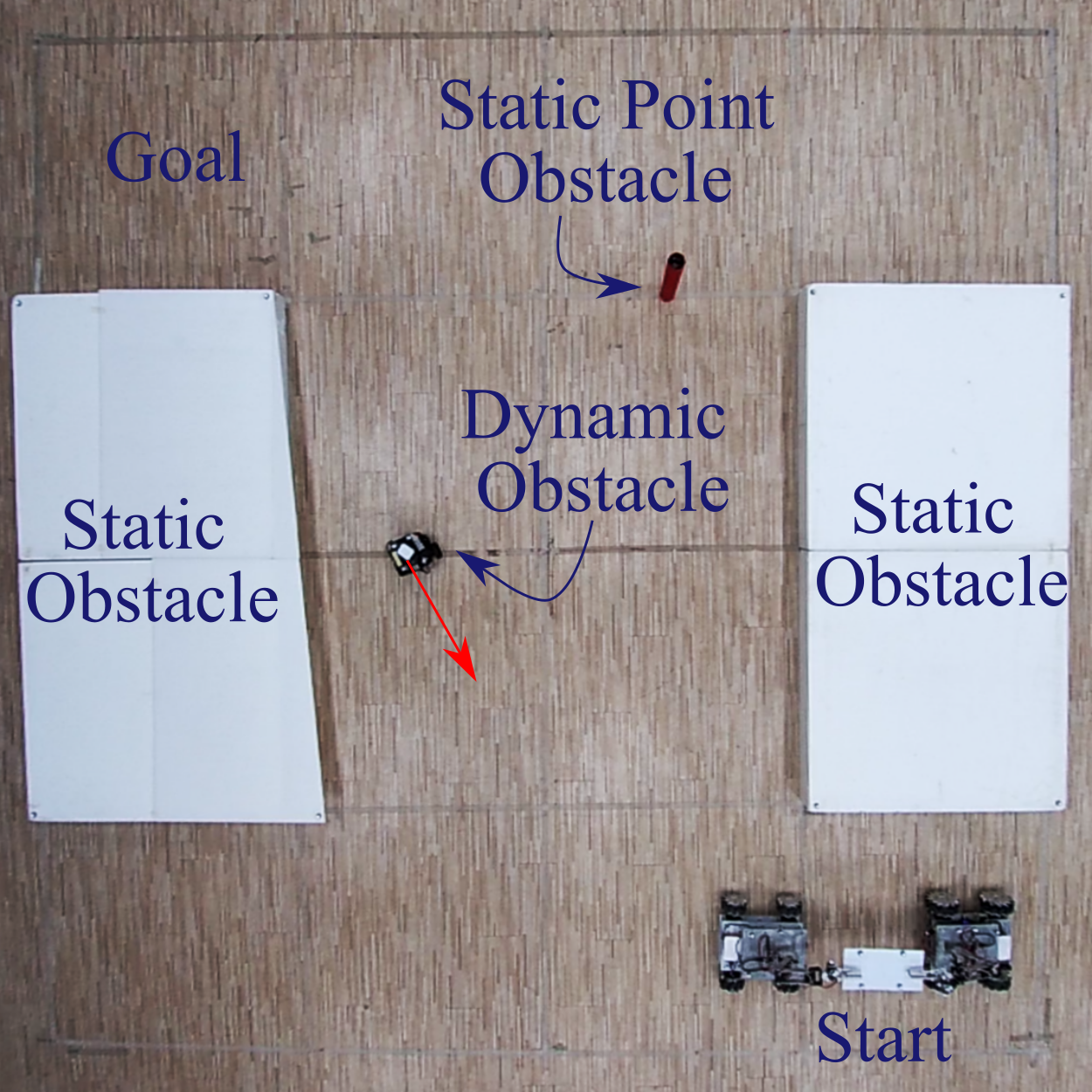}	
		\subcaption{\label{fig:expa} $t = 0\ s$}
	\end{subfigure}
	\begin{subfigure}[t]{0.24\textwidth}
		\includegraphics[width=110px]{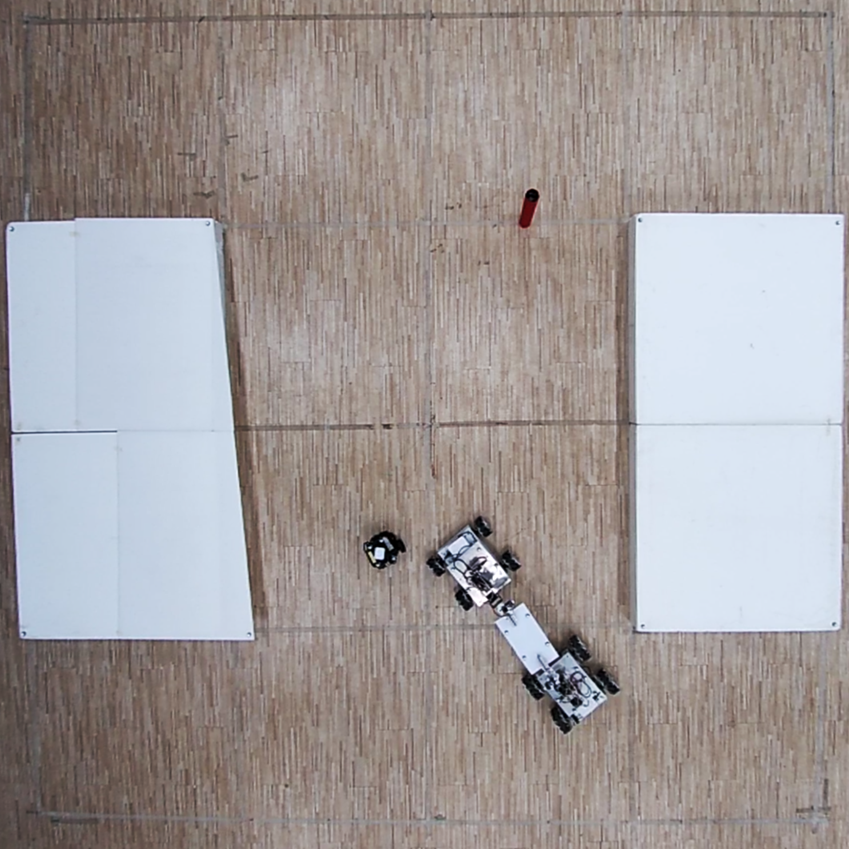}
		\subcaption{\label{fig:expc} $t = 6.10\ s$}
	\end{subfigure}
	\begin{subfigure}[t]{0.24\textwidth}
		\includegraphics[width=110px]{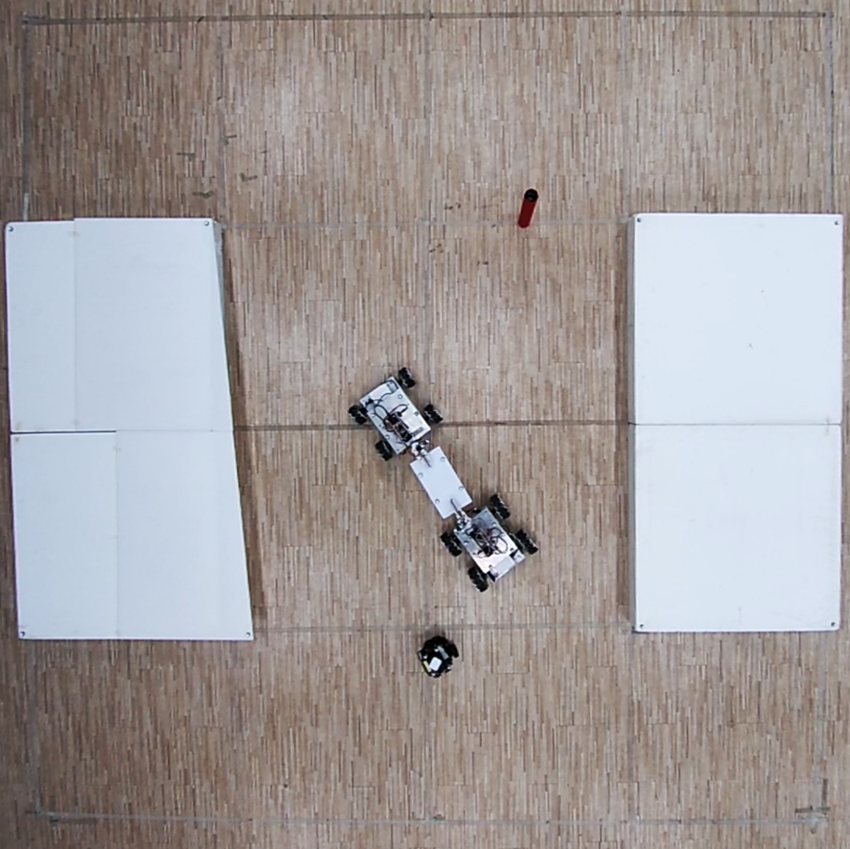}
		\subcaption{\label{fig:expd} $t = 12.10\ s$}
	\end{subfigure}
	\begin{subfigure}[t]{0.24\textwidth}
		\includegraphics[width=110px]{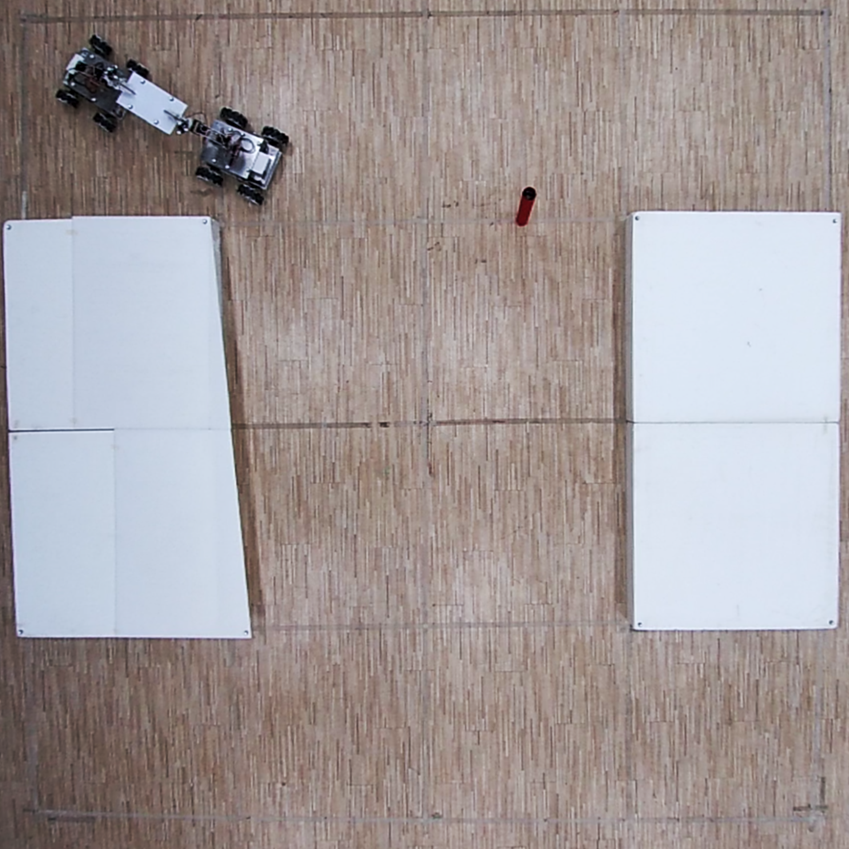}
		\subcaption{\label{fig:expe} $t = 30\ s$}
	\end{subfigure}
	\caption{Two MMR transporting a rectangular plate-like object in a dynamic environment Fig. \ref{fig:expa}. The MMRs encounter a dynamic obstacle and start avoidance maneuver (Fig. \ref{fig:expc}). It successfully avoids the dynamic obstacle \ref{fig:expd} and reaches the goal point \ref{fig:expe}}
	\label{fig:ExpSnap}
\end{figure}

We conducted hardware experiments using our in-house developed ROS-enabled MMRs have been performed in an environment measuring $4\ m\times4\ m$ with static and dynamic obstacles. These experiments assesed the motion planning algorithm in Section \ref{MotionPlanning}. Each holonomic mobile base is equiped four mecanum wheels, each driven by a geared motor with an encoder. Fig. \ref{fig:HoloSetup} shows two holonomic MMRs with the grasped object (Plywood) transporting the object in an indoor environment shown in Fig. \ref{fig:expa}.

Our developed manipulator is the same as the one used in the simulation described in Table \ref{Tab:0} except for joint 5. We fixed the joint five because of the hardware limitations. The DH parameter of the gripper has been updated as $d = 0.120\ m, a = 0, \alpha = 0$, and $\theta = 0$ after fixing joint 5.

\begin{figure*}[p]
	\begin{subfigure}{\textwidth}
	\centering
	\includegraphics[width = 475px]{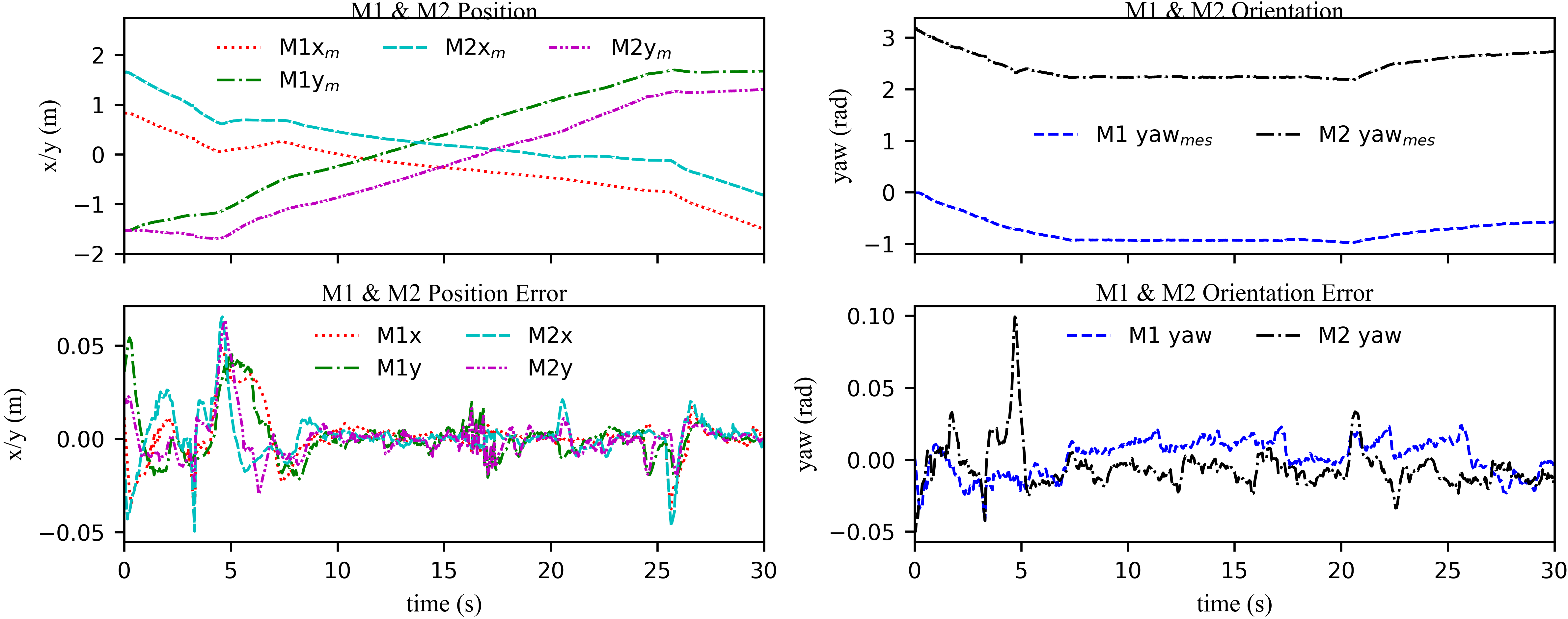}
	\subcaption{\label{fig:MMRBasePose} MMRs' base position and orientation (yaw) tracking performance. M1 is for MMR 1 and M2 is for MMR 2}
	\end{subfigure}
	\begin{subfigure}{\textwidth}
	\centering
	\includegraphics[width = 475px]{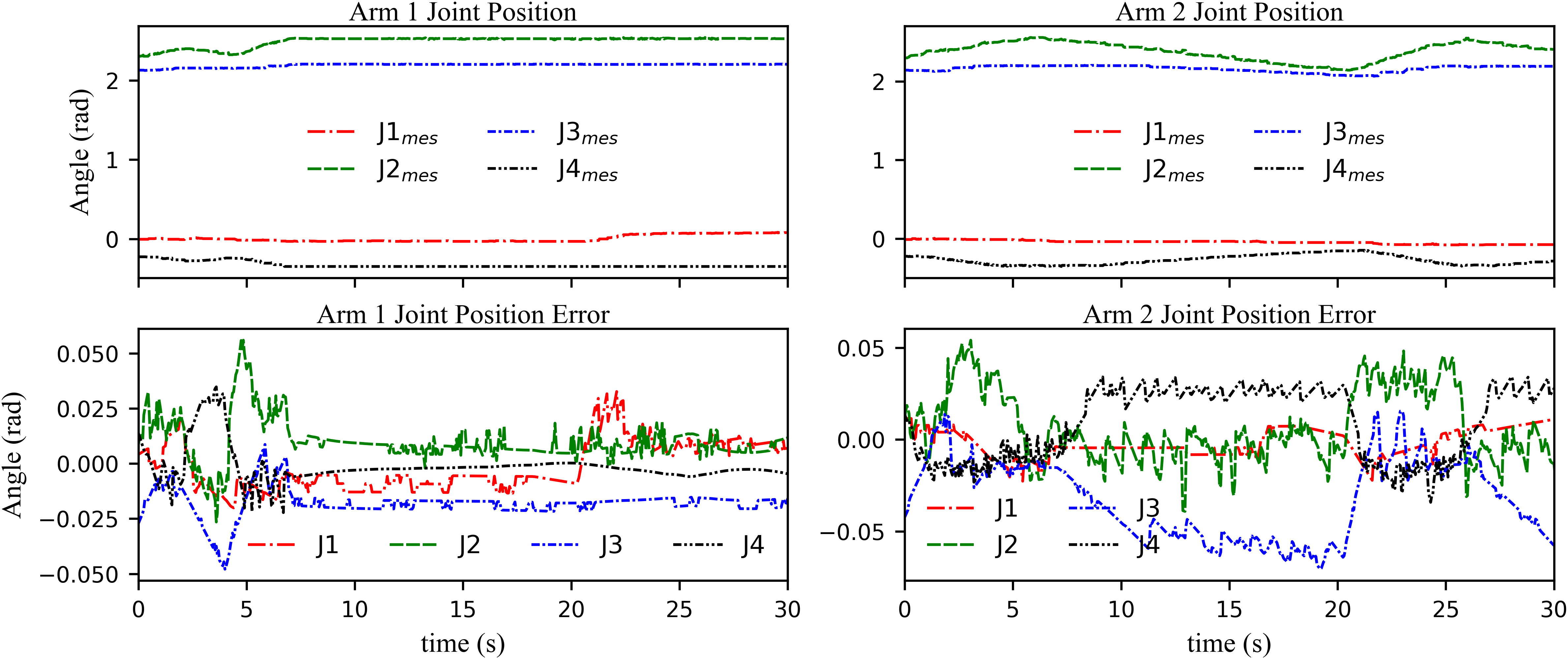}
	\subcaption{\label{fig:ArmJointState} Manipulator arm joint states tracking performance}
	\end{subfigure}
	\begin{subfigure}{\textwidth}
	\centering
	\includegraphics[width = 475px]{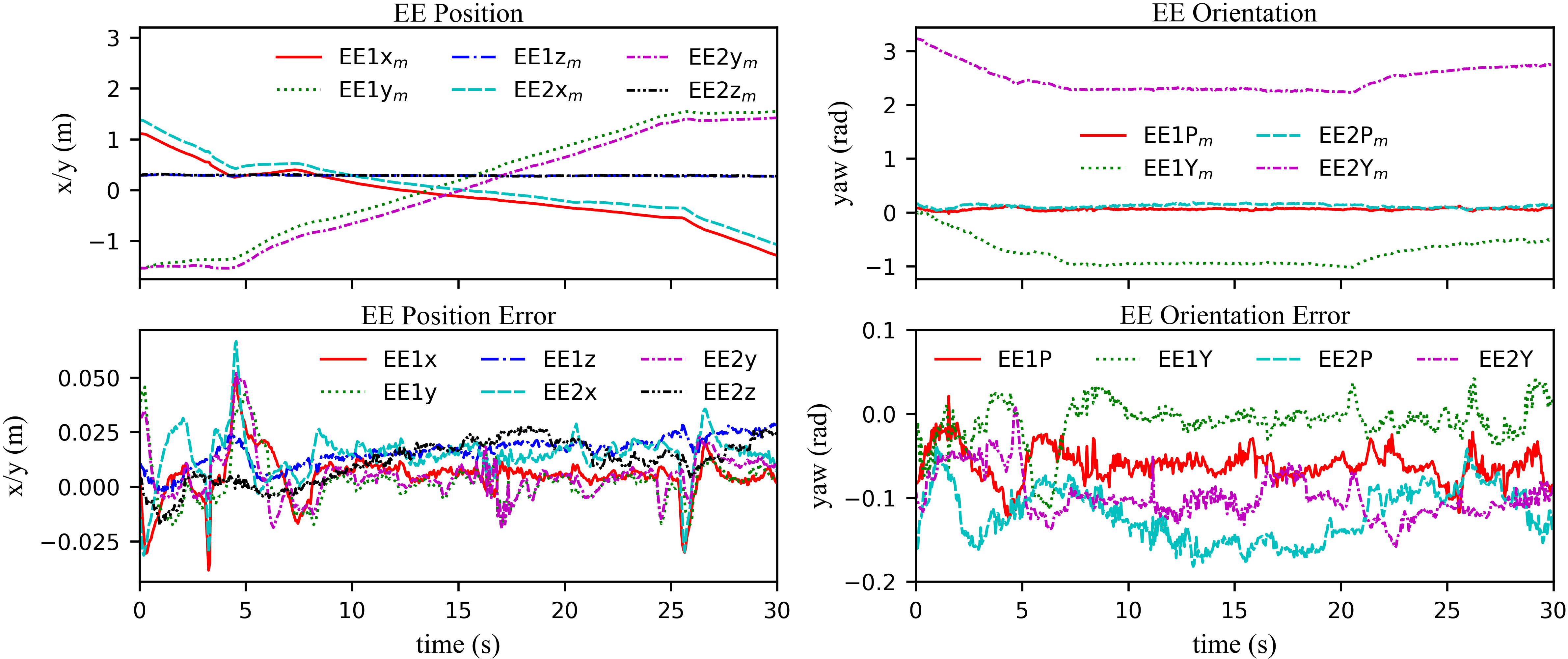}
	\subcaption{\label{fig:EEPose} EE position and orientation tracking performance in task space}
	\end{subfigure}
	\caption{The performances of the MMRs during collaborative object transportation experiment}
\end{figure*}

The trajectory and the control input generated by the online motion planner (Section \ref{OMP}) for the mobile base and the manipulator arm is transmitted to the respective MMRs. The local PID controllers ensure the mobile base's trajectory tracking and the manipulators' joint trajectory tracking. The trajectories of each mobile base and the EE of MMRs are measured using an external motion capture system.
Fig. \ref{fig:EEPose} shows the snap of the object transport from the start (Fig. \ref{fig:expa}) to the goal (Fig. \ref{fig:expe}). During the transportation the system encounters a dynamic obstacle and initiates an avoidance maneuver. Fig. \ref{fig:expc} shows when it comes closest and finally avoids Fig. \ref{fig:expd} the obstacles to reach the goal Fig. \ref{fig:expe}.

\subsubsection{MMR's base trajectory tracking}
A hardware-level PID controller ensures the reference trajectory tracking for the mobile base. Fig. \ref{fig:MMRBasePose} shows the trajectory tracking performance of the two MMRs' base. The position tracking error remains within $\pm0.01\ m$ except at $5\ s$, when it takes a sharp turn and rotation cause an error to increases to $0.06\ m$ due to the wheel traction loss. The orientation tracking performance remains within $\pm\ 0.1\ rad$ except for the traction loss observed at $5\ s$.

\subsubsection{MMR's manipulator arm}

\begin{figure}[htbp]
    \centerline{\includegraphics[width = 200px]{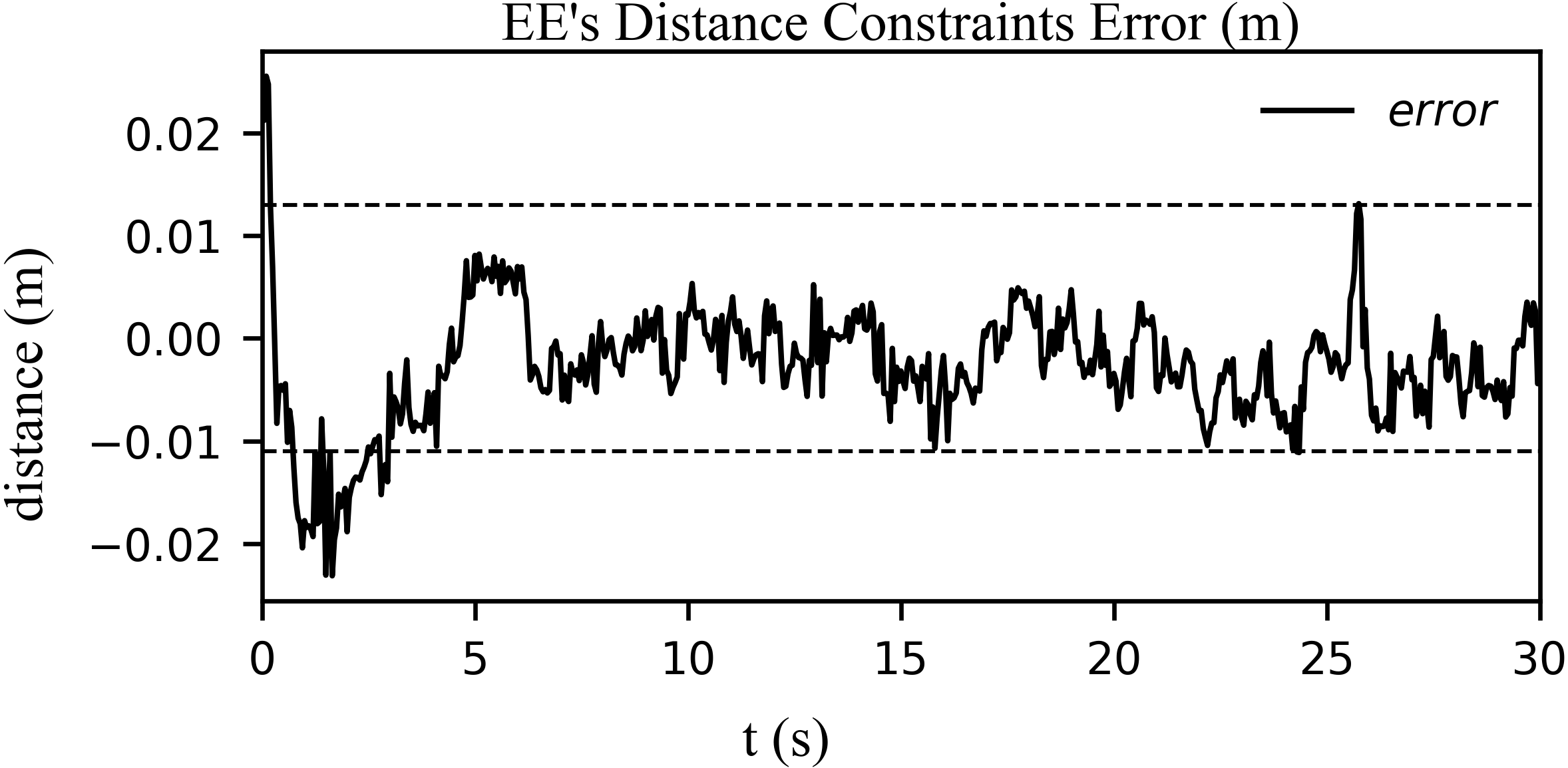}}
    \caption{Distance Constraint of the two MMR's EE.}
    \label{fig:EEConstr}
\end{figure}

The manipulators' arm joint actuators follow the joint trajectories from the online motion planner (Section \ref{OMP}) using hardware-level PID controls. Fig. \ref{fig:ArmJointState} shows the joint trajectory performance of each joint of the two manipulators measured from the encoder of the actuators. The trajectory tracking error remains within $0.07\ rad$. The EEs' trajectory tracking performance in the task space is vital for stable formation maintenance and, more importantly, the grasp constraints (Section \ref{gc}). The task space trajectory of the EE was measured externally using an external motion capture system and shown in Fig. \ref{fig:EEPose}. The position tracking error remains within $\pm\ 0.025\ m$ except at $t=5$ and $26\ s$ when the traction loss of the mobile base occurs. The distance between the two EE'ss grasp point is $0.25\ m$, and the change in the distance has been plotted as an error in Fig. \ref{fig:EEConstr} to verify the grasp constraint. The initial grasp error touches $- 0.02\ m$ and settles within $0.01\ m$ afterward. Other than the initial error, the error remains within $-0.011\ m$ and $+0.013\ m$. The collaborative trajectory tracking task has been accomplished.

The trajectory tracking performance and the stable grasp maintenance show the efficacy of the joint motion planner for the base and the manipulator arm. It is also evident that a successful motion plan and the local individual MMRs' PID controller successfully execute rigid object transportation tasks.

The motion planning algorithm for the hardware experiment was performed on a desktop equipped with Intel Core i7-8700 CPU and 16 GB RAM. The computing time for a planning horizon is as follows.

\begin{table}[h]
	\centering
	\caption{Computation Time (in seconds) of the Experiments for a planning horizon.}
	\begin{tabular}{@{}lcccc@{}}
		\hline
		& Min& Mean & Max & SD\\
		\hline
		Experiments &$0.195$&$0.387$ & $0.751$& $0.161$ \\
		\hline
	\end{tabular}
	\label{tab:ComputationTime}
\end{table}

\subsection{Scalability Analysis}\label{ScalabilityAnalysis}
We analyze the scalability of the MMRs for the computational complexity in this section. The scalability would not be impacted by the offline path planning as the path is computed prior, considering the static obstacles. Hence, the computational time does not impact real-time performance. The online planner's performance dictates the real-time use case and the scalability. The online motion planner's computational complexity is discussed in the following.

For $n$ number of MMRs the motion planning optimization has the following number of constraints and variables which impact the computational complexity.\\

	$(1)$ \textbf{Number of constraints}:
	\begin{itemize}
		\item The number of formation constraints required is $(n - 1)\times 3$; for the 2D formation model, two for position and one for orientation between two robots are required.
		\item Static obstacle avoidance: $ (n + 1)\times n_f$, one set of constraints for each MMR and one for the object is required, and $n_f$ is the number of faces of the convex obstacle-free space.
		\item Dynamic obstacle avoidance: $(2n +1)\times n_{dyn}$ for each MMR, two constraints, one for the mobile base and one for the manipulator, are required per dynamic obstacles. 
		\item State transition constraints: $6n$, Six equality constraints for system dynamics for each MMR.
		\item \textbf{Total no of constraints:} $3(n-1) + n_f(n+1) + n_{dyn}(2n+1) + 6n$
	\end{itemize}
	$(2)$ \textbf{Total number of variables:} The states and control inputs are decision variables of the motion planning optimization. Hence the number of variable is $n\times 2\times 6 = 12n$ \\

	The computation complexity varies linearly as indicated by the number of constraints and variables.\\
	\textbf{Environmental Complexity:} The number or the shape of the static obstacles have very narrow roles in the planning complexity. The number of faces $n_f$ of the static obstacle-free convex polygon linearly time complexity on the motion planning. The complexity of the planning varies linearly with the number of dynamic obstacles $n_{dyn}$ it encounters.
	So, the complexity of planning increases linearly with the number of MMRs. The number of iterations required for solution convergence may increase as the environmental congestion increases with the number of MMRs. Thus, it reduces the solution space for MMRs.

	Our computational time analysis for scalability for offline path and online motion planning is performed in the same environment as Section 5.2.1 Case 1. We performed the motion planning simulations on a laptop with an AMD Rayzen 5800H CPU, 16 GB RAM, and Ubuntu 22.04 OS. The OS was changed as compared to the result presented in Section 5.2.1 Case 1. We notice that running the same analysis in Ubuntu computation time improves by nearly $40\%$.



\begin{table}[h]
	\centering
	\caption{Computation Time (in second) of the offline path planner and the online motion planner for different numbers of MMRs}
	\begin{tabular}{|l|c|c|c|c|c|}
		\hline
		& Path Planning & \multicolumn{4}{c|}{Motion Planning}\\
		\hline
		& &  Min & Mean & Max & SD\\
		\hline
		2 MMRs & $2.25$ & $0.23$ & $0.32$ & $0.36$ & $0.027$\\
		\hline
		3 MMRs & $2.25$ & $0.39$ & $0.47$ & $0.57$ & $0.06$ \\
		\hline
		4 MMRs & $2.18$ & $0.52$ & $0.61$ & $0.68$ & $0.05$\\
		\hline
		5 MMRs & $2.32$ & $0.65$ & $0.83$ & $1.09$ & $0.13$\\
		\hline
	\end{tabular}
	\label{tab:ScaleCompTime}
\end{table}

From the analysis result presented in Table \ref{tab:ScaleCompTime}, it can be concluded that the offline path planner required almost similar time while the motion planner computation time varies almost linearly with the number of robots.

\subsection{Comparative Analysis}\label{ComparativeAnalysis}
We compare our proposed planning framework for dynamic environment with a state-of-the-art algorithm by \textit{AlonsoMora et al.} \cite{2017_AlonsoMora}. We also compare our planning algorithm with the algorithm proposed \textit{Zhang et al.} \cite{2025_zhang}.

\begin{figure}[htbp]
	\begin{subfigure}{0.24\textwidth}
		\centering
		\includegraphics[width=114px]{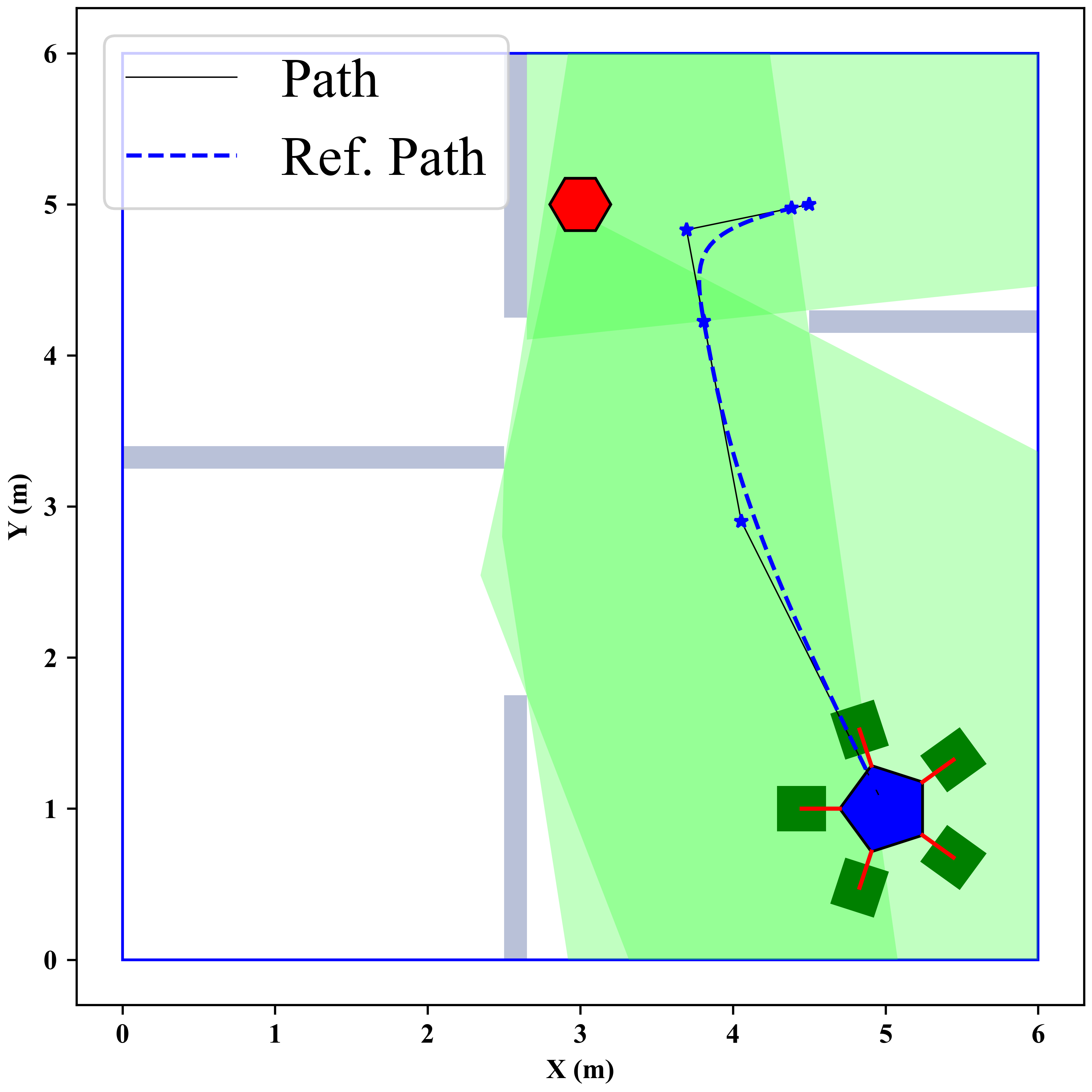}
		\subcaption{\label{fig4a} $t = 0\ s$}
	\end{subfigure}
	\begin{subfigure}{0.24\textwidth}
		\centering
		\includegraphics[width=114px]{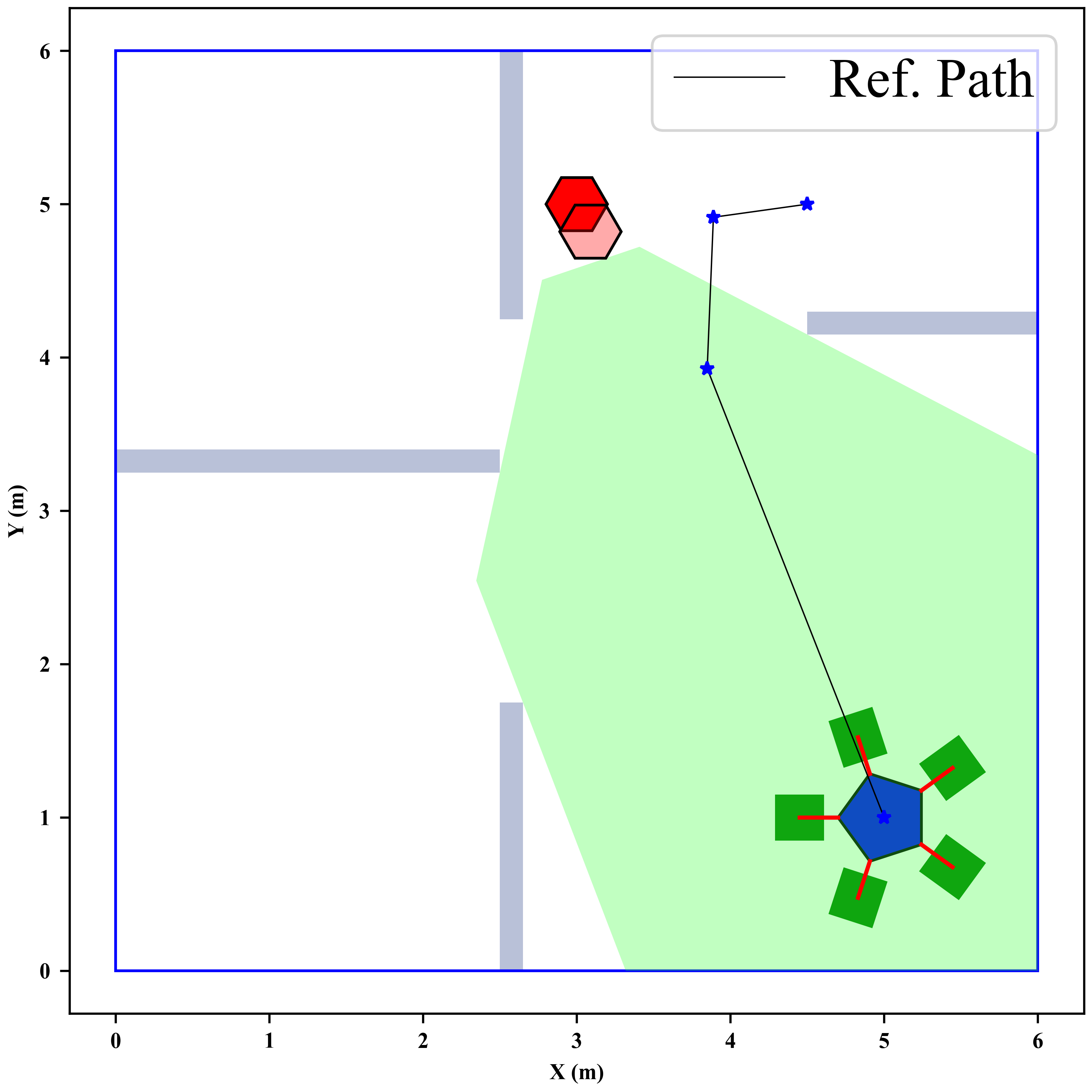}
		\subcaption{\label{fig4e} $t = 0\ s$}
	\end{subfigure}
	\begin{subfigure}{0.24\textwidth}
		\centering
		\includegraphics[width=114px]{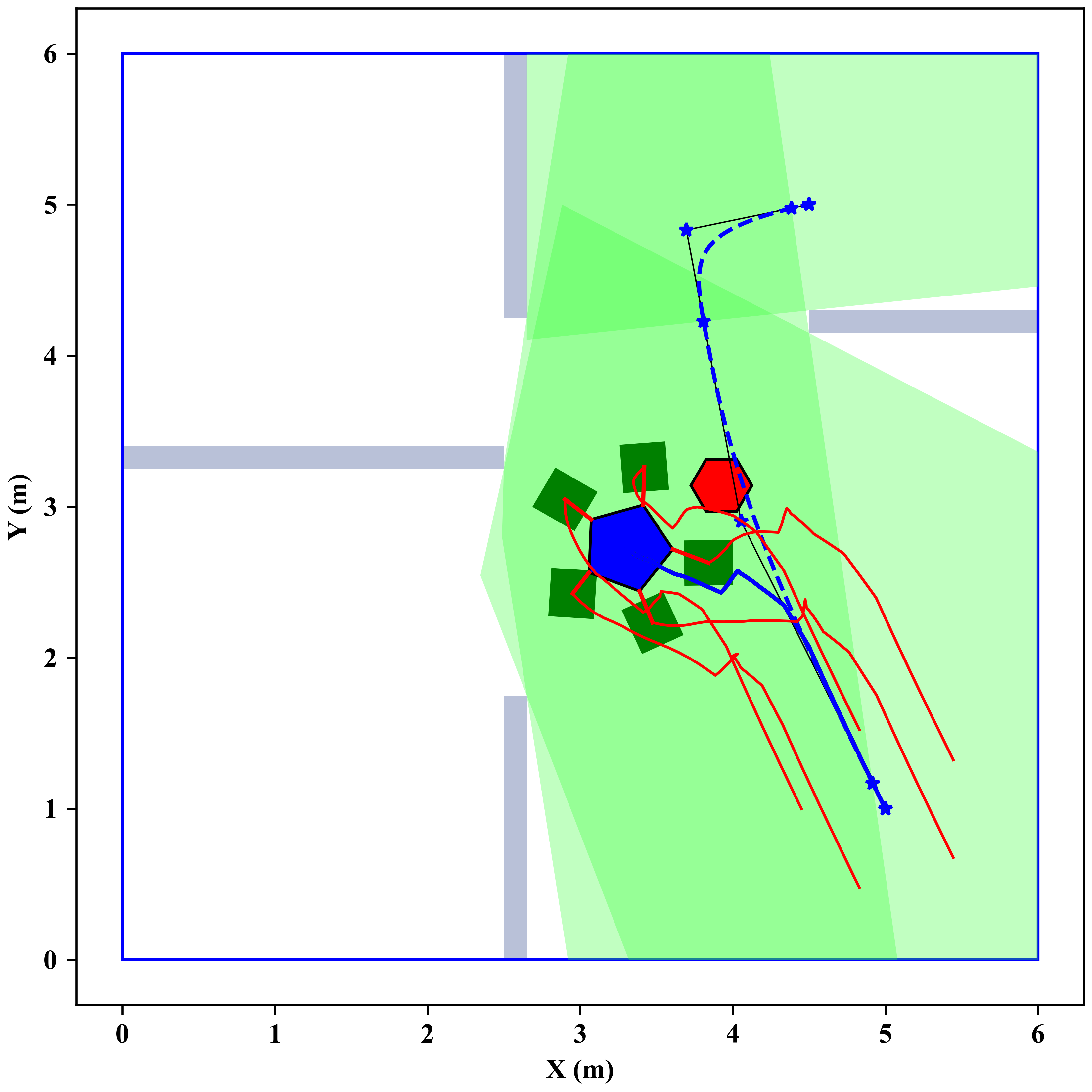}
		\subcaption{\label{fig4b} $t = 20.75\ s$}
	\end{subfigure}
	\begin{subfigure}{0.24\textwidth}
		\centering
		\includegraphics[width=114px]{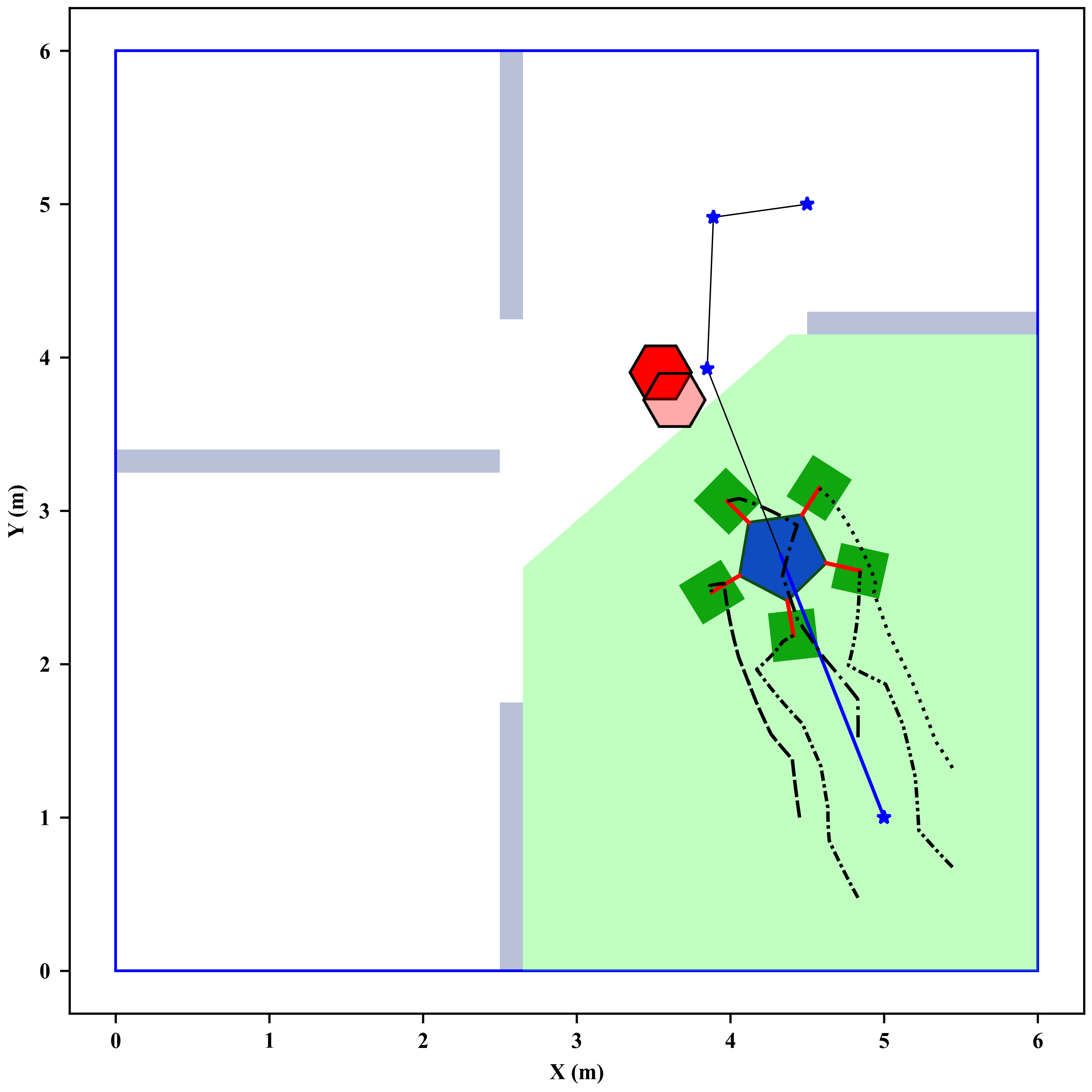}
		\subcaption{\label{fig4f} $t = 12.25\ s$}
	\end{subfigure}
	
	\begin{subfigure}{0.24\textwidth}
		\centering
		\includegraphics[width=114px]{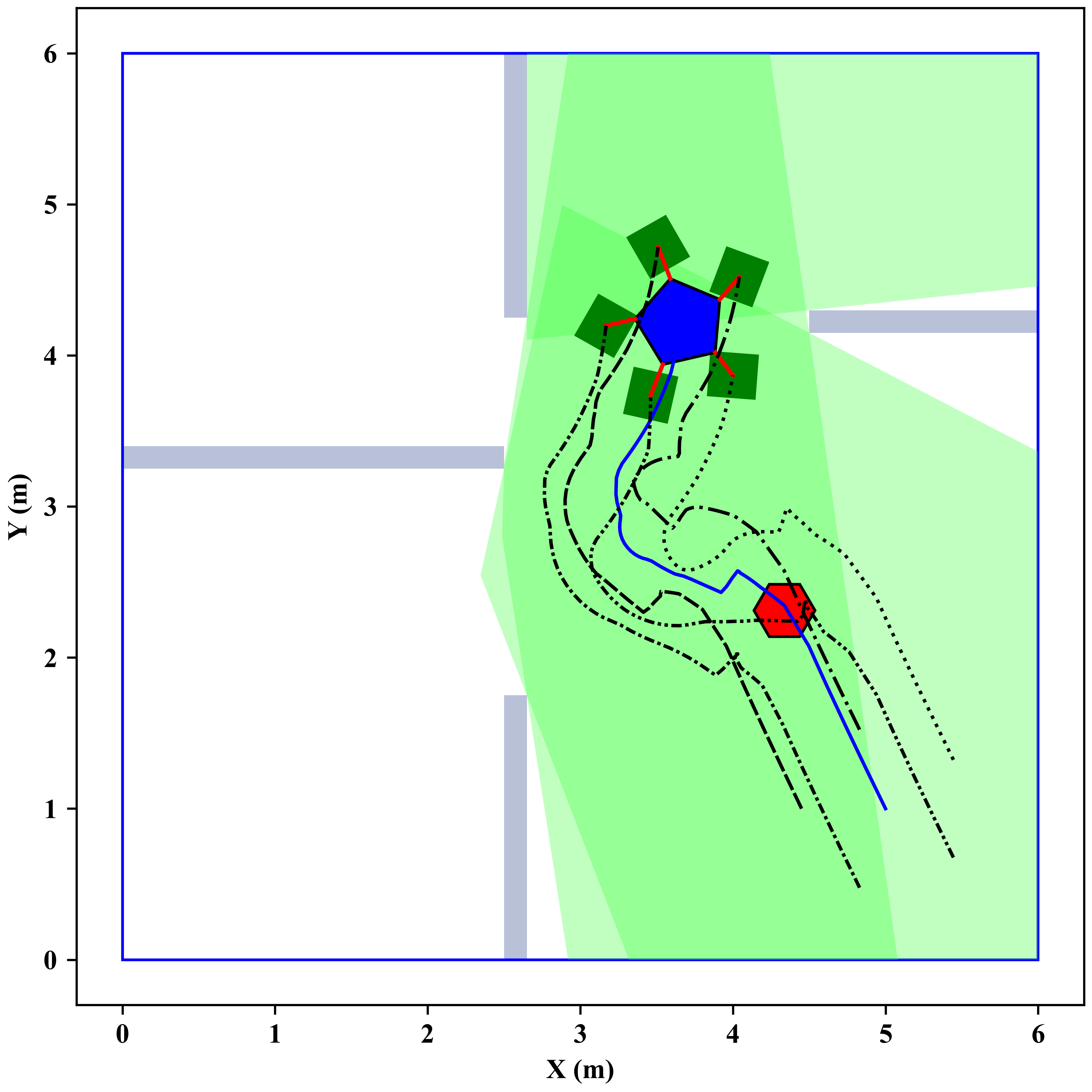}
		\subcaption{\label{fig4c} $t = 30\ s$}
	\end{subfigure}
	\begin{subfigure}{0.24\textwidth}
		\centering
		\includegraphics[width=114px]{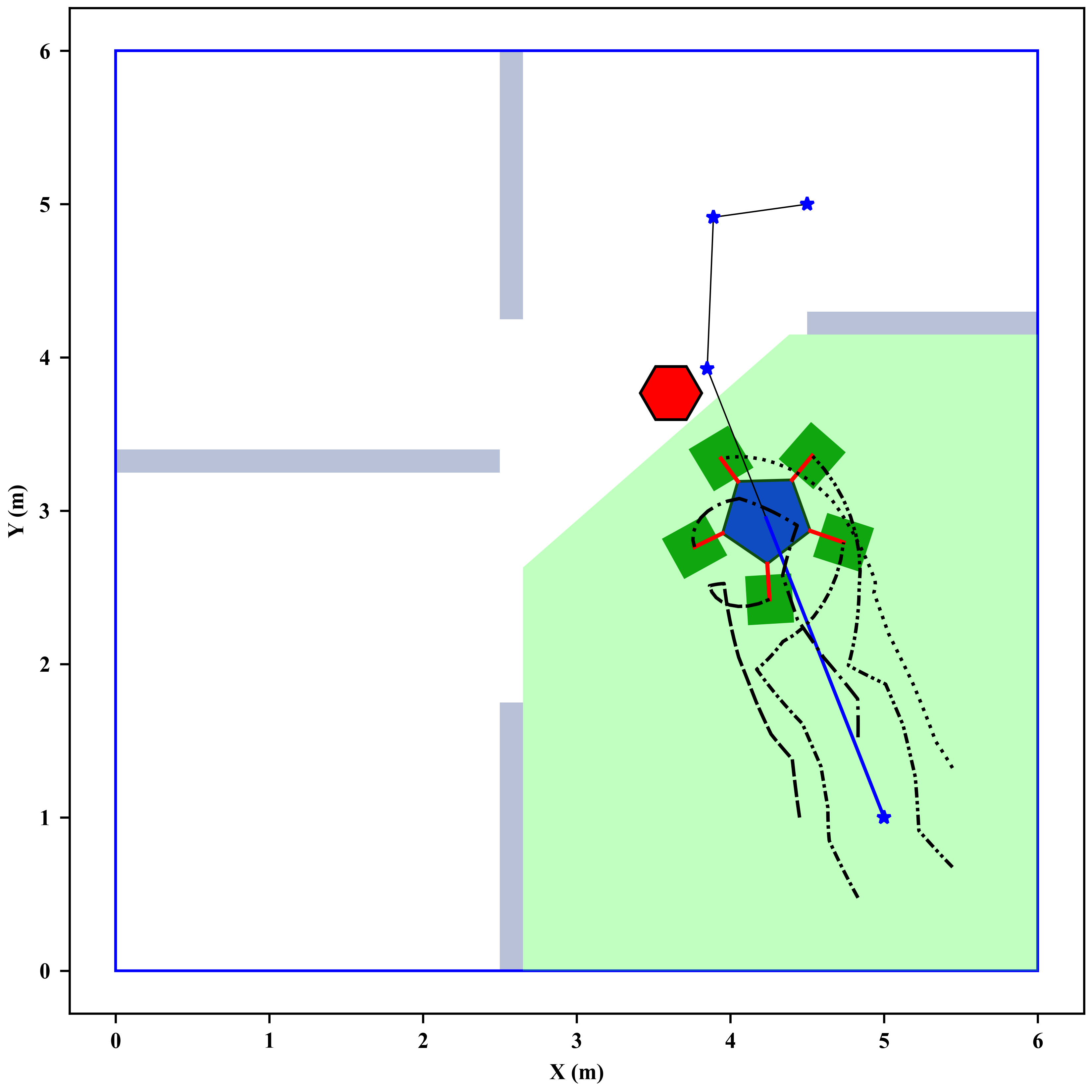}
		\caption{\label{fig4g} $t = 14\ s$}
	\end{subfigure}
	
	\begin{subfigure}{0.24\textwidth}
		\centering
		\includegraphics[width=114px]{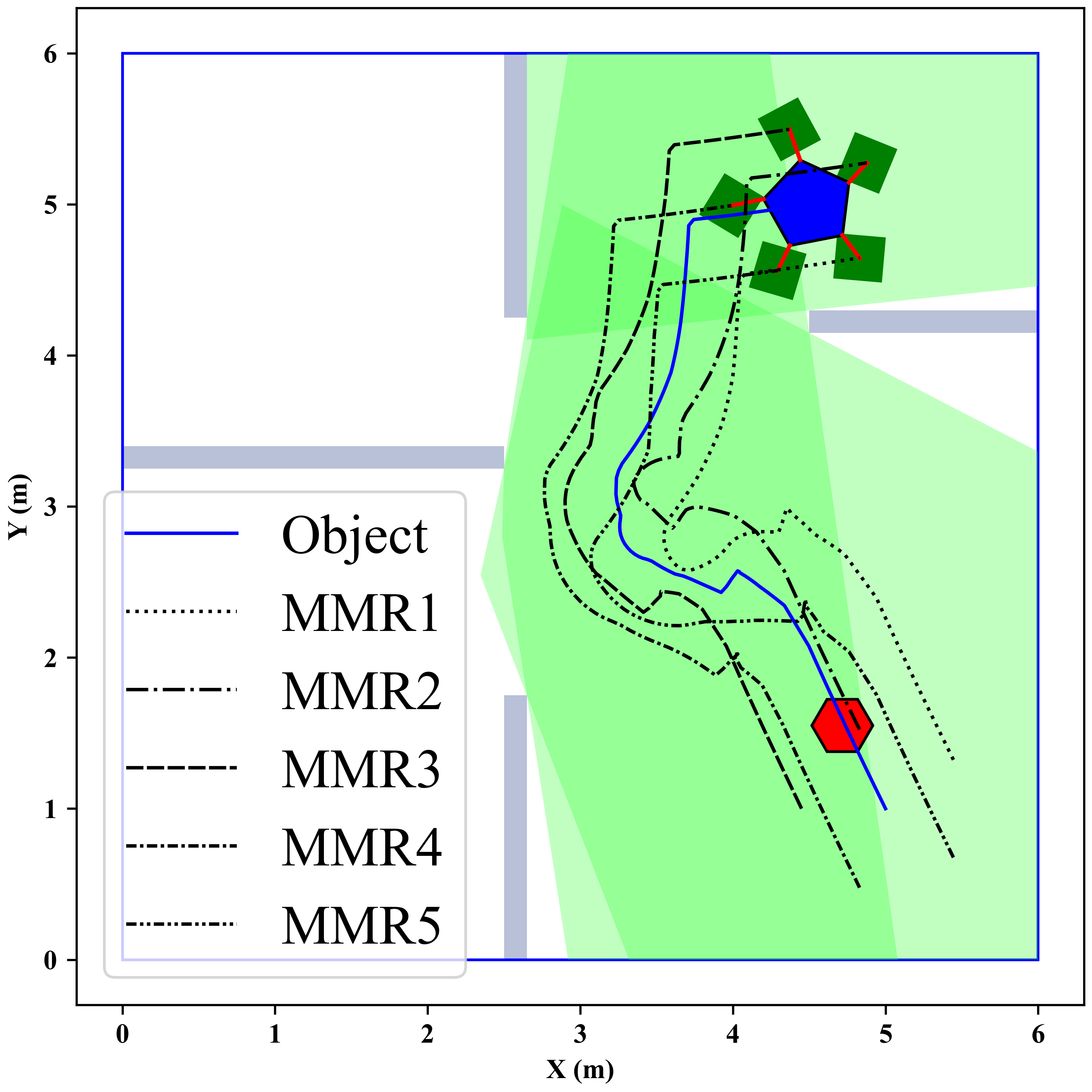}
		\subcaption{\label{fig4d} $t = 38.5\ s$}
	\end{subfigure}
	\begin{subfigure}{0.24\textwidth}
		\centering
		\includegraphics[width=114px]{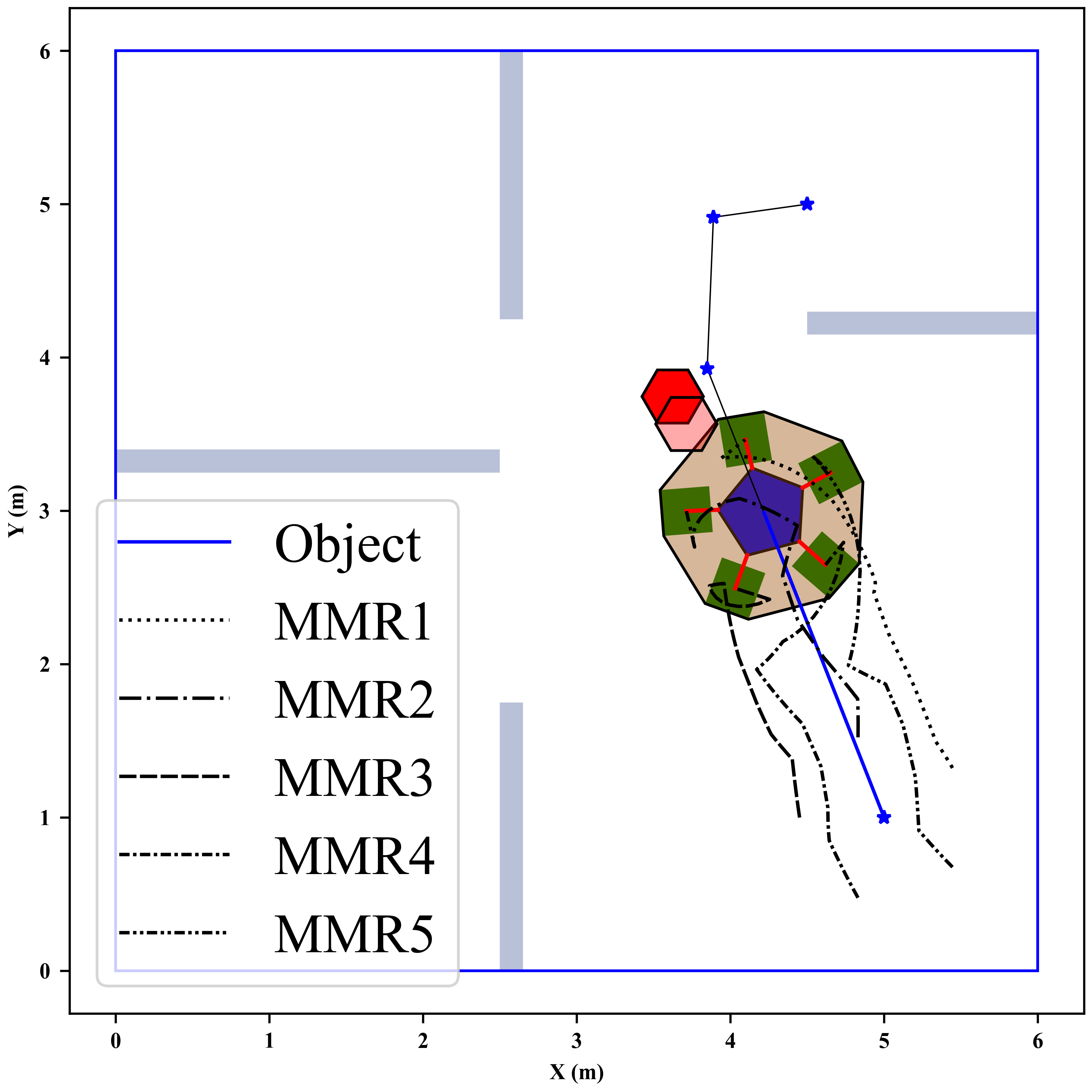}
		\caption{\label{fig4h} $t = 14.25\ s$}
	\end{subfigure}
	\caption{The deep red hexagon is the current state of the obstacles, while the lighter reds are the future states. Figs. in the left column show the motion plan obtained from the proposed algorithm. The motion plan using the algorithm proposed by \textit{Alonso-Mora et al.} \cite{2017_AlonsoMora} is shown in the right column.}
	\label{fig4}
\end{figure}

\begin{figure}[htbp]
	\centerline{\includegraphics[width = 80px]{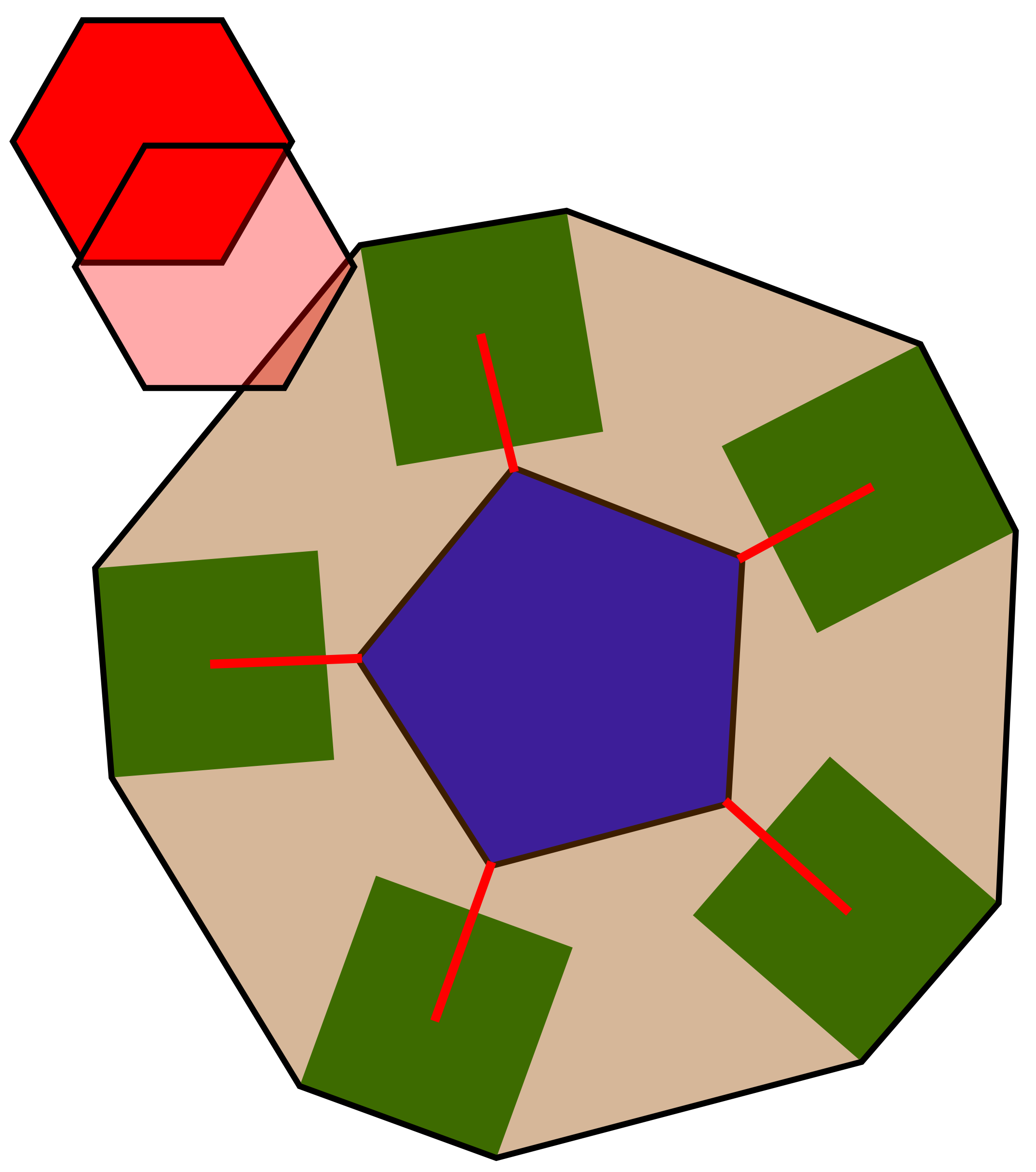}}
	\caption{The deep red hexagon is the dynamic obstacle in the current state, and the light red one is at the end of the horizon $\tau$. The light convex brown region contains the MMRs and the grasped object in the current state. The dynamic obstacle within the horizon intersects with MMRs}
	\label{fig:collision}
\end{figure}

We have separately employed our proposed algorithm and the state-of-the-art algorithm by \textit{AlonsoMora et al.} \cite{2017_AlonsoMora} to plan the motion for cooperative object transportation in the same environment. The results from our proposed motion planning are compared with the other method. Fig. \ref{fig4a}, \ref{fig4b}, \ref{fig4c} and \ref{fig4d} shows four snapshots of object transportation maneuver obtained from our proposed online motion planning algorithm in Section \ref{MotionPlanning}, where the light green regions indicate the static obstacle-free region around the path. The obtained motion plan using the technique by \textit{Alonso-Mora et al.} \cite{2017_AlonsoMora} is presented till the planner fails to plan motion using four snapshots in Fig. \ref{fig4e}, \ref{fig4f}, \ref{fig4g} and \ref{fig4h}. The light green area indicates a complete obstacle-free convex region around the formation for the horizon time $T_h$. The region defined by the dynamic obstacle for the time horizon $T_h$ intersects with the convex hull by the collaborative manipulators shown in Fig. \ref{fig:collision}. Thus, the position-time space embedded obstacle-free convex region-based method for the dynamic obstacle fails to find an obstacle-free convex region at time instant $t=14.25\ s$ shown in Fig. \ref{fig4f} and the MMRs collide with the dynamic obstacle. The failure to find an obstacle-free region at $t= 14\ s$ while the obstacle is very close to the formation leads to a collision with the dynamic obstacle. Compared to the method proposed by \textit{Alonso-Mora et al.} \cite{2017_AlonsoMora}, our proposed algorithm in Section \ref{MotionPlanning} successfully computes a collision-free motion plan.

\textcolor{blue}{
\begin{table}[h]
 	\centering
	\caption{Computation Time (in seconds) Comparison with the state-of-the-art algorithm.}
	\begin{tabular}{@{}llllll@{}}
		\hline
 		  & Min& Mean & Max & SD\\
 		\hline
 		Proposed&$0.27$&$0.48$ & $0.62$& $0.09$ \\
 		\hline
         \textit{AlonsoMora et al.}\cite{2017_AlonsoMora}&$0.46$&$0.86$ & $1.26$& $0.26$ \\
         \hline
	\end{tabular}
	\label{tab:ComputationTimeComp}
 \end{table}
}

We compare the computational time in Python implementation in Table \ref{tab:ComputationTime} for an online motion planning horizon. We have run the motion planning algorithm a Laptop equipped with AMD Rayzen 5800H CPU, 16 GB RAM  Windows 11 OS. The dynamic obstacle trajectory has been altered to $p_{dyn} = [0.5+0.056t, 1.0-0.112t]$, making both planning algorithms work. The computation time statistics are shown in Table \ref{tab:ComputationTimeComp}. The computation time of our proposed algorithm is lower. The proposed approach demonstrates the real-time performance of less than $1\ s$.

\begin{figure}[htbp]
	\centerline{\includegraphics[width = 180px]{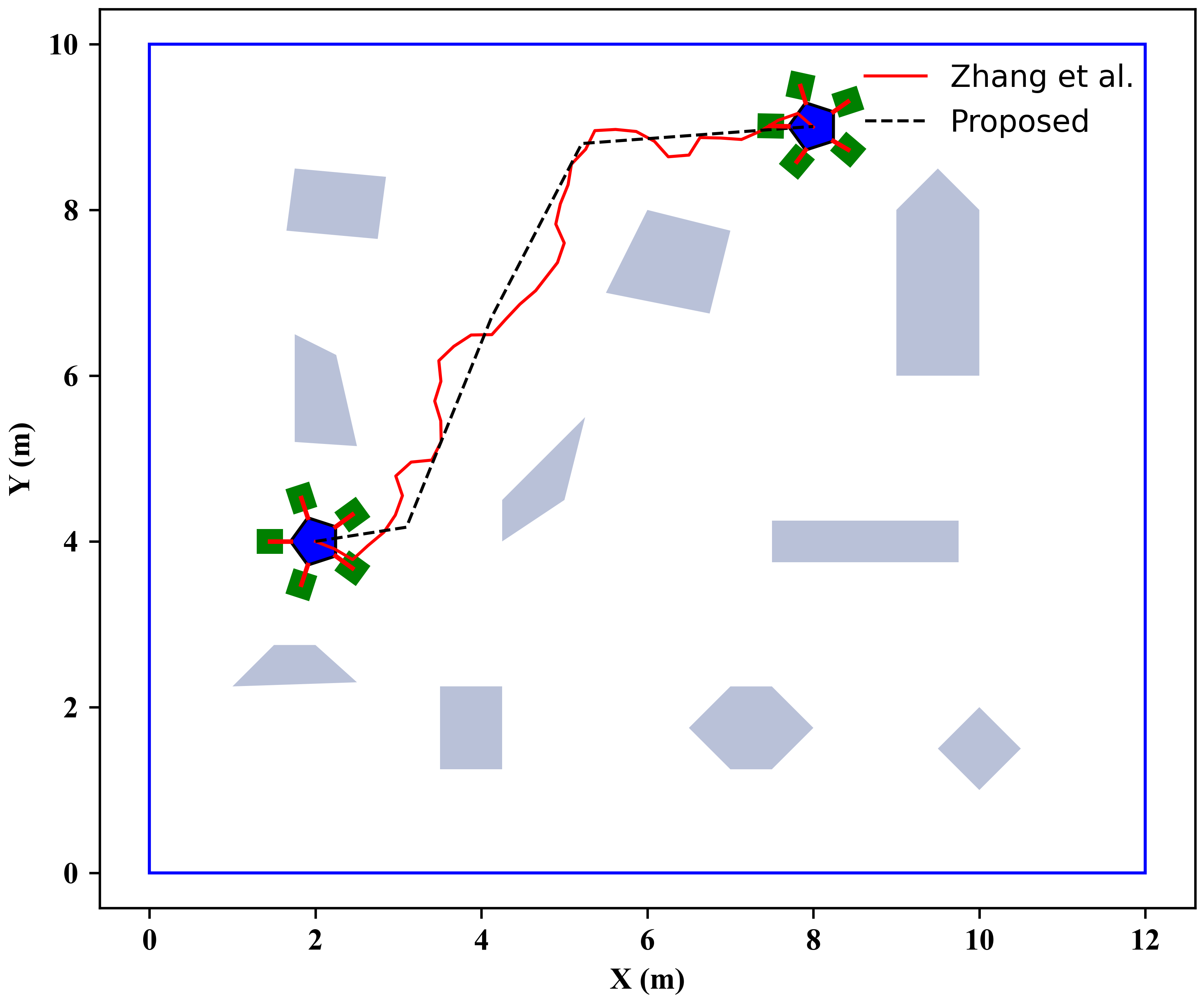}}
	\caption{Comparison of static obstacle free path computed by our proposed algorithm and the algorithm by \textit{Zhang et al.} \cite{2025_zhang}.}
	\label{fig:HP_IRIS_compare}
\end{figure}

We have compared our proposed planner with the algorithm by \textit{Zhang et al.} \cite{2025_zhang}. The planning algorithm by \textit{Zhang et al.} computes paths for static environments. In this comparison, we compare the path planning algorithm's performance. Both path-planning algorithms are stochastic. Hence, every plan differs from the previous. We have simulated both algorithms 10 times and compared the path lengths in Table \ref{tab:PathComp}. The computed path length by our proposed planner is less, indicating the efficiency of the proposed path planner. Fig. \ref{fig:HP_IRIS_compare} shows the minimum length path of the computed paths by both algorithms.

\begin{table}[h]
	\centering
	\caption{Path length (in meter) comparison.}
	\begin{tabular}{@{}lllll@{}}
		\hline
		& Min& Mean & Max & SD\\
		\hline
		Proposed&$9.00$&$9.09$ & $9.47$& $0.18$ \\
		\hline
		 \textit{Zhang et al.} \cite{2025_zhang}&$10.18$&$11.07$ & $11.89$& $0.58$ \\
		\hline
	\end{tabular}
	\label{tab:PathComp}
\end{table}
\section{Conclusion}
The proposed NMPC-based motion planning algorithm generates kinodynamically feasible, collision-free trajectories of MMRs in dynamic environments by simultaneously optimizing both the robot's base and the arm. Our algorithm identifies narrow regions between static obstacles, such as doors and corridors, which more accurately defines the obstacle-free polygons in those areas, enhancing the feasibility of finding a feasible path. We successfully demonstrated the algorithm's effectiveness in transporting objects in an environment with single and dual dynamic obstacles. Our analysis of computational times across various simulation environments and real experiments shows that the algorithm can be implemented in real-time with grasp error within $\pm 0.013m$. In a comparison study with state-of-the-art motion planning techniques, our approach outperformed others, with some competing methods failing and lower computational time 0.62 s, nearly $50\%$ less. Our planning algorithm is designed to work directly for convex static obstacles. We use two approaches for non-convex obstacles: one involves obtaining the convex hull of the obstacles with any convex hull computation algorithm, and the other splits the non-convex obstacle into multiple non-overlapping convex obstacles. 

Cooperative object transportation may encounter a challenge due to the wheel slip of the MMR base and the rigid connection between the object and the manipulators. Our simulations indicate that the motion planner may fail if the obstacle velocities exceed the maximum attainable formation velocities. We plan to develop a distributive motion planning algorithm with efficient online task allocation in the future.
\section*{Appendix}\label{append:1}
The joint position of $i-th$ MMR's manipulator arm shown in Fig. \ref{fig:1a}, whose base is attached to the mobile base at $p_i$, is indicated by $q_{f,i}=[q_{f,i1},q_{f,i2},\cdots,q_{f,in_i}]^T$. The EE position $p_{ee,i} = p_i + f_{xy,i}(q_{f, i})$, where $f_{xy,i}$ is a mapping function of 2D task space (Fig. \ref{fig:1b}) to the joint space. The first and the third joints of the reduced joint space $q_{a,i}=[q_{i1},q_{i2},q_{i3}]^T$ are revolute and placed at the two ends of the manipulator arm, and the second is a prismatic joint placed in the middle of the manipulators. The reduced revoule-prismatic-revolute (RPR) joint space with a fixed height $f_{z,i}(q_{f, i}) = constant$ exactly represents the manipulator arm in a 3D workspace. The original joint displacement $q_{f,i}$ is mapped into $q_{a,i}$. The mapping is $\boldsymbol{f}_{xy,i}(q_{f, i}) =\ <q_{i2},\ q_{i1}>$, where $q_{i2} = \vert \boldsymbol{f}_{xy,i}(q_{f, i})\vert$, $q_{i1}$ and $q_{i3}$ are the angle of the vector value of $\boldsymbol{f}_{xy,i}(q_{af, i})$ measured at $\{b_i\}$ and $\{o\}$ respectively.


\normalsize
\bibliography{ref}

@article{1996_khatib,
author = {Khatib, O. and Yokoi, K. and Chang, K. and Ruspini, D. and Holmberg, R. and Casal, A.},
title = {Coordination and decentralized cooperation of multiple mobile manipulators},
journal = {Journal of Robotic Systems},
volume = {13},
number = {11},
pages = {755-764},
doi = {https://doi/10.1002/(SICI)1097-4563(199611)13:11<755::AID-ROB6>3.0.CO;2-U},
url = {https://onlinelibrary.wiley.com/https://onlinelibrary.wiley.com/doi/10.1002/(SICI)1097-4563(199611)13:11<755::AID-ROB6>3.0.CO;2-U},
year = {1996}
}

@INPROCEEDINGS{1997_Desai,
	author={Desai, J.P. and Kumar, V.},
	booktitle={Proceedings of International Conference on Robotics and Automation},
	title={Nonholonomic motion planning for multiple mobile manipulators},
	year={1997},
	volume={4},
	number={},
	pages={3409-3414 vol.4},
	doi={10.1109/ROBOT.1997.606863}}

@ARTICLE{2003_Tanner,
  author={Tanner, H.G. and Loizou, S.G. and Kyriakopoulos, K.J.},
  journal={IEEE Transactions on Robotics and Automation}, 
  title={Nonholonomic navigation and control of cooperating mobile manipulators}, 
  year={2003},
  volume={19},
  number={1},
  pages={53-64},
  doi={10.1109/TRA.2002.807549}}

@INPROCEEDINGS{2015_Deits,
	author={Deits, Robin and Tedrake, Russ},
	booktitle={2015 IEEE International Conference on Robotics and Automation (ICRA)}, 
	title={Efficient mixed-integer planning for UAVs in cluttered environments}, 
	year={2015},
	volume={},
	number={},
	pages={42-49},
	doi={10.1109/ICRA.2015.7138978}}

@INPROCEEDINGS{2013_Yang,
  author={Hyunsoo Yang and Lee, Dongjun},
  booktitle={2013 IEEE International Conference on Robotics and Automation}, 
  title={Cooperative grasping control of multiple mobile manipulators with obstacle avoidance}, 
  year={2013},
  volume={},
  number={},
  pages={836-841},
  doi={10.1109/ICRA.2013.6630670}}

@INPROCEEDINGS{2013_Erhart,
	author={Erhart, Sebastian and Hirche, Sandra},
	booktitle={2013 IEEE/RSJ International Conference on Intelligent Robots and Systems}, 
	title={Adaptive force/velocity control for multi-robot cooperative manipulation under uncertain kinematic parameters}, 
	year={2013},
	volume={},
	number={},
	pages={307-314},
	doi={10.1109/IROS.2013.6696369}}

@article{2017_AlonsoMora,
	author = {Javier Alonso-Mora and Stuart Baker and Daniela Rus},
	title ={Multi-robot formation control and object transport in dynamic environments via constrained optimization},
	journal = {The International Journal of Robotics Research},
	volume = {36},
	number = {9},
	pages = {1000-1021},
	year = {2017},
	doi = {10.1177/0278364917719333},
	URL = {https://doi.org/10.1177/0278364917719333},
	eprint = {https://doi.org/10.1177/0278364917719333}}

@ARTICLE{2017_Jiao,
  author={Jiao, Jile and Cao, Zhiqiang and Gu, Nong and Nahavandi, Saeid and Yang, Yuequan and Tan, Min},
  journal={IEEE Systems Journal}, 
  title={Transportation by Multiple Mobile Manipulators in Unknown Environments With Obstacles}, 
  year={2017},
  volume={11},
  number={4},
  pages={2894-2904},
  doi={10.1109/JSYST.2015.2416215}}

@ARTICLE{2018_Cao,
  author={Cao, Zhiqiang and Gu, Nong and Jiao, Jile and Nahavandi, Saeid and Zhou, Chao and Tan, Min},
  journal={IEEE Systems Journal}, 
  title={A Novel Geometric Transportation Approach for Multiple Mobile Manipulators in Unknown Environments}, 
  year={2018},
  volume={12},
  number={2},
  pages={1447-1455},
  doi={10.1109/JSYST.2016.2581171}}

@INPROCEEDINGS{2019_Tallamraju,
  author={Tallamraju, Rahul and Salunkhe, Durgesh H. and Rajappa, Sujit and Ahmad, Aamir and Karlapalem, Kamalakar and Shah, Suril V.},
  booktitle={2019 IEEE 15th International Conference on Automation Science and Engineering (CASE)}, 
  title={Motion Planning for Multi-Mobile-Manipulator Payload Transport Systems}, 
  year={2019},
  volume={},
  number={},
  pages={1469-1474},
  doi={10.1109/COASE.2019.8842840}}

@article{1993_Donald,
	author = {Donald, Bruce and Xavier, Patrick and Canny, John and Reif, John},
	title = {Kinodynamic Motion Planning},
	year = {1993},
	issue_date = {Nov. 1993},
	publisher = {Association for Computing Machinery},
	address = {New York, NY, USA},
	volume = {40},
	number = {5},
	issn = {0004-5411},
	url = {https://doi.org/10.1145/174147.174150},
	doi = {10.1145/174147.174150},
	journal = {J. ACM},
	month = {nov},
	pages = {1048–1066},
	numpages = {19}
}

@ARTICLE{1996_Kavraki,
  author={Kavraki, L.E. and Svestka, P. and Latombe, J.-C. and Overmars, M.H.},
  journal={IEEE Transactions on Robotics and Automation}, 
  title={Probabilistic roadmaps for path planning in high-dimensional configuration spaces}, 
  year={1996},
  volume={12},
  number={4},
  pages={566-580},
  doi={10.1109/70.508439}}

@article{2001_Lavalle,
    author = {Steven M. LaValle and James J. Kuffner, Jr.},
    title ={Randomized Kinodynamic Planning},
    journal = {The International Journal of Robotics Research},
    volume = {20},
    number = {5},
    pages = {378-400},
    year = {2001},
    doi = {10.1177/02783640122067453},
    URL = {https://doi.org/10.1177/02783640122067453},
    eprint = {https://doi.org/10.1177/02783640122067453}
}

@book{2006_Lavalle,
    place={Cambridge},
    title={Planning Algorithms},
    DOI={10.1017/CBO9780511546877},
    publisher={Cambridge University Press},
    author={LaValle, Steven M.},
    year={2006}}

@INPROCEEDINGS{2014_Saha,
  author={Saha, Indranil and Ramaithitima, Rattanachai and Kumar, Vijay and Pappas, George J. and Seshia, Sanjit A.},
  booktitle={2014 IEEE/RSJ International Conference on Intelligent Robots and Systems}, 
  title={Automated composition of motion primitives for multi-robot systems from safe LTL specifications}, 
  year={2014},
  volume={},
  number={},
  pages={1525-1532},
  doi={10.1109/IROS.2014.6942758}}

@Article{2019_Andersson,
  author = {Joel A E Andersson and Joris Gillis and Greg Horn
            and James B Rawlings and Moritz Diehl},
  title = {{CasADi} -- {A} software framework for nonlinear optimization
           and optimal control},
  journal = {Mathematical Programming Computation},
  volume = {11},
  number = {1},
  pages = {1--36},
  year = {2019},
  publisher = {Springer},
  doi = {10.1007/s12532-018-0139-4}
}

@INPROCEEDINGS{2020_Shorinwa,
  author={Shorinwa, Ola and Schwager, Mac},
  booktitle={2020 IEEE/RSJ International Conference on Intelligent Robots and Systems (IROS)}, 
  title={Scalable Collaborative Manipulation with Distributed Trajectory Planning}, 
  year={2020},
  volume={},
  number={},
  pages={9108-9115},
  doi={10.1109/IROS45743.2020.9340957}}

@ARTICLE{2021_Wu,
  author={Wu, Chu and Fang, Hao and Yang, Qingkai and Zeng, Xianlin and Wei, Yue and Chen, Jie},
  journal={IEEE Transactions on Cybernetics}, 
  title={Distributed Cooperative Control of Redundant Mobile Manipulators With Safety Constraints}, 
  year={2021},
  volume={},
  number={},
  pages={1-13},
  doi={10.1109/TCYB.2021.3104044}}

@INPROCEEDINGS{2021_Spahn,
  author={Spahn, Max and Brito, Bruno and Alonso-Mora, Javier},
  booktitle={2021 IEEE International Conference on Robotics and Automation (ICRA)}, 
  title={Coupled Mobile Manipulation via Trajectory Optimization with Free Space Decomposition}, 
  year={2021},
  volume={},
  number={},
  pages={12759-12765},
  doi={10.1109/ICRA48506.2021.9561821}}

@INPROCEEDINGS{2022_Dhaval,
  author={Gujarathi, Dhaval and Saha, Indranil},
  booktitle={2022 IEEE/RSJ International Conference on Intelligent Robots and Systems (IROS)}, 
  title={MT*: Multi-Robot Path Planning for Temporal Logic Specifications}, 
  year={2022},
  volume={},
  number={},
  pages={13692-13699},
  doi={10.1109/IROS47612.2022.9981504}}

@article{2013_Ulusoy,
author = {Alphan Ulusoy and Stephen L. Smith and Xu Chu Ding and Calin Belta and Daniela Rus},
title ={Optimality and Robustness in Multi-Robot Path Planning with Temporal Logic Constraints},
journal = {The International Journal of Robotics Research},
volume = {32},
number = {8},
pages = {889-911},
year = {2013},
doi = {10.1177/0278364913487931},
URL = {https://doi.org/10.1177/0278364913487931},
eprint = {https://doi.org/10.1177/0278364913487931}
}

@techreport{1998_Lavalle,
	author = {Steven M. LaValle},
	title = {Rapidly-exploring random trees: A new tool for path planning},
	type = {Technical Report},
	institution = {Iowa State University},
	month = {October},
	year = {1998},
	address = {Ames, IA, USA},
	volume = {TR 98-11}
}

@ARTICLE{2022_Pan,
  author={Pan, Zhenhua and Zhang, Chengxi and Xia, Yuanqing and Xiong, Hao and Shao, Xiaodong},
  journal={IEEE Transactions on Circuits and Systems II: Express Briefs}, 
  title={An Improved Artificial Potential Field Method for Path Planning and Formation Control of the Multi-UAV Systems}, 
  year={2022},
  volume={69},
  number={3},
  pages={1129-1133},
  doi={10.1109/TCSII.2021.3112787}}

@ARTICLE{2018_Wen,
  author={Wen, Guoxing and Chen, C. L. Philip and Liu, Yan-Jun},
  journal={IEEE Transactions on Industrial Electronics}, 
  title={Formation Control With Obstacle Avoidance for a Class of Stochastic Multiagent Systems}, 
  year={2018},
  volume={65},
  number={7},
  pages={5847-5855},
  doi={10.1109/TIE.2017.2782229}}

@article{2024_Liu,
title = {A path planning algorithm for three-dimensional collision avoidance based on potential field and B-spline boundary curve},
journal = {Aerospace Science and Technology},
volume = {144},
pages = {108763},
year = {2024},
issn = {1270-9638},
doi = {https://doi.org/10.1016/j.ast.2023.108763},
url = {https://www.sciencedirect.com/science/article/pii/S1270963823006594},
author = {Mingjie Liu and Hongxin Zhang and Jian Yang and Tiezhu Zhang and Caihong Zhang and Lan Bo},
}

@ARTICLE{2020_Liu,
  author={Liu, Ya and Huang, Panfeng and Zhang, Fan and Zhao, Yakun},
  journal={IEEE Transactions on Control Systems Technology}, 
  title={Distributed Formation Control Using Artificial Potentials and Neural Network for Constrained Multiagent Systems}, 
  year={2020},
  volume={28},
  number={2},
  pages={697-704},
  doi={10.1109/TCST.2018.2884226}}

@ARTICLE{2016_Faulwasser,
  author={Faulwasser, Timm and Findeisen, Rolf},
  journal={IEEE Transactions on Automatic Control}, 
  title={Nonlinear Model Predictive Control for Constrained Output Path Following}, 
  year={2016},
  volume={61},
  number={4},
  pages={1026-1039},
  doi={10.1109/TAC.2015.2466911}}

@INPROCEEDINGS{2016_Petitti,
	author={Petitti, Antonio and Franchi, Antonio and Di Paola, Donato and Rizzo, Alessandro},
	booktitle={2016 IEEE International Conference on Robotics and Automation (ICRA)}, 
	title={Decentralized motion control for cooperative manipulation with a team of networked mobile manipulators}, 
	year={2016},
	volume={},
	number={},
	pages={441-446},
	doi={10.1109/ICRA.2016.7487164}}

@article{2020_Mao,
   author = {Run Mao and Hongli Gao and Liang Guo},
   doi = {10.1155/2020/4183427},
   issn = {1024-123X},
   journal = {Mathematical Problems in Engineering},
   month = {6},
   pages = {1-16},
   title = {A Novel Collision-Free Navigation Approach for Multiple Nonholonomic Robots Based on ORCA and Linear MPC},
   volume = {2020},
   year = {2020},
}

@ARTICLE{2021_Zhang,
  author={Zhang, Xiaojing and Liniger, Alexander and Borrelli, Francesco},
  journal={IEEE Transactions on Control Systems Technology}, 
  title={Optimization-Based Collision Avoidance}, 
  year={2021},
  volume={29},
  number={3},
  pages={972-983},
  doi={10.1109/TCST.2019.2949540}}

@ARTICLE{2022_Hu,
  author={Hu, Jiawei and Liu, Wenhang and Zhang, Heng and Yi, Jingang and Xiong, Zhenhua},
  journal={IEEE Robotics and Automation Letters}, 
  title={Multi-Robot Object Transport Motion Planning With a Deformable Sheet}, 
  year={2022},
  volume={7},
  number={4},
  pages={9350-9357},
  doi={10.1109/LRA.2022.3191190}}

@article{2023_Xu,
   author = {Pengjie Xu and Jingtao Zhang and Yuanzhe Cui and Kun Zhang and Qirong Tang},
   doi = {10.1007/s12555-021-0925-z},
   issn = {1598-6446},
   issue = {4},
   journal = {International Journal of Control, Automation and Systems},
   month = {4},
   pages = {1296-1308},
   title = {Modeling and Coordinated Control of Multiple Mobile Manipulators with Closed-chain Constraints},
   volume = {21},
   year = {2023},
}

@ARTICLE{2004_Belta,
  author={Belta, C. and Kumar, V.},
  journal={IEEE Transactions on Robotics}, 
  title={Abstraction and control for Groups of robots}, 
  year={2004},
  volume={20},
  number={5},
  pages={865-875},
  doi={10.1109/TRO.2004.829498}}

@inproceedings{2008_Michael,
   author = {Nathan Michael and Vijay Kumar},
   doi = {10.15607/RSS.2008.IV.006},
   isbn = {9780262513098},
   journal = {Robotics: Science and Systems IV},
   month = {6},
   publisher = {Robotics: Science and Systems Foundation},
   title = {Controlling Shapes of Ensembles of Robots of Finite Size with Nonholonomic Constraints},
   year = {2008},
}

@ARTICLE{2001_Egerstedt,
  author={Egerstedt, M. and Xiaoming Hu},
  journal={IEEE Transactions on Robotics and Automation}, 
  title={Formation constrained multi-agent control}, 
  year={2001},
  volume={17},
  number={6},
  pages={947-951},
  doi={10.1109/70.976029}}

@ARTICLE{2024_Pei,
  author={Pei, Liuao and Lin, Junxiao and Han, Zhichao and Quan, Lun and Cao, Yanjun and Xu, Chao and Gao, Fei},
  journal={IEEE Robotics and Automation Letters}, 
  title={Collaborative Planning for Catching and Transporting Objects in Unstructured Environments}, 
  year={2024},
  volume={9},
  number={2},
  pages={1098-1105},
  doi={10.1109/LRA.2023.3335770}}

@INPROCEEDINGS{2012_Krontiris,
	author={Krontiris, Athanasios and Louis, Sushil and Bekris, Kostas E.},
	booktitle={2012 IEEE International Conference on Robotics and Automation}, 
	title={Multi-level formation roadmaps for collision-free dynamic shape changes with non-holonomic teams}, 
	year={2012},
	volume={},
	number={},
	pages={1570-1575},
	doi={10.1109/ICRA.2012.6225372}}

@INPROCEEDINGS{2018_Culbertson,
	author={Culbertson, Preston and Schwager, Mac},
	booktitle={2018 IEEE International Conference on Robotics and Automation (ICRA)}, 
	title={Decentralized Adaptive Control for Collaborative Manipulation}, 
	year={2018},
	volume={},
	number={},
	pages={278-285},
	doi={10.1109/ICRA.2018.8461263}}

@ARTICLE{2017_Dai,
	author={Dai, Gong-Bo and Liu, Yen-Chen},
	journal={IEEE Transactions on Industrial Electronics}, 
	title={Distributed Coordination and Cooperation Control for Networked Mobile Manipulators}, 
	year={2017},
	volume={64},
	number={6},
	pages={5065-5074},
	doi={10.1109/TIE.2016.2642880}}

@ARTICLE{2018_Marino,
	author={Marino, Alessandro},
	journal={IEEE Transactions on Control Systems Technology}, 
	title={Distributed Adaptive Control of Networked Cooperative Mobile Manipulators}, 
	year={2018},
	volume={26},
	number={5},
	pages={1646-1660},
	doi={10.1109/TCST.2017.2720673}}

@ARTICLE{2020_Ren,
	author={Ren, Yi and Sosnowski, Stefan and Hirche, Sandra},
	journal={IEEE Transactions on Robotics}, 
	title={Fully Distributed Cooperation for Networked Uncertain Mobile Manipulators}, 
	year={2020},
	volume={36},
	number={4},
	pages={984-1003},
	doi={10.1109/TRO.2020.2971416}}

@INPROCEEDINGS{2018_Verginis,
  author={Verginis, Christos K. and Nikou, Alexandros and Dimarogonas, Dimos V.},
  booktitle={2018 European Control Conference (ECC)}, 
  title={Communication-based Decentralized Cooperative Object Transportation Using Nonlinear Model Predictive Control}, 
  year={2018},
  volume={},
  number={},
  pages={733-738},
  doi={10.23919/ECC.2018.8550305}}

@INPROCEEDINGS{2017_Nikou,
  author={Nikou, Alexandres and Verginis, Christos and Heshmati-alamdari, Shahab and Dimarogonas, Dimos V.},
  booktitle={2017 25th Mediterranean Conference on Control and Automation (MED)}, 
  title={A Nonlinear Model Predictive Control scheme for cooperative manipulation with singularity and collision avoidance}, 
  year={2017},
  volume={},
  number={},
  pages={707-712},
  doi={10.1109/MED.2017.7984201}}

@ARTICLE{2022_Vlantis,
	author={Vlantis, Panagiotis and Bechlioulis, Charalampos P. and Kyriakopoulos, Kostas J.},
	journal={Robotics},
	title={Mutli-Robot Cooperative Object Transportation with Guaranteed Safety and Convergence in Planar Obstacle Cluttered Workspaces via Configuration Space Decomposition},
	volume={11},
	year={2022},
	number={6},
	doi={10.3390/robotics11060148}}

@ARTICLE{2021_Koung,
	author={Koung, Daravuth and Kermorgant, Olivier and Fantoni, Isabelle and Belouaer, Lamia},
	journal={IEEE Robotics and Automation Letters}, 
	title={Cooperative Multi-Robot Object Transportation System Based on Hierarchical Quadratic Programming}, 
	year={2021},
	volume={6},
	number={4},
	pages={6466-6472},
	doi={10.1109/LRA.2021.3092305}}

@article{2018_Tuci,
	author = {Elio Tuci and Muhanad H. M. Alkilabi and Otar Akanyeti},
	doi = {10.3389/frobt.2018.00059},
	issn = {2296-9144},
	journal = {Frontiers in Robotics and AI},
	month = {5},
	title = {Cooperative Object Transport in Multi-Robot Systems: A Review of the State-of-the-Art},
	volume = {5},
	year = {2018},
}

@article{2020_Feng,
	author = {Zhi Feng and Guoqiang Hu and Yajuan Sun and Jeffrey Soon},
	doi = {10.1016/j.arcontrol.2020.02.002},
	issn = {13675788},
	journal = {Annual Reviews in Control},
	title = {An overview of collaborative robotic manipulation in multi-robot systems},
	volume = {49},
	year = {2020},
}

@article{2021_Ramalepa,
	author = {Larona Pitso Ramalepa and Rodrigo S. Jamisola},
	doi = {10.1007/s11633-021-1299-7},
	issn = {1476-8186},
	issue = {4},
	journal = {International Journal of Automation and Computing},
	month = {8},
	pages = {536-555},
	title = {A Review on Cooperative Robotic Arms with Mobile or Drones Bases},
	volume = {18},
	year = {2021},
}

@ARTICLE{2024_Jaafar,
	author={Jaafar, Hussein Ali and Kao, Cheng-Hao and Saeedi, Sajad},
	journal={IEEE Control Systems Letters}, 
	title={MR.CAP: Multi-Robot Joint Control and Planning for Object Transport}, 
	year={2024},
	volume={8},
	number={},
	pages={139-144},
	doi={10.1109/LCSYS.2024.3349989}}

@ARTICLE{2024_Kennel,
  author={Kennel-Maushart, Florian and Coros, Stelian},
  journal={IEEE Robotics and Automation Letters}, 
  title={Payload-Aware Trajectory Optimisation for Non-Holonomic Mobile Multi-Robot Manipulation With Tip-Over Avoidance}, 
  year={2024},
  volume={9},
  number={9},
  pages={7669-7676},
  doi={10.1109/LRA.2024.3427555}}

@article{2008_Cheng,
	author = {Cheng, Peng and Fink, Jonathan and Kumar, Vijay and Pang, Jong-Shi},
	title = {Cooperative Towing With Multiple Robots},
	journal = {Journal of Mechanisms and Robotics},
	volume = {1},
	number = {1},
	pages = {011008},
	year = {2008},
	month = {08},
	issn = {1942-4302},
	doi = {10.1115/1.2960539},
	url = {https://doi.org/10.1115/1.2960539},
}

@article{2018_Venkatesan,
	author = {Venkatesan, Vinoth and Seymour, Joseph and Cappelleri, David J.},
	title = {Micro-Assembly Sequence and Path Planning Using Subassemblies},
	journal = {Journal of Mechanisms and Robotics},
	volume = {10},
	number = {6},
	pages = {061015},
	year = {2018},
	month = {10},
	issn = {1942-4302},
	doi = {10.1115/1.4041333},
	url = {https://doi.org/10.1115/1.4041333},
}

@article{2022_Shantanu,
	author = {Thakar, Shantanu and Srinivasan, Srivatsan and Al-Hussaini, Sarah and Bhatt, Prahar M. and Rajendran, Pradeep and Jung Yoon, Yeo and Dhanaraj, Neel and Malhan, Rishi K. and Schmid, Matthias and Krovi, Venkat N. and Gupta, Satyandra K.},
	title = {A Survey of Wheeled Mobile Manipulation: A Decision-Making Perspective},
	journal = {Journal of Mechanisms and Robotics},
	volume = {15},
	number = {2},
	pages = {020801},
	year = {2022},
	month = {07},
	issn = {1942-4302},
	doi = {10.1115/1.4054611},
	url = {https://doi.org/10.1115/1.4054611},
}

@article{2023_Fan,
	author = {Fan, Changxiang and Zeng, Fan and Shirafuji, Shouhei and Ota, Jun},
	title = {Development of a Three-Mobile-Robot System for Cooperative Transportation},
	journal = {Journal of Mechanisms and Robotics},
	volume = {16},
	number = {2},
	pages = {021008},
	year = {2023},
	month = {03},
	issn = {1942-4302},
	doi = {10.1115/1.4056771},
	url = {https://doi.org/10.1115/1.4056771},
}

@article{2024_Borate,
	author = {Borate, Suraj and Rana, Rwik and Venkatesh, Praveen and Vadali, Madhu},
	title = {FF-RRT*: A Sampling-Based Planner for Multirobot Global Formation Path Planning},
	journal = {Journal of Mechanisms and Robotics},
	volume = {16},
	number = {10},
	pages = {104502},
	year = {2024},
	month = {09},
	issn = {1942-4302},
	doi = {10.1115/1.4066129},
	url = {https://doi.org/10.1115/1.4066129},
}

@article{2025_Li,
	author = {Li, Ce and He, Zhiyuan and Zheng, Huijiang},
	year = {2025},
	month = {03},
	pages = {1-49},
	title = {A dual-robot system design for in-cabin pose adjusting of heavy parts},
	journal = {Journal of Mechanisms and Robotics},
	doi = {10.1115/1.4068313}
}

@misc{2025_zhang,
	title={A Novel Semi-Coupled Hierarchical Motion Planning Framework for Cooperative Transportation of Multiple Mobile Manipulators}, 
	author={Heng Zhang and Haoyi Song and Wenhang Liu and Xinjun Sheng and Zhenhua Xiong and Xiangyang Zhu},
	year={2025},
	eprint={2208.08054},
	archivePrefix={arXiv},
	primaryClass={cs.RO},
	url={https://arxiv.org/abs/2208.08054}, 
}


\end{document}